%% file: colm2026_conference.tex
\documentclass{article} 
\usepackage[final]{colm2026_conference}

\usepackage{microtype}
\usepackage{hyperref}
\usepackage{url}
\usepackage{booktabs}
\usepackage{graphicx}
\usepackage{caption}
\usepackage{CJKutf8}
\usepackage{subcaption}
\usepackage{multirow}
\usepackage{makecell}
\usepackage{booktabs}
\usepackage{multirow}
\usepackage{makecell}
\usepackage{tabularx}
\usepackage{siunitx}
\usepackage{array}
\usepackage{lipsum}
\usepackage{amsmath}
\usepackage{graphicx}
\usepackage{wrapfig}

\usepackage[table]{xcolor}
\usepackage{lineno}
\usepackage{amssymb}

\definecolor{correctgreen}{HTML}{1D9E75}
\definecolor{partialamber}{HTML}{BA7517}
\definecolor{wrongred}{HTML}{E24B4A}

\newcommand{\correct}[1]{\textcolor{correctgreen}{\textbf{#1}}}
\newcommand{\partialcolor}[1]{\textcolor{partialamber}{\textbf{#1}}}
\newcommand{\wrong}[1]{\textcolor{wrongred}{\textbf{#1}}}

\newcommand{\squiggleZ}{\raisebox{-0.35\height}{\includegraphics[height=2em]{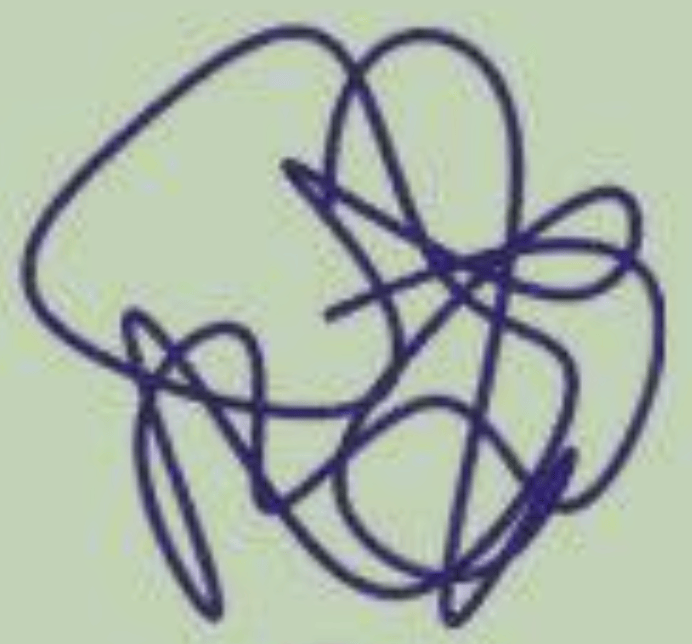}}}

\newcommand{\squiggleO}{\raisebox{-0.35\height}{\includegraphics[height=2em]{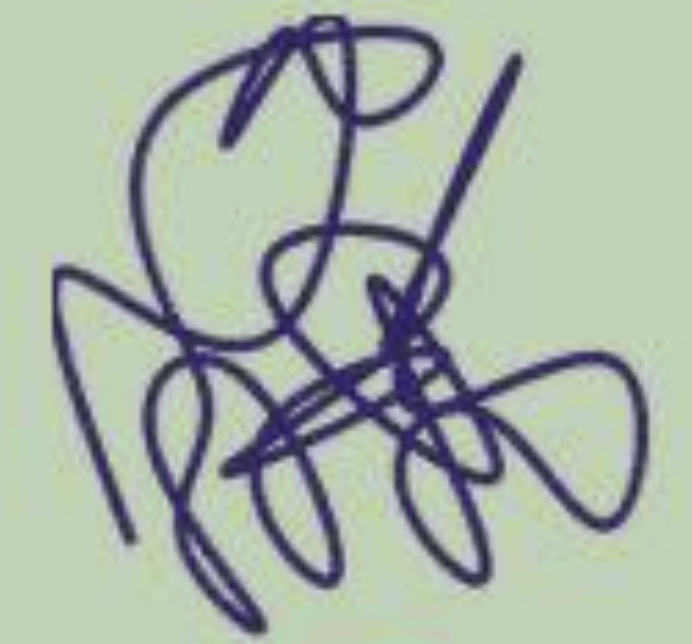}}}

\definecolor{darkblue}{rgb}{0, 0, 0.5}
\definecolor{mynavyblue}{RGB}{0, 32, 96}
\definecolor{mydarkorange}{RGB}{180, 60, 0}

\hypersetup{colorlinks=true, citecolor=darkblue, linkcolor=darkblue, urlcolor=darkblue}

\title{VLMs Need Words: Vision Language Models Ignore Visual Detail In Favor of Semantic Anchors}

\author{Haz Sameen Shahgir$^{1}$, Xiaofu Chen$^{2}$, Yu Fu$^{1}$, Erfan Shayegani $^{1}$\\ \textbf{Nael Abu-Ghazaleh$^1$}, \textbf{Yova Kementchedjhieva$^2$}, \textbf{Yue Dong$^1$}\\ [5pt]
University of California, Riverside $^1$, MBZUAI$^2$\\ [5pt]
\texttt{hshah057@ucr.edu}, \texttt{yued@ucr.edu}
}
\begin{document}

\ifcolmsubmission
\linenumbers
\fi

\maketitle

\input{sections/1_abstract}
\input{sections/2_introduction}
\input{sections/4_knowledge_and_gap}
\input{sections/5_teaching_names}
\input{sections/6_are_VLMs_limited}
\input{sections/3_related_work}

\input{sections/7_conclusion}

\bibliography{colm2026_conference}
\bibliographystyle{colm2026_conference}

\appendix
\input{sections/8_appendix}

\end{document}

%% file: sections/1_abstract.tex
\begin{abstract}

Vision-language models (VLMs) have achieved impressive performance across a wide range of multimodal tasks. However, they often fail on tasks that require fine-grained visual perception, even when the required information is still present in their internal representations. Prior work has attributed this ``hidden-in-plain-sight'' gap to the language model, but the cause remains unexplained. In this work, we demonstrate that this gap arises from the language model's lack of semantic labels for fine-grained visual details: when visual entities can be mapped to known concepts, VLMs bypass visual comparison and reason through language; when they cannot, VLMs resort to brittle and hallucinated descriptions. We verify this across semantic correspondence, synthetic shape matching, and face matching, and find that VLMs perform much better when the relevant entities are nameable than when they are unnamable. Mechanistically, Logit Lens analysis confirms that VLMs explicitly recover semantic labels for nameable entities and surface more unique tokens compared to unnameable entities. Furthermore, we show that this limitation can be addressed: teaching completely arbitrary names for unknown entities improves performance. More importantly, task-specific finetuning yields even stronger generalization without relying on language priors, i.e. through real visual perception. Our findings suggest that current VLM failures on visual tasks reflect a learned shortcut rather than a fundamental limitation of multimodal reasoning\footnote{Code and data are publicly available at \href{https://github.com/Patchwork53/VLMs-Need-Words-COLM2026}{github.com/Patchwork53/VLMs-Need-Words-COLM2026}}.

\end{abstract}

%% file: sections/2_introduction.tex
\section{Introduction}
Vision Language Models (VLMs) have achieved impressive performance across a wide range of multimodal tasks, from visual question answering and image captioning \citep{openai2024gpt4technicalreport} to document understanding and visual grounding \citep{bai2023qwenvl}. These models are pre-trained on billions of image-text pairs and further finetuned on curated instruction-following datasets, with the implicit promise that this pipeline produces general-purpose multimodal reasoners capable of handling novel multimodal problems out of the box.

Yet this promise remains unfulfilled. VLMs consistently fail on tasks that demand fine-grained visual perception. They struggle with chart and diagram understanding \citep{mathverse}, misinterpret optical illusions that humans resolve effortlessly \citep{illusionvqa}, falter on abstract visual reasoning \citep{arc-agi}, and most strikingly, cannot solve basic visual discrimination tasks that three-year-old children handle with ease \citep{babyvision}. These failures are not domain- or model-specific; they emerge whenever models must reason from pixels instead of matching patterns. These failures are often mitigated through task-specific supervision, but such improvements do not explain why VLMs pretrained on web-scale data fail to generalize in the first place. We argue that the failure is rooted in the standard pretraining-and-SFT pipeline itself, which encourages models to rely on semantic shortcuts rather than learning transferable visual skills.


Recently, \citet{fu2025hiddenplainsightvlms} and \citet{liu2025visual} have shown that VLMs' internal representations retain enough visual information to solve many of these challenging tasks, even when the model's verbal output fails. This ``hidden-in-plain-sight'' gap was identified as a shortcoming of the language model (LM) backbone, but its underlying mechanism was left unexplored. At the same time, other work has documented a systematic over-reliance on language priors: VLMs underperform on tasks that resist mediation through language \citep{fu2024blink} and exhibit biases inherited from their LM backbone \citep{vlmsarebiased}.

In this work, we connect and explain these observations. We hypothesize that VLMs short-circuit visual reasoning by mapping visual entities to discrete semantic labels in the language space. When a visual entity can be named, the model bypasses pixel-level comparison and transfers the task to the language space. When no crisp label exists, the model still attempts the same linguistic strategy, generating approximate, often hallucinated descriptions that actively degrade its reasoning. This explains both the hidden-in-plain-sight gap and why failures systematically cluster around novel or sub-semantic visual content.

We test this hypothesis through a visual correspondence task, where models must identify matching entities between two images, a capability that underlies real-world applications from medical differential diagnosis to temporal change detection in videos. Correspondence is a natural testbed because the degree to which target entities admit semantic labels varies naturally across settings: some keypoints on an object have well-known names (e.g., the \textit{pedal} of a bicycle; see Figure~\ref{fig:overview}, point D), while others do not (e.g., the joint between the head tube and top tube of a bicycle; see Figure~\ref{fig:overview}, point A). We extend this contrast by constructing synthetic tasks that pair common shapes ("star," "circle") against procedurally generated ones, and recognizable celebrity faces against AI-generated unknown faces.

Across all three settings, we find the same pattern. VLMs perform substantially better on nameable entities than on unnameable ones, even when representation probing confirms that internal features contain sufficient information in both cases. Chain-of-thought reasoning disproportionately benefits named entities, consistent with the model leveraging verbal reasoning to recover and match discrete labels. Logit Lens analysis~\citep{nostalgebraist2020logitlens} confirms the mechanism: for known entities, hidden states progressively resolve from semantically unrelated tokens through approximate descriptors to exact labels (e.g., "triangle" to "pointed" to "star"), while unknown entities remain semantically indiscernible.

We further show that teaching VLMs completely arbitrary names for novel shapes substantially closes the performance gap. Once a shape has a name, the model can short-circuit visual comparison altogether, matching shapes by their learned labels rather than their pixels. The Logit Lens signature of the name-trained models confirms this observation: downstream accuracy grows proportionally to the increase in semantic discernibility.

However, semantic labeling is sufficient but not necessary for closing this gap. We investigate what task-specific finetuning changes in VLMs. When finetuned directly on the correspondence task, VLMs outperform name-trained models while exhibiting lower semantic discernibility. This indicates that task-specific finetuning teaches a distinct mechanism --- direct visual comparison --- that does not route through language. Together, these findings suggest that current VLM failures on visual tasks reflect a learned shortcut induced by the standard pretraining-and-SFT pipeline, not a fundamental limitation of the architecture.

\begin{enumerate}
    \item We show, across semantic, shape, and face correspondence tasks, that VLM performance depends strongly on whether the target visual entity admits a semantic label, rather than on whether the internal representations contain sufficient information to solve the task.
    \item We provide mechanistic evidence via Logit Lens analysis that known entities become progressively more semantically explicit inside the language model, while unknown entities remain indiscernible, and that teaching arbitrary names to unknown entities closes this discernibility gap.
    \item We demonstrate that semantic labeling is sufficient but not necessary for closing the representation-to-output gap: teaching names enables a linguistic shortcut, while direct task finetuning elicits genuine visual reasoning rather than language-mediated semantic reasoning.
    
\end{enumerate}

%% file: sections/4_knowledge_and_gap.tex
\section{The Performance Gap Between Textual Response and Representation Space is Larger for Unknown Entities}
\label{sec:the_gap}

\begin{figure}
    \centering
    \includegraphics[width=1\linewidth]{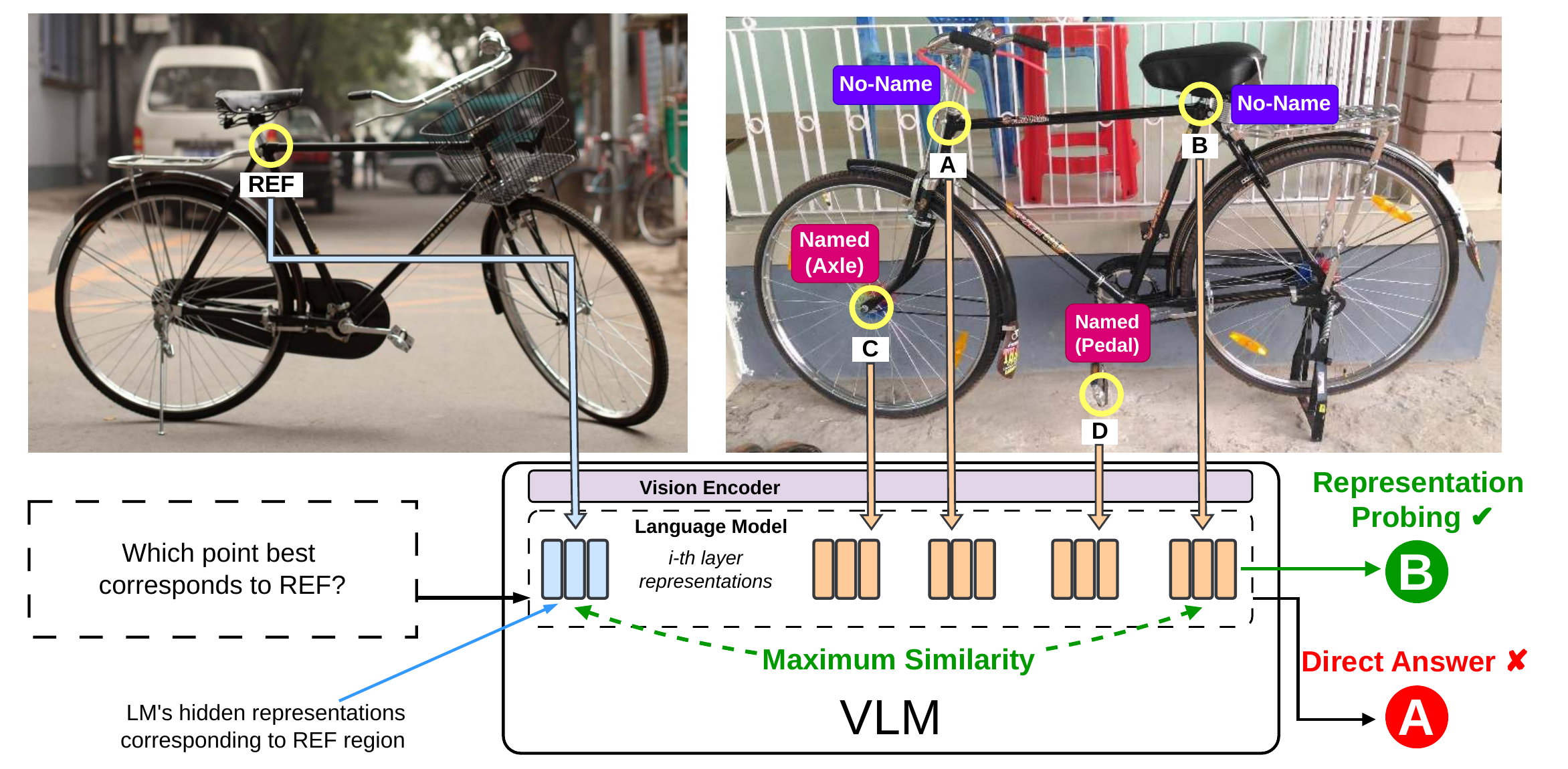}
    \caption{Overview of the correspondence task framed as multiple-choice VQA. The VLM must match the reference point ``REF'' in the first image to one of four candidate regions (A, B, C, D) in the second image. Regions can either be \textcolor{purple}{Named} or have \textcolor{violet}{No-Name}. We evaluate VLMs in three setups: Direct answer, Chain-of-Thought, and Representation Probing.}
    \label{fig:overview}
    \vspace{-1em}
\end{figure}

In this section, we first study the role of semantic anchors by comparing VLM performance on semantic correspondence for named vs. unnamed entities in real images, then isolate this effect using synthetically generated 2D object and face correspondence tasks. Beyond fine-grained visual correspondence, we also study the effect of semantic anchors on Art Style Matching, where VLMs need to identify one of four image options that has the same art style as a target image. This task requires holistic understanding of  

\paragraph{Experimental Setup}
For all correspondence tasks in this section, we follow the methodology of \citet{fu2025hiddenplainsightvlms} and add Chain-of-Thought (CoT) evaluation to study the effect of textual reasoning on inherently vision-heavy tasks. We frame the problem as a multiple-choice visual question answering (MC-VQA) task. We evaluate VLMs in three setups: 1) MC-VQA with Direct Answer, 2) MC-VQA with CoT, and 3) Representation Probing as shown in Figure \ref{fig:overview}. For MC-VQA, we present a VLM with two images and prompt it to find a point in the second image that matches a reference point in the first image. We annotate the reference point with ``REF'' and different regions in the second image with the four options (``A'', ``B'', ``C'', and ``D''). For Direct, we prepend \textit{``The correct answer is (''} to the assistant response and directly generate the answer. This is the setup explored by \citet{fu2025hiddenplainsightvlms, liu2025visual}, which requires the VLM to complete the task entirely through non-verbal reasoning in a single forward pass. In our work, we also test Chain-of-Thought, which not only allows VLMs to reason through text but also allows them to attend to the visual tokens multiple times before generating the answer. We append \textit{``Think step by step before choosing an option.''} to the original prompt. We impose no output-format constraints on the VLM, thereby avoiding the diversion of model capacity from the correspondence task to format compliance. Instead, we use Qwen3-8B to extract the selected answer option from the VLM’s free-form response. We use the default sampling parameters for all VLMs. In the Representation Probing setup, we define a square region of $30\times30$ pixels for each point of interest and extract the corresponding visual tokens. For each layer, we compute the similarity between the hidden representations of the ``REF'' point in the first image and those of the four options in the second image. A sample is considered correct if the ground-truth option has the highest similarity. Since each region maps to multiple tokens, we use the MaxSim operator \citep{maxsim} instead of cosine similarity, which requires a lossy pooling step.

\subsection{Testing Nameable and Unnameable Key Points in Semantic Correspondence}

\citet{fu2025hiddenplainsightvlms} and \citet{liu2025visual} have demonstrated that using VLMs' internal representations consistently outperforms their textual output on semantic correspondence on the  SPairs71k \citep{min2019spair} dataset. SPair-71k consists of a total of 70,958 image pairs from PASCAL~3D+ \citep{pascal3d} and PASCAL~VOC~2012 \citep{pascalvoc} spanning 18 object categories encompassing vehicles (e.g., ``aeroplane''), ordinary objects (e.g., ``bottle''), and animals (e.g., ``sheep'').

We isolate the effect of nameability by benchmarking VLMs on named and unnamed splits of the SPair-71k test set images. SPair-71k provides consistent keypoint annotations across all images within a category. We manually categorize each keypoint as either \textit{Named} or \textit{No-Name}: keypoints corresponding to well-known parts (e.g., pedal, seat, handlebar) are \textit{Named}, while keypoints at ambiguous locations (e.g., the junction between the handlebars and the stem) are \textit{No-Name}. Each task is assigned to the split corresponding to its reference keypoint, resulting in 11,362 Named and 11,792 No-Name samples.

\begin{table}[h]
\centering
\resizebox{0.85\linewidth}{!}{%
\begin{tabular}{lllccc|cc}
\toprule
\textbf{Model} & \textbf{Size} & \textbf{Subset}
  & \makecell{\textbf{Direct}\\\textbf{(D)}}
  & \makecell{\textbf{CoT}\\\textbf{(C)}}
  & \makecell{\textbf{C$-$D}\\\textbf{$\Delta$}}
  & \makecell{\textbf{Rep. Probe}\\\textbf{(R)}}
  & \makecell{\textbf{R$-$max(D,C)}\\\textbf{$\Delta$}} \\
\midrule
\multirow{6}{*}{Qwen3VL}
& \multirow{2}{*}{2B} & Named   & 36.6 & 57.4 & \textbf{20.8} & 68.7 & 11.3 \\
&                     & No-Name & 32.4 & 42.2 & 9.8           & 60.2 & \textbf{18.0} \\
\cmidrule{2-8}
& \multirow{2}{*}{4B} & Named   & 52.5 & 62.9 & \textbf{10.4} & 68.4 & 5.5 \\
&                     & No-Name & 37.2 & 46.7 & 9.5           & 62.1 & \textbf{15.4} \\
\cmidrule{2-8}
& \multirow{2}{*}{8B} & Named   & 53.8 & 65.3 & \textbf{11.5} & 68.6 & 3.3 \\
&                     & No-Name & 40.3 & 47.9 & 7.6           & 61.6 & \textbf{13.7} \\
\midrule
\multirow{6}{*}{InternVL3.5}
& \multirow{2}{*}{2B}  & Named   & 28.1 & 34.0 & \textbf{5.9}  & 56.8 & \textbf{22.8} \\
&                      & No-Name & 24.8 & 30.0 & 5.2           & 47.4 & 17.4 \\
\cmidrule{2-8}
& \multirow{2}{*}{8B}  & Named   & 34.3 & 46.7 & \textbf{12.4} & 56.7 & 10.0 \\
&                      & No-Name & 28.2 & 34.4 & 6.2           & 51.1 & \textbf{16.7} \\
\cmidrule{2-8}
& \multirow{2}{*}{14B} & Named   & 31.9 & 43.4 & \textbf{11.5} & 56.4 & 13.0 \\
&                      & No-Name & 28.4 & 36.5 & 8.1           & 53.8 & \textbf{17.3} \\
\bottomrule
\end{tabular}
}

\caption{Semantic correspondence results on SPairs71k. R denotes the representation-probing accuracy from the \textbf{best-performing layer} of the language model. We report layer-wise performance in Appendix Fig. \ref{fig:rep_probe_layer_sem_corr}. The R$-$max(D, C) $\Delta$ column shows the gap between Representation Probing and the best textual output.}
\label{tab:sem_corr}

\end{table}

Table \ref{tab:sem_corr} shows results across two VLM families on the Named and No-Name subsets of SPairs71k. First, the accuracy of VLMs on the No-Name subset is lower than on the Named subset for all model sizes, across both model families, and in both the Direct and CoT setups. This shows that VLMs struggle with the correspondence task when the reference point lacks a definitive semantic label. Second, we find that the gap between Representation Probing and best textual strategy (R$-$max(D, C) $\Delta$) is consistently larger for the No-Name subset: Qwen3VL-8B achieves an R$-$max(D, C) gap of just 3.3\% on Named keypoints, nearly closing the gap when semantic anchors are available, but 13.7\% on No-Name, while unnamed entities remain bottlenecked. The pattern holds for InternVL3.5 at 8B and 14B. In contrast, InternVL3.5-2B appears to lack the baseline capability required for the task, achieving only 28.1\% accuracy even on the Named subset, barely above the 25\% random-chance baseline. We report the performance of Gemma-3-4B and 12B in Appendix \ref{app:Gemma-3-Sem-Corr}.

Chain-of-thought reasoning disproportionately benefits named entities: Qwen3VL-2B gains $+20.8\%$ from CoT on Named keypoints but only $+9.8\%$ on No-Name. Inspecting the CoT chain (See Appendix Figure \ref{fig:sem_corr_cot_example}), we find that VLMs explicitly generate the names of the points when available, effectively converting the visual correspondence task to a verbal string-matching task. This CoT advantage narrows with scale (Qwen3VL-8B: $+11.5\%$ Named vs.\ $+7.6\%$ No-Name), consistent with larger models performing more of this reasoning internally. Representation Probing performance, by contrast, remains stable across model sizes within each family (e.g., 68.4--68.7\% Named and 60.2--62.1\% No-Name for Qwen3VL). Since models within a family share similar vision encoders, this stability indicates that the LLM layers do not destroy visual information; rather, larger LMs are more effective at surfacing it through text.

\subsection{Confirming the Effect of Semantic Anchors on Shape and Face Correspondence}

\begin{figure}[h]
    \centering   
    \begin{subfigure}[b]{0.23\textwidth}
        \centering
        \includegraphics[width=\textwidth]{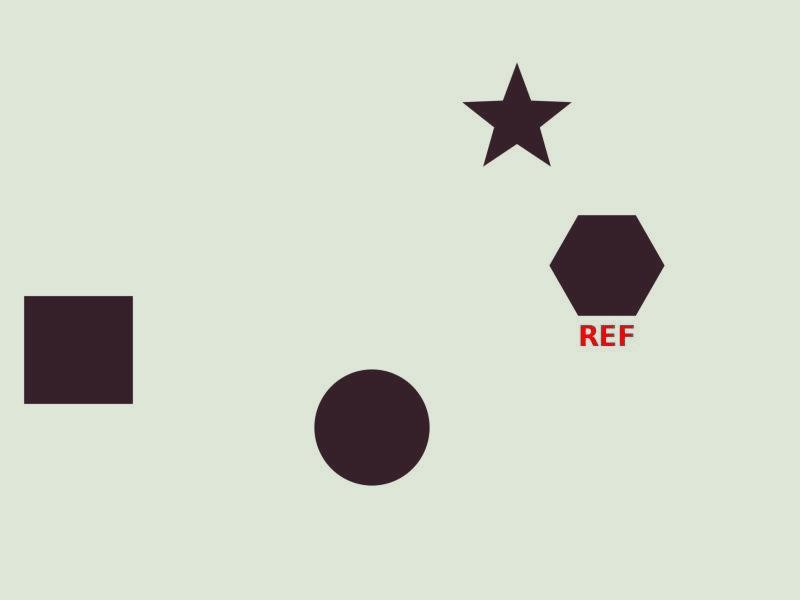}
        \caption{Known Ref.}
    \end{subfigure}
    \hfill
    \begin{subfigure}[b]{0.23\textwidth}
        \centering
        \includegraphics[width=\textwidth]{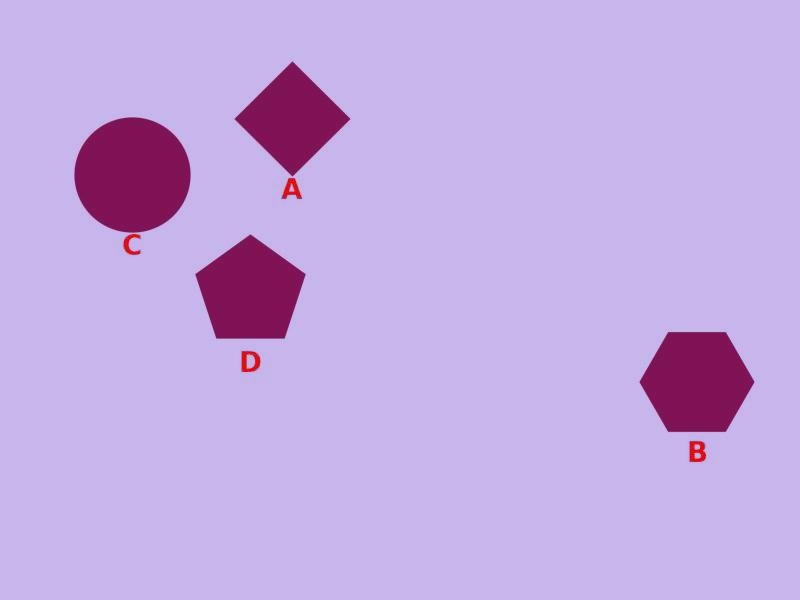}
        \caption{Known Tgt.}
    \end{subfigure}
    \hfill
    \begin{subfigure}[b]{0.23\textwidth}
        \centering
        \includegraphics[width=\textwidth]{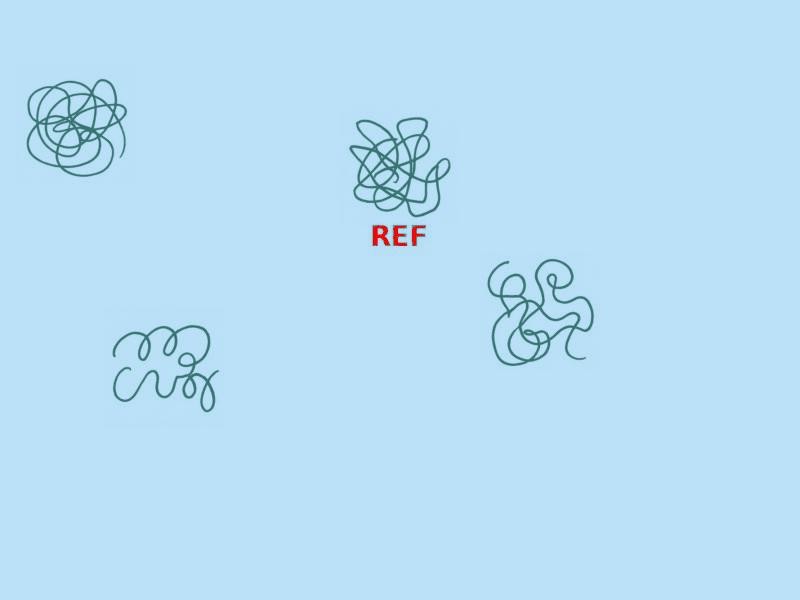}
        \caption{Unknown Ref. }
        \label{fig:shape_corr_c}
    \end{subfigure}
    \hfill
    \begin{subfigure}[b]{0.23\textwidth}
        \centering
        \includegraphics[width=\textwidth]{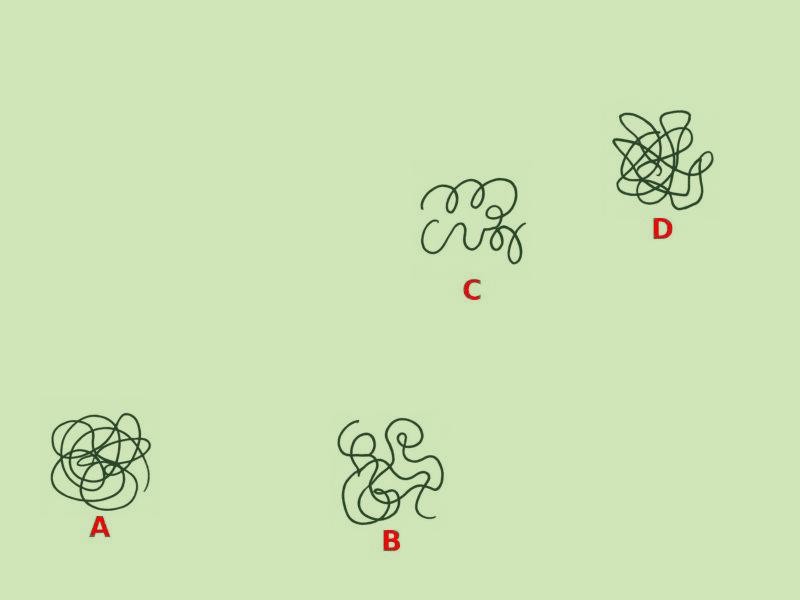}
        \caption{Unknown Tgt.}
        \label{fig:shape_corr_d}
    \end{subfigure}
    \caption{Example of our 2D shape correspondence task. Face correspondence example is presented in Appendix Fig. \ref{fig:face_corr_task}.}
    \label{fig:shape_corr}

    \vspace{-1em}
\end{figure}

The SPairs71k Semantic Correspondence task uses real-world images of entities subject to factors such as occlusion, varying resolution, varying entity size relative to the image, and visual artifacts. Since VLMs operate by converting non-overlapping image patches into tokens, the effects of these factors are complex and hard to control.
To study the effect of semantic anchors in isolation, we design the 2D object correspondence and face correspondence tasks. This synthetic setup allows us to control resolution and entity size, and ensures there is no partial occlusion or visual artifacts. Figure \ref{fig:shape_corr} shows an example of the 2D object correspondence task. To construct the dataset, we use 10 common shapes ("square", "circle", "star", etc.) as known 2D objects and generate unknown 2D shapes ("squiggles") using a randomized spline-generation algorithm whose complexity can be controlled. Squiggles are generated by iteratively sampling $n$ anchor points. A cubic spline is then fit through the resulting anchors and rendered with $4\times$ supersampling for anti-aliasing.

For the face correspondence task, we first collect synthetic faces from FluxSynID \citep{ismayilov2025fluxsynid} and a single photograph of well-known celebrities. We deliberately restrict the dataset to a single demographic group so that models cannot solve the task using coarse cues such as ethnicity or gender. We use East Asian male identities because our pilot experiments found that the evaluated VLMs recognized celebrities from this group more consistently than celebrities from other groups, such as African male celebrities. We then prompt \texttt{Nano-Banana-2} \citep{nano_banana} with one example image to generate 4 images per person in different poses, lighting conditions, and environments.  Because both the known and unknown celebrity images are generated through the same \texttt{Nano-Banana-2} pipeline, any model-specific visual artifacts are controlled for. We verify that the VLMs recognize each celebrity via knowledge probing and construct VLM-specific splits of the known celebrity dataset accordingly. The probing methodology and exact splits are reported in Appendix \ref{app:faces_dataset}. Our probing experiments reveal that InternVL3.5 is unable to recognize most celebrities. We therefore evaluate the Gemma3 model family to ensure our findings generalize across VLM architectures. We discuss InternVL3.5's inability to identify celebrities and its performance on Shape Correspondence in Appendix \ref{app:internvl_appendix}.

\begin{table}[h]
\centering
\small
\setlength{\tabcolsep}{4pt}
\renewcommand{\arraystretch}{1.15}
\resizebox{\linewidth}{!}{%
\begin{tabular}{@{} l c c c c c c c | c c c c c @{}}
\toprule
\multicolumn{2}{c}{\textbf{Model}} & \textbf{Known}
  & \multicolumn{5}{c|}{\textbf{Faces}}
  & \multicolumn{5}{c@{}}{\textbf{2D Shapes}} \\
\cmidrule(lr){4-8} \cmidrule(l){9-13}
 & & 
  & \makecell{\textbf{Direct}\\\textbf{(D)}}
  & \makecell{\textbf{CoT}\\\textbf{(C)}}
  & \makecell{\textbf{C$-$D}\\\textbf{$\Delta$}}
  & \makecell{\textbf{Rep. Probe}\\\textbf{(R)}}
  & \makecell{\textbf{R$-$max(D,C)}\\\textbf{$\Delta$}}
  & \makecell{\textbf{Direct}\\\textbf{(D)}}
  & \makecell{\textbf{CoT}\\\textbf{(C)}}
  & \makecell{\textbf{C$-$D}\\\textbf{$\Delta$}}
  & \makecell{\textbf{Rep. Probe}\\\textbf{(R)}}
  & \makecell{\textbf{R$-$max(D,C)}\\\textbf{$\Delta$}} \\
\midrule

\multirow{6}{*}{\rotatebox[origin=c]{90}{Qwen3VL}}
  & \multirow{2}{*}{2B}  & \checkmark
    & 77.1 & 59.2 & $-$17.9 & 97.2 & 20.1
    & 54.1 & 97.3 & 43.2 & 100  & 2.7 \\
  &                      & $\times$  
    & 41.1 & 37.4 & $-$3.7  & 70.8 & \textbf{29.7}
    & 29.0 & 27.3 & $-$1.7 & 74.2 & \textbf{45.2} \\
  \cmidrule(lr){2-13}
  & \multirow{2}{*}{4B}  & \checkmark
    & 84.2 & 83.2 & $-$1.0 & 93.2 & 9.0
    & 93.5 & 99.4 & 5.9 & 100 & 0.6 \\
  &                      & $\times$  
    & 56.9 & 55.7 & $-$1.2 & 73.4 & \textbf{16.5}
    & 48.4 & 40.0 & $-$8.4 & 91.7 & \textbf{43.3} \\
  \cmidrule(lr){2-13}
  & \multirow{2}{*}{8B}  & \checkmark
    & 83.9 & 85.2 & 1.3 & 92.8 & 7.6
    & 99.7 & 99.9 & 0.2 & 100 & 0.1 \\
  &                      & $\times$  
    & 65.7 & 63.8 & $-$1.9 & 74.1 & \textbf{8.4}
    & 57.1 & 37.7 & $-$19.4 & 86.1 & \textbf{29.0} \\
\midrule

\multirow{4}{*}{\rotatebox[origin=c]{90}{Gemma3}}
  & \multirow{2}{*}{4B}  & \checkmark
    & 49.8 & 52.7 & 2.9 & 61.2 & 8.5
    & 50.6 & 70.5 & 19.9 & 98.9 & 28.4 \\
  &                      & $\times$  
    & 32.4 & 31.9 & $-$0.5 & 42.9 & \textbf{10.5}
    & 30.5 & 32.3 & 1.8 & 91.7 & \textbf{59.4} \\
  \cmidrule(lr){2-13}
  & \multirow{2}{*}{12B} & \checkmark
    & 50.4 & 48.6 & $-$1.8 & 51.8 & 1.4
    & 72.7 & 91.8 & 19.1 & 98.5 & 6.7 \\
  &                      & $\times$  
    & 36.5 & 34.2 & $-$2.3 & 42.1 & \textbf{5.6}
    & 40.2 & 42.9 & 2.7 & 89.3 & \textbf{46.4} \\
\bottomrule
\end{tabular}%
}

\caption{Comparison of Direct, Chain-of-Thought, and Representational Probe accuracy across tasks, models, and subsets. R denotes the representation-probing accuracy from the \textbf{best-performing layer} of the language model. We report layer-wise performance in Appendix \ref{app:rep_probe_layers}. $\Delta$ columns show the gain of CoT over Direct (C$-$D) and of the representational probe over the stronger of the two baselines (R$-$max(D,C)). Bold entries indicate that the probing substantially exceeds verbal output for unknown entities.}
\label{tab:shape_face_corr}
\vspace{-2em}
\end{table}

\paragraph{Results}

Table \ref{tab:shape_face_corr} shows that in this controlled synthetic setting, the gap between known and unknown entities observed in Section~\ref{sec:the_gap} is amplified. The gap between  Rep.\ Probing and the best textual baseline (R$-$max(D,C) $\Delta$) is consistently larger for unknown entities across all models on both tasks. On 2D shapes, the gap is particularly stark: Qwen3VL-2B achieves just 29.0\% Direct accuracy on unknown shapes versus 74.2\% via Rep.\ Probing, a 45.2 percentage point gap, while known shapes are saturated (100\% Rep.\ Probe, 97.3\% CoT).

Unlike the semantic correspondence results, CoT can be harmful on these tasks: several models show negative C$-$D deltas on unknown entities (e.g., Qwen3VL-8B drops 19.4 points with CoT on unknown shapes), suggesting that without a semantic anchor, verbal reasoning devolves into hallucinated descriptions that actively mislead (see Appendix~\ref{app:cot_examples}). As before, Representation Probing performance remains stable across model sizes within each family, while larger models narrow the textual output gap.

\paragraph{Beyond fine-grained correspondence: Art-Style Matching}
To test whether this effect extends to more holistic visual judgments, we additionally evaluate art-style matching \citep{fu2025hiddenplainsightvlms}, where models identify which target painting shares the reference painting's style. Art style depends on global properties such as texture, color usage, brushwork, and composition. We find that VLMs perform better on samples involving artists they can recognize. Recognizing the artist allows the model to avoid visually comparing the paintings' artistic styles. We discuss this experiment in Appendix~\ref{app:art_style}.

\subsection{Logit Lens Reveals VLMs Explicitly Recover Semantic Anchors}

We use Logit Lens \citep{nostalgebraist2020logitlens, nostalgebraist2021logitlensextensions} to directly decode the hidden representations of visual tokens inside the language model, confirming that VLMs explicitly align known visual tokens to semantic labels. We refer to the top-1 decoded token as the \emph{Logit Lens token} from here on. Since each entity is patched into multiple visual tokens, we approximate the semantic discernibility of two entities by measuring the difference between their Logit Lens token sets. Specifically, we compute the Jaccard Distance $ D_J(A, B \mid L) = 1 - \frac{|\,\mathrm{LL}_L(\mathcal{V}_A) \;\cap\; \mathrm{LL}_L(\mathcal{V}_B)\,|}{|\,\mathrm{LL}_L(\mathcal{V}_A) \;\cup\; \mathrm{LL}_L(\mathcal{V}_B)\,|} $
where $\mathrm{LL}_L(\mathcal{V}_X)$ denotes the set of Logit Lens tokens for entity $X$'s visual tokens at layer $L$.

Since each target image contains 4 entities, we compute the Jaccard Distance for all $\binom{4}{2}$ pairs per image and average over the entire dataset to obtain the \emph{Mean Jaccard Distance} for each layer. A \textbf{higher} Jaccard Distance implies that the visual tokens of two entities are \textbf{more differentiable} in semantic space and therefore easier for VLMs to tell apart. We would expect known shapes and faces to surface more unique Logit Lens tokens and therefore higher Jaccard Distance. We note that Jaccard Distance \textbf{underestimates} semantic discernibility for three reasons: (1) we only use the top-1 decoded Logit Lens token; (2) it does not account for synonyms; and (3) since the shapes/faces are naturally similar, they share some Logit Lens tokens such as ``polygon'', ``geometry'', ``eye'', ``nose'', names of colors for shapes, and ethnicity for faces. Furthermore, prior work has shown that Logit Lens itself underestimates the interpretability of visual tokens in early layers \citep{krojer2026latentlens}, since the unembedding matrix is trained to decode only the last layer. However, since our goal is to study the \emph{relative} difference between known and unknown entities rather than absolute interpretability performance, Logit Lens suffices.

\begin{figure}[h]
    \centering   
    \begin{subfigure}[t]{0.32\textwidth}
        \centering
        \includegraphics[width=\textwidth]{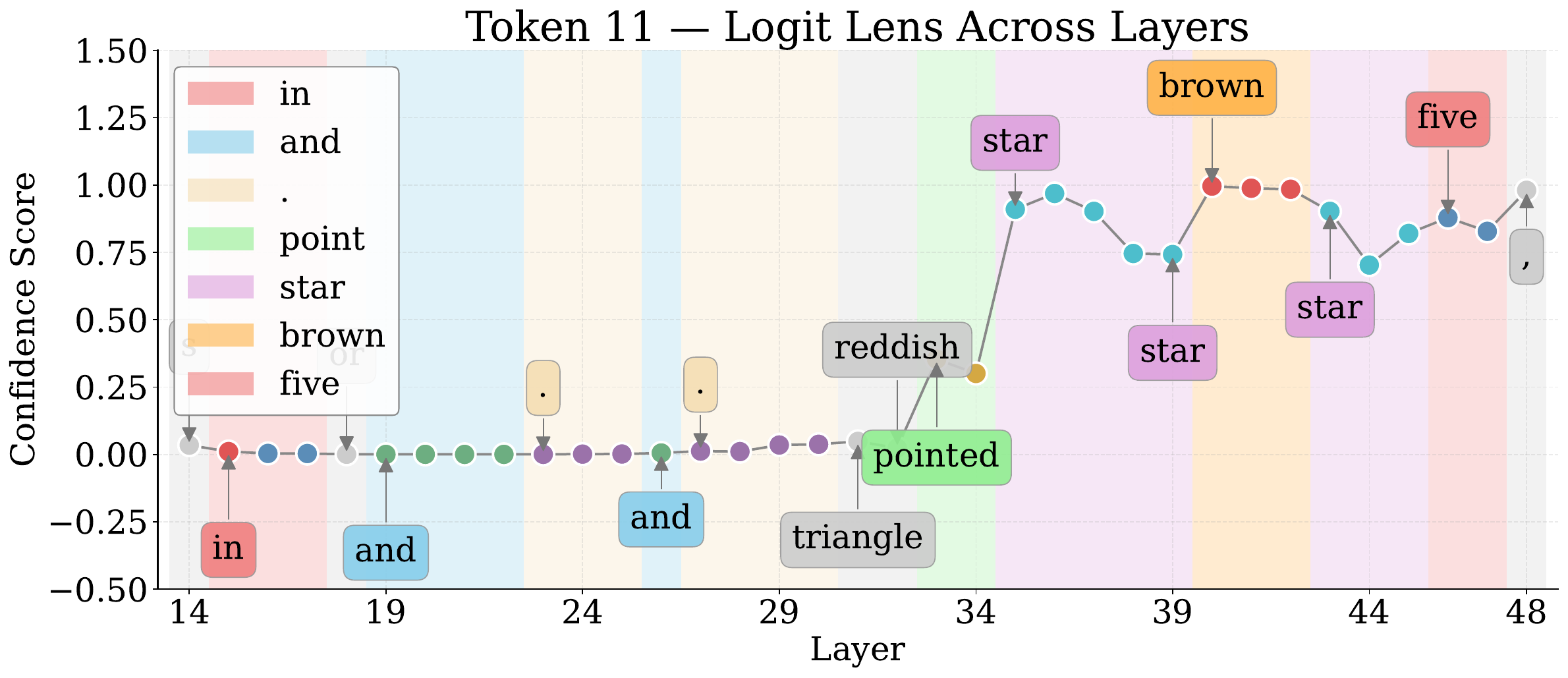}
        \caption{\scalebox{0.7}{Logit Lens for Known Shape (Star)}}
        \label{fig:gemma_logit_a}
    \end{subfigure}
    \hfill 
    \begin{subfigure}[t]{0.32\textwidth}
        \centering
        \includegraphics[width=\textwidth]{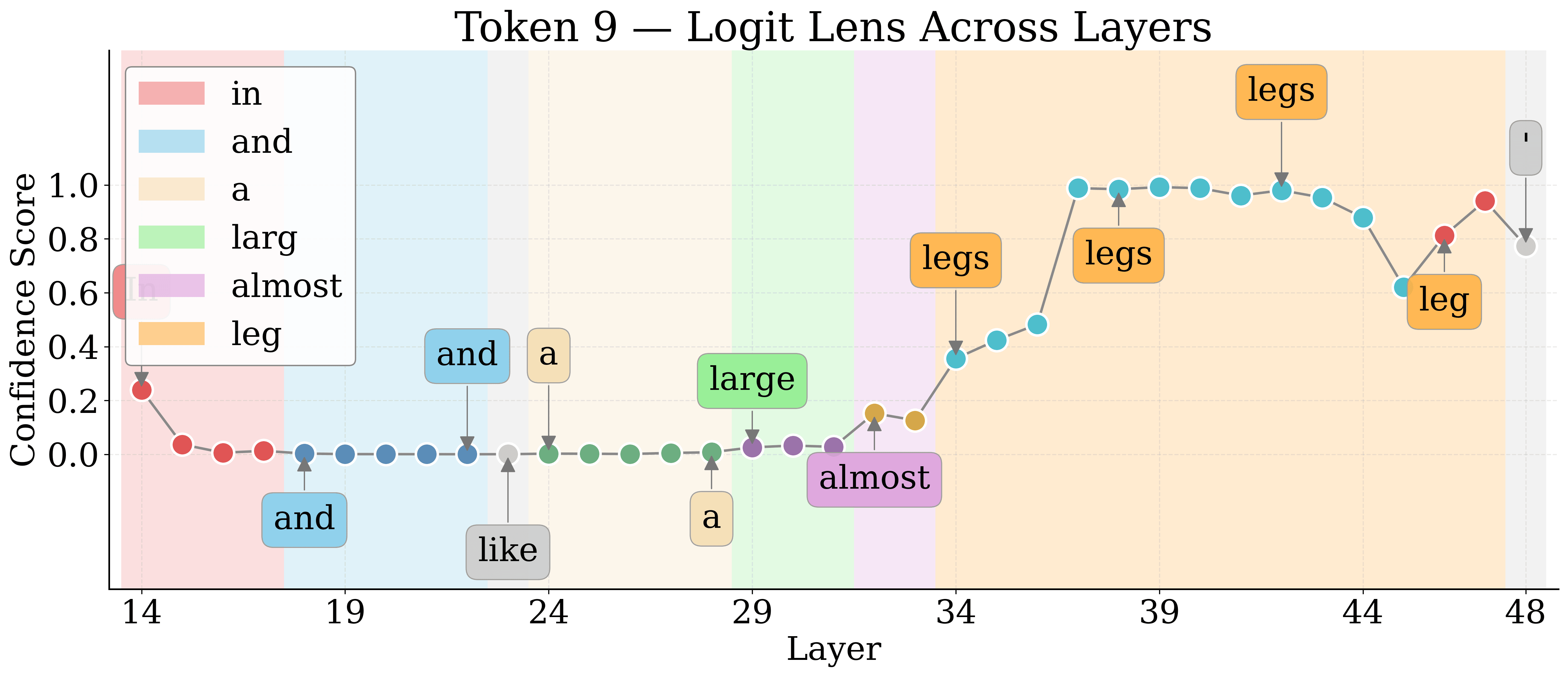}
        \caption{\scalebox{0.7}{Logit Lens for Unknown Shape \protect\squiggleZ}}
        \label{fig:gemma_logit_b}
    \end{subfigure}
    \hfill 
    \begin{subfigure}[t]{0.32\textwidth}
        \centering
        \includegraphics[width=\textwidth]{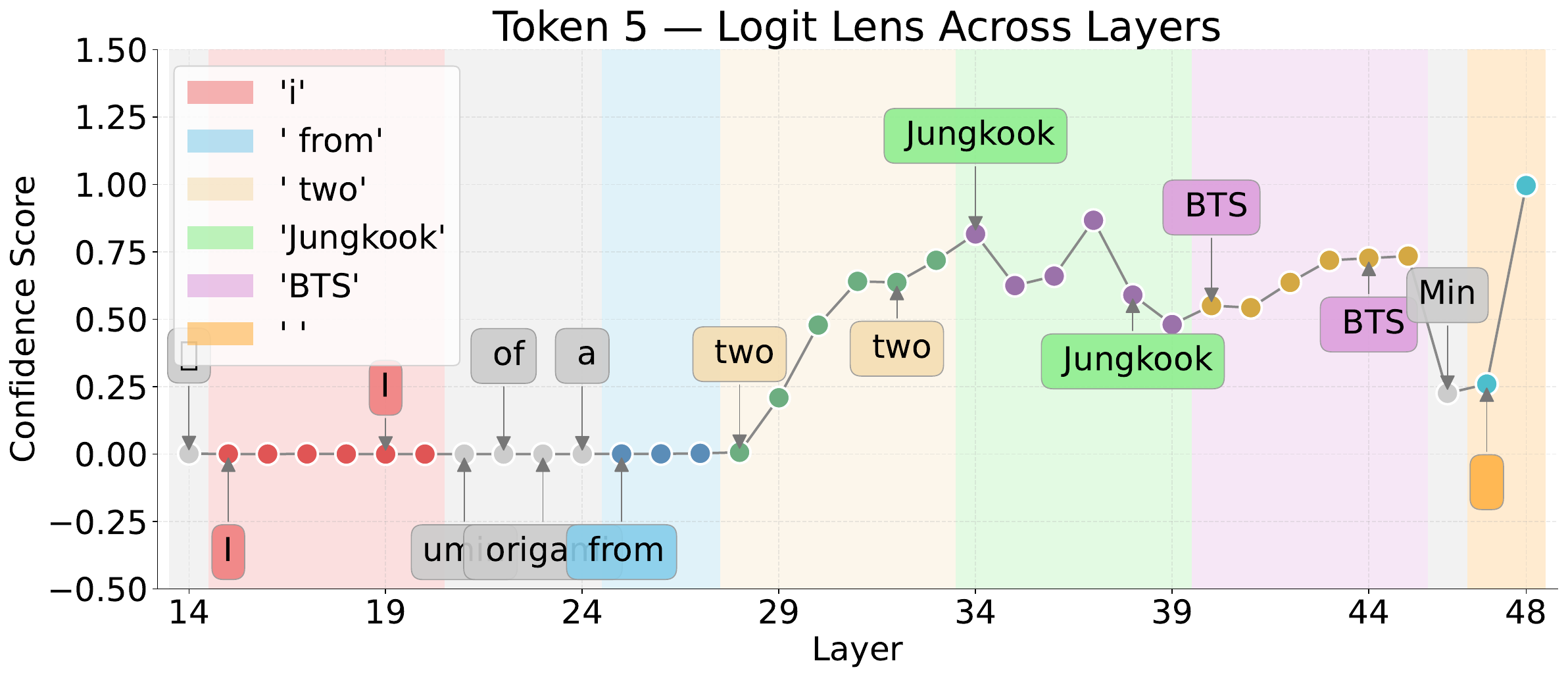}
        \caption{\scalebox{0.7}{Logit Lens for Known Face (Jungkook)}}
        \label{fig:gemma_logit_c}
    \end{subfigure}
    \begin{subfigure}[b]{0.24\textwidth}
        \centering
        \includegraphics[width=\textwidth]{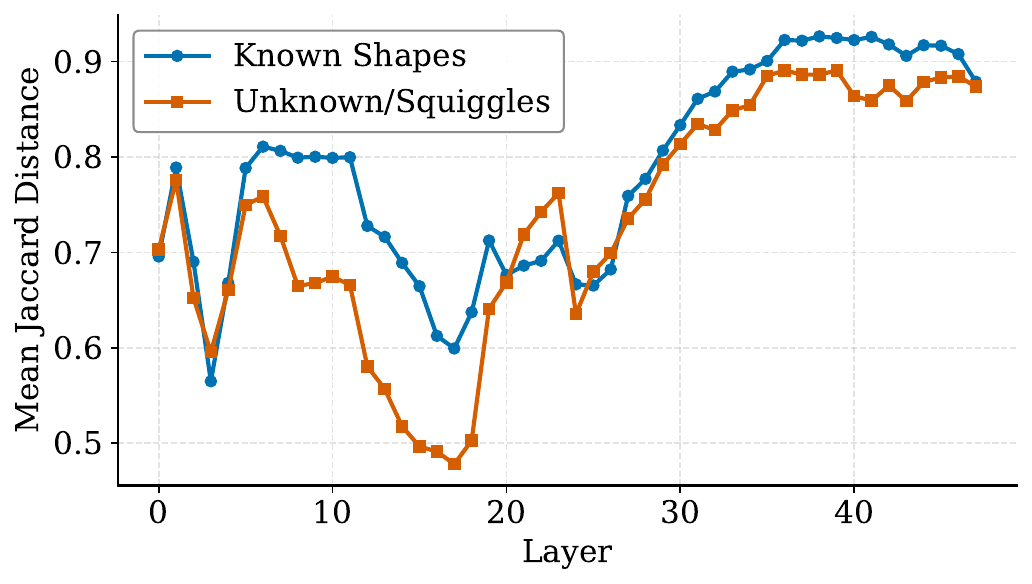}
        \caption{\scalebox{0.8}{Gemma - Shapes}}
        \label{fig:jaccard_d}
    \end{subfigure}
    \hfill
    \begin{subfigure}[b]{0.24\textwidth}
        \centering
        \includegraphics[width=\textwidth]{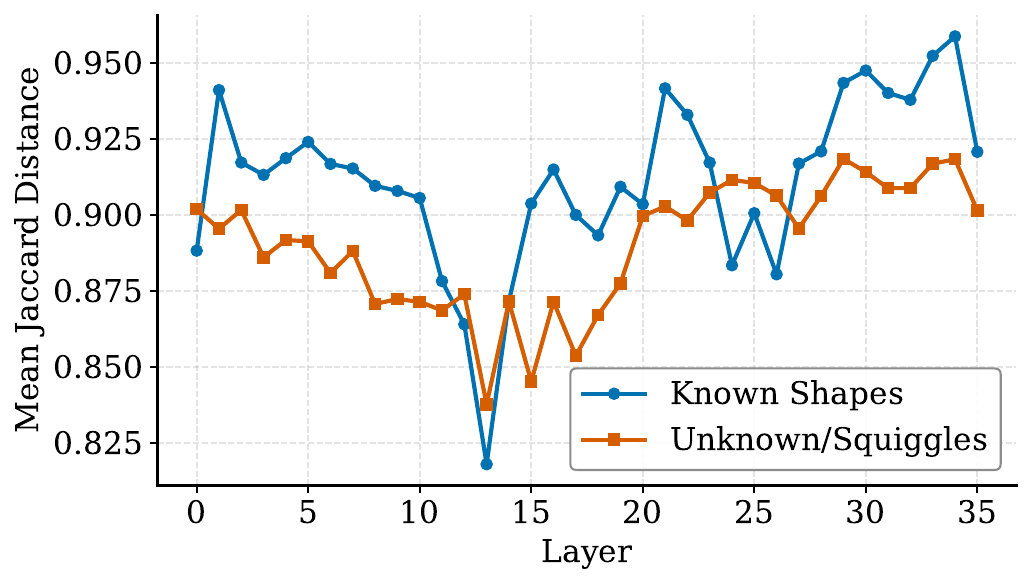}
        \caption{\scalebox{0.8}{Qwen - Shapes}}
        \label{fig:jaccard_e}
    \end{subfigure}
    \hfill
    \begin{subfigure}[b]{0.24\textwidth}
        \centering
        \includegraphics[width=\textwidth]{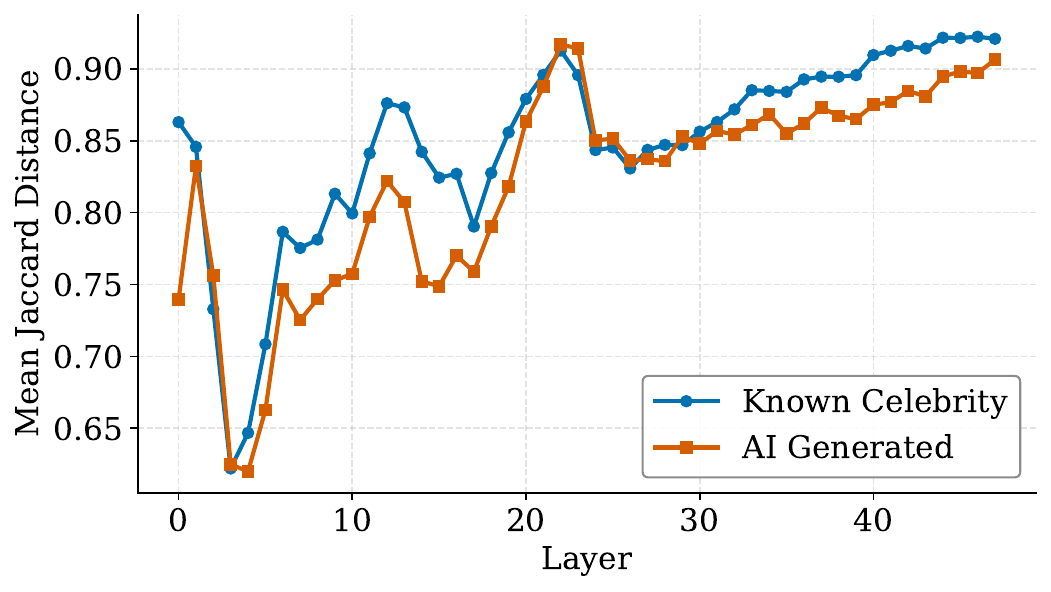}
        \caption{\scalebox{0.8}{Gemma - Faces}}
        \label{fig:jaccard_f}
    \end{subfigure}
    \hfill
    \begin{subfigure}[b]{0.24\textwidth}
        \centering
        \includegraphics[width=\textwidth]{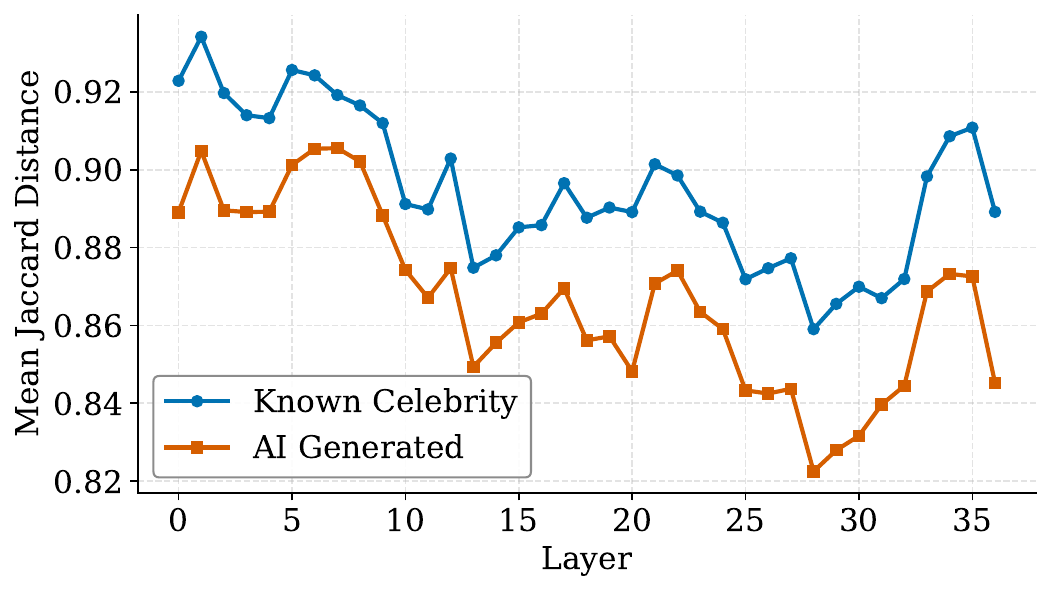}
        \caption{\scalebox{0.8}{Qwen - Faces}}
        \label{fig:jaccard_g}
    \end{subfigure}

    \caption{Logit Lens analysis. \textbf{Top row:} layerwise decoded tokens for a known shape (star) and a known face (Jungkook) in Gemma3-12B, showing the progression from semantically unrelated tokens to exact labels and encyclopedic associations. For the unknown shape, the Logit Lens probe surfaces a hallucinated token (``legs"). \textbf{Bottom row:} Mean Jaccard Distance across layers for shapes and faces in Gemma3-12B and Qwen3VL-8B. Known entities (\textcolor{blue}{blue}) yield consistently higher Mean Jaccard Distance than unknown entities (\textcolor{orange}{orange}), confirming greater semantic discernibility in the hidden representations.}

    \vspace{-1em}
\end{figure}

\paragraph{Results} Figures~\ref{fig:gemma_logit_a} and~\ref{fig:gemma_logit_c} visualize the Logit Lens trajectory of individual visual tokens in Gemma3-12B for two Known Entities: a 2D star shape and a photograph of the celebrity. The decoded token evolves through distinct phases as it passes through the LM transformer blocks. From Layer~0 to~$\sim$30, the Logit Lens token is semantically unrelated to the visual input (e.g., ``and,'' ``in,'' ``from'') and carries near-zero probability mass. Around Layer~$\sim$31--33, a transitional phase emerges, more pronounced for the shape, where approximate descriptors appear (``triangle,'' ``reddish,'' ``pointed'' for the star in Fig.~\ref{fig:gemma_logit_a}). By Layer~$\sim$34, the top-1 probability surges and the decoded token snaps to the correct semantic label: ``star'' and ``Jungkook'' in Figs.~\ref{fig:gemma_logit_a} and~\ref{fig:gemma_logit_b}, respectively. In the deepest layers, both tokens continue to evolve, accumulating higher-level associations: ``five'' for the star (reflecting the five-pointed shape) and ``BTS'' for the face (the celebrity's group). This layerwise progression, from noise, through approximate descriptors, to an exact semantic anchor, and finally to encyclopedic associations, demonstrates that VLMs explicitly recover discrete names for known visual entities within their intermediate representations. In contrast, Figure~\ref {fig:gemma_logit_b} shows that a visual token for an Unknown 2D shape spuriously maps to the ``legs" token. We show additional examples in Appendix~\ref{app:logit_lens_extra}.

Figures~\ref{fig:jaccard_d}--\ref{fig:jaccard_g} show that known shapes and faces have consistently higher Mean Jaccard Distance than unknown ones from Layer~4 onwards in both Gemma3-12B and Qwen3VL-8B. This confirms that Logit Lens produces more unique tokens for known entities and that the underlying hidden representations are more semantically discernible for the language model across both model families. Because semantically meaningful tokens emerge only after approximately Layer 30 in Gemma3-12B, the higher Mean Jaccard Distance observed in these later layers is particularly informative. Although the metric fluctuates considerably in earlier layers, it becomes consistently higher for known entities beyond Layer 30, indicating that their representations are more semantically distinguishable to the language model in the layers where meaningful visual semantics emerge.



%% file: sections/5_teaching_names.tex
\section{Teaching Arbitrary Names}
\label{sec:teaching_names}

In Section \ref{sec:the_gap}, we showed that VLMs' performance on the semantic, shape, and face correspondence tasks is greatly affected when the model lacks a semantic label for the reference entity, despite the internal representations encoding enough visual information. In this section, we investigate whether teaching VLMs an arbitrary name for previously unknown shapes can close this gap on the shape correspondence task. Once a shape has a name, the VLM can use its default strategy of short-circuiting visual comparison, instead matching shapes by their learned labels: \textit{``REF is John and C is John, so the answer is C.''}

\paragraph{Experimental Setup}
We finetune Qwen3VL-2B and Gemma3-4B to teach them arbitrary names for each squiggle shape. We present the image containing a shape and ask, "What is the name of this object?" and train the model to generate the name. To prevent overfitting, we train the VLMs on a varied set of single-image tasks and aggressively augment the images. All tasks are single images with at most two shapes per image, and are deliberately designed to be distinct from the original 2D shape correspondence task, which had two images with four shapes per image. This ensures the VLM is not inadvertently trained to compare shapes between images. Examples of the tasks can be found in Appendix \ref{app:teaching_names}.

We test three sets of names: ordinary object names that already map to a unique visual object (cup, anchor, feather, etc.), human names that are known to LLMs but lack a unique visual counterpart (John, Mary, Charles, etc.), and randomly generated 6-character strings that are completely unknown (``0QK2Z2'', ``5F1FT3'', ``OZ0W0M'', etc.).

\begin{figure}[h]
\centering
\begin{minipage}{0.50\textwidth}
    \centering
    \resizebox{\textwidth}{!}{%
    \begin{tabular}{llcc}
    \toprule
    Model & Eval. Strategy& Name Set& Accuracy \\
    \midrule
    \multirow{5}{*}{Qwen3VL-2B}
      & Rep. Probe        &          & 74.2 \\
    \cmidrule{2-4}
      & \multirow{4}{*}{VQA}         & Baseline & 29.0   \\
      &                              & Random   & 62.8 \\
      &                              & Human    & 70.2 \\
      &                              & Ordinary & 86.0   \\
    \midrule
    \multirow{5}{*}{Gemma3-4b}
      & Rep. Probe        &          & 91.7 \\
    \cmidrule{2-4}
      & \multirow{4}{*}{VQA}         & Baseline & 30.5 \\
      &                              & Random   & 65.1 \\
      &                              & Human    & 41.0   \\
      &                              & Ordinary & 50.5 \\
    \bottomrule
    \end{tabular}
    }
    \captionof{table}{\footnotesize Direct VQA accuracy on unknown shape correspondence after learning arbitrary names. Rep.\ Probe (pre-finetuning) shown as reference.}

    \label{tab:teaching_names}
\end{minipage}
\hfill
\begin{minipage}{0.48\textwidth}
    \centering
    \includegraphics[width=\textwidth]{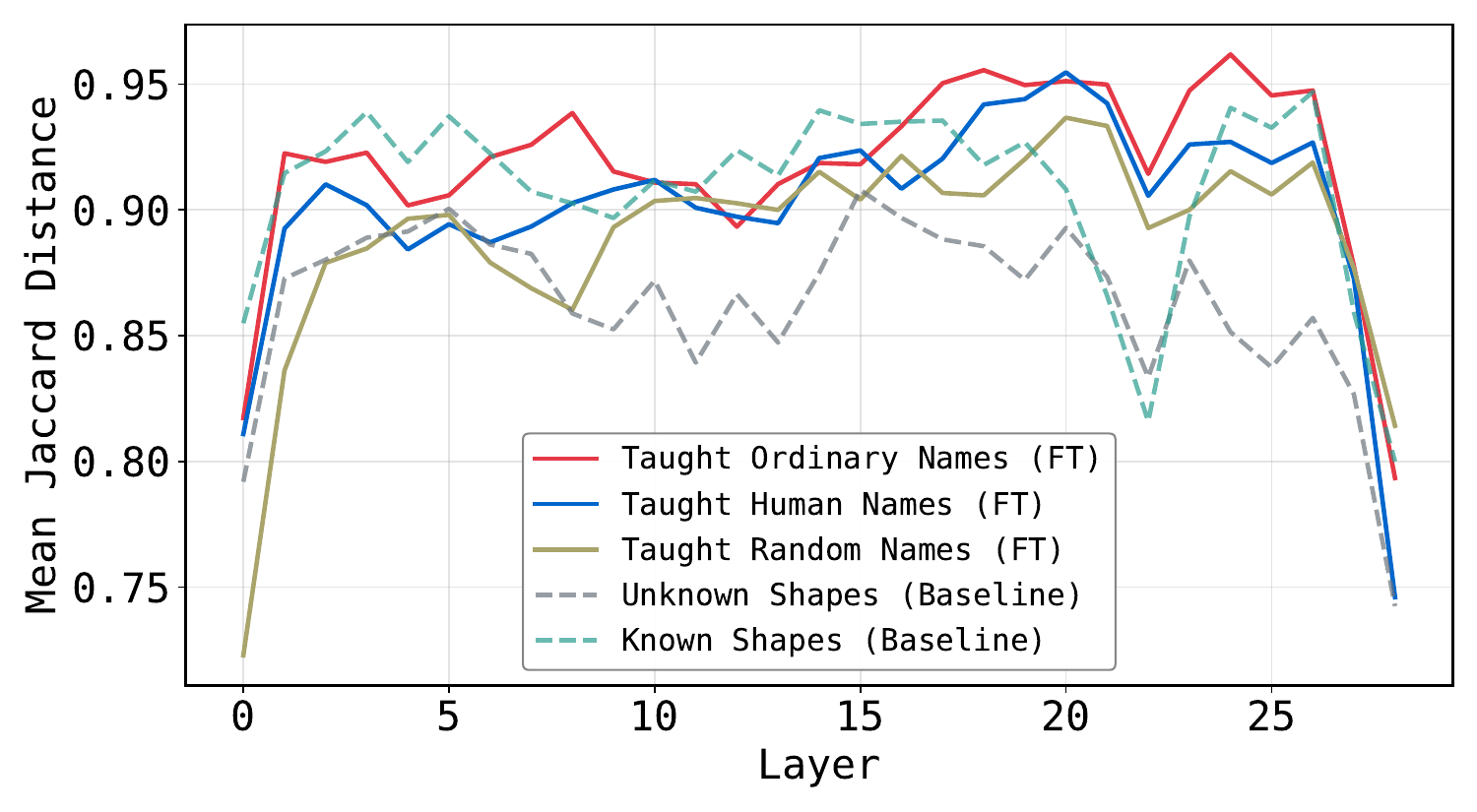}
    \captionof{figure}{\footnotesize Mean Jaccard Distance across layers for Qwen3VL-2B after learning arbitrary names. Finetuned models close the gap between the unknown and the known shapes baselines.}
    \label{fig:teaching_names_logit}
\end{minipage}
\end{figure}

\paragraph{Results}
Table \ref{tab:teaching_names} shows the results of teaching VLMs arbitrary names for each unknown squiggle object. All three name types consistently improve VQA accuracy over the baseline.\footnote{We show the representation probing after finetuning in Appendix Figure \ref{fig:probe_after_names}.} For Qwen3VL-2B, ordinary names reach 86.0\% (from 29.0\% baseline), surpassing even the pre-finetuning Representation Probe of 74.2\%. Human names reach 70.2\% and random names 62.8\%. Qwen3VL-2B benefits most from ordinary names (86.0\%), while Gemma3-4B benefits most from random names (65.1\%). The three name sets differ in average tokenization length (1 token for ordinary, ${\sim}$1.4 for human, ${\sim}$4.7 for random), which possibly interacts with how easily each model learns and uses them. Figure \ref{fig:teaching_names_logit} confirms the mechanism: Logit Lens reveals that finetuned Qwen models close the Mean Jaccard Distance gap between unknown and known shape baselines, and the ordering (Ordinary $>$ Human $>$ Random) matches the VQA accuracy ordering, confirming a positive correlation between semantic discernibility and downstream performance. Inspecting the Chain-of-Thought reasoning confirms that the VLM \textbf{short-circuits} pixel comparison using the learned semantic anchors (see Appendix~\ref{app:cot_examples}). Instead of referring to fine-grained visual details, the VLM simply reasons \textit{``REF is a brick and Choice D is a brick so the answer is D.''}.

%% file: sections/6_are_VLMs_limited.tex
\section{Is Semantic Alignment A Requirement or A Side Effect?}

In this section, we investigate whether semantic anchoring is a requirement for good VLM performance or a side effect of VLM pretraining. Note that the need for semantic anchors would limit VLMs to the granularity of existing vocabulary, which is undesirable.

To test this, we finetune VLMs on shape correspondence using one set of shapes and evaluate on entirely different shapes the model has never seen, including a different shape family (mazes). A model that has learned to compare visual details directly should generalize across shape families. A model that relies on internally assigned semantic labels has no labels for novel shapes and should fail.

\begin{figure}[h]
    \centering   
    \begin{subfigure}[c]{0.49\textwidth}
        \centering
        \includegraphics[width=\textwidth]{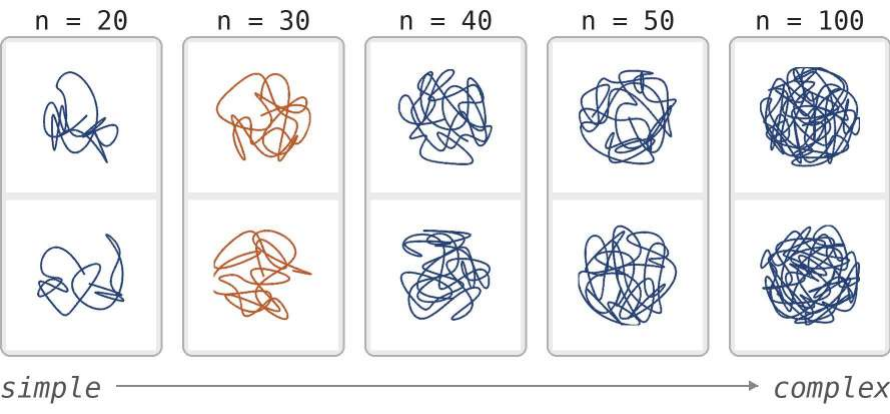}
        \caption{Generated Squiggles}
    \end{subfigure}
    \hfill
    \begin{subfigure}[c]{0.49\textwidth}
        \centering
        \includegraphics[width=\textwidth]{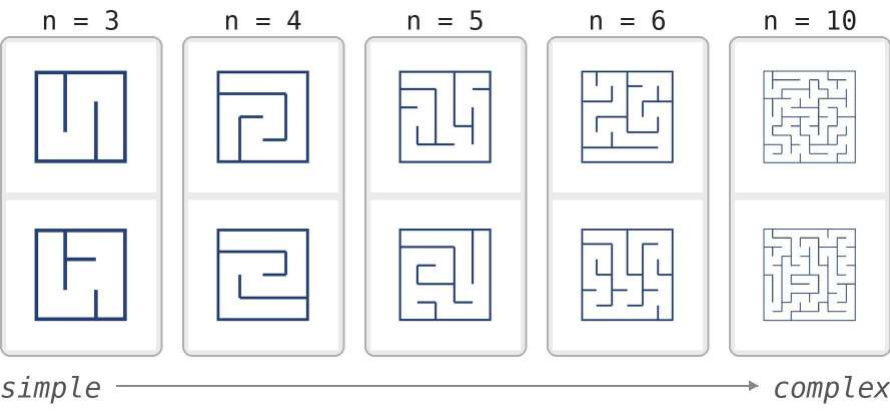}
        \caption{Generated Mazes}
    \end{subfigure}
    \caption{Set of procedurally-generated shapes tested. $n$ denotes the shape complexity variable: number of anchor points for squiggles and grid size for mazes. \textcolor{mydarkorange}{Orange} denotes the train dataset (Squiggles n=30) and \textcolor{blue}{blue} denotes the test dataset. }
    \vspace{-1em}
\end{figure}

\paragraph{Experimental Setup}
We finetune Qwen3VL-2B and Gemma3-4B on shape correspondence using squiggles with medium complexity ($n{=}30$ anchor points). Our training dataset consists of 1000 synthetically generated image pairs of the type shown in Figure \ref{fig:shape_corr_c} and \ref{fig:shape_corr_d}. We evaluate on held-out squiggles with both simpler and more complex configurations, procedurally generated mazes whose rectilinear grid structure bears no geometric resemblance to the shapes seen during training, and on semantic and face correspondence tasks that represent extremely out-of-distribution domains.


\begin{figure}[!h]
\centering
\begin{minipage}[b]{0.50\textwidth}
    \centering
    \resizebox{\textwidth}{!}{%
    \begin{tabular}{ll|ccc|ccc}
        \toprule
         & & & Squiggle& & & Maze& \\
        \midrule
        Model & Domain& n& Base& FT& n& Base& FT \\
              & Shift  & & Acc. & Acc. & & Acc. & Acc. \\
        \midrule
        \multirow{6}{*}{Qwen3VL-2B}& ID & 30 & 29 & 99.3 & & & \\
        \cmidrule{2-8}
        & \multirow{5}{*}{OOD} & 20  & 36.7 & 100   & 3 & 36.8 & 99.0 \\
        &                      & 30  & 28.8 & 98.7  & 4 & 34.2 & 99.4 \\
        &                      & 40  & 27.8 & 93.0  & 5 & 28.6 & 90.5 \\
        &                      & 50  & 25.9 & 76.1  & 6 & 25.9 & 82.2 \\
        &                      & 100 & 28.7 & 61.4  & 10& 24.0 & 62.8 \\
        \midrule
        \multirow{6}{*}{Gemma3-4b}& ID & 30 & 30.3 & 100 & & & \\
        \cmidrule{2-8}
        & \multirow{5}{*}{OOD} & 20  & 36.1 & 100   & 3 & 36.3 & 99.9 \\
        &                      & 30  & 31.9 & 98.7  & 4 & 37.3 & 99.1 \\
        &                      & 40  & 26.7 & 94.0  & 5 & 31.9 & 99.3 \\
        &                      & 50  & 27.2 & 82.7  & 6 & 28.6 & 97.1 \\
        &                      & 100 & 27.0 & 63.2  & 10& 28.0 & 81.1 \\
        \bottomrule
    \end{tabular}
    }
    \captionof{table}{\footnotesize Direct VQA accuracy on shape correspondence after finetuning on squiggles ($n{=}30$). Base Acc. is pre-finetuning. FT Acc. is after finetuning.}
    \label{tab:on_task}
\end{minipage}
\hfill
\begin{minipage}[b]{0.48\textwidth}
    \centering
    \includegraphics[width=\textwidth]{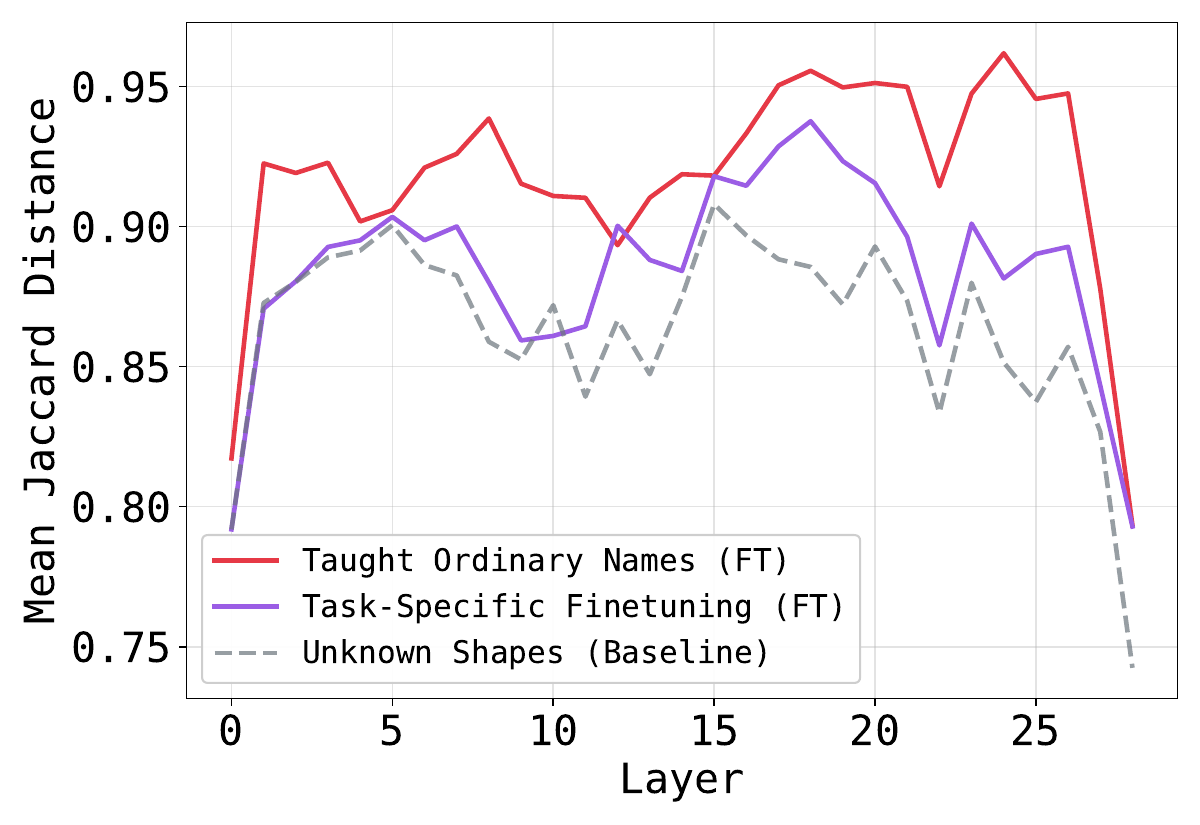}
    \captionof{figure}{\footnotesize Task-specific finetuning has lower Jaccard Distance than teaching ordinary names, despite higher VQA accuracy (98.7\% vs 86.0\%).}
    \label{fig:logits_after_ft}
\end{minipage}

\end{figure}

\paragraph{Results}

Table \ref{tab:on_task} shows that both Qwen3VL-2B and Gemma3-4B exhibit strong OOD generalization after finetuning, achieving near-perfect accuracy on held-out squiggles of similar or lower complexity, with graceful degradation as complexity increases. Critically, they also generalize to mazes despite being trained only on squiggles, with Gemma3-4B reaching 99.3\% on complex $n=5$ maze structures. This rules out the internal labeling hypothesis: VLMs generalize to unseen shapes, including an entirely different shape family, indicating they have learned visual comparison rather than memorizing specific shapes. 

We further find that finetuning on squiggles improves performance on face and semantic correspondence tasks, too, which are extremely out-of-distribution with respect to squiggles. Finetuned Qwen3VL-2B improves by $16\%$ on face correspondence and $10.7\%$ on semantic correspondence. Finetuned Gemma3-4B improves by $8.7\%$ and $5.8\%$ on face and semantic correspondence, respectively. The full results can be found in Appendix~\ref{app:face_sem_after_ft}.

Figure \ref{fig:logits_after_ft} reveals that task-specific finetuning has a lower Mean Jaccard Distance than teaching ordinary names (Section~\ref{sec:teaching_names}), but it achieves higher VQA accuracy (98.7\% vs 86.0\%). This suggests that task-specific finetuning teaches the VLM to leverage fine-grained visual information through a mechanism beyond semantic discernibility. Prior work has shown that task-specific finetuning improves VLM performance on visual correspondence tasks \citep{fu2025hiddenplainsightvlms, liu2025visual}, but the underlying mechanism was unclear. Our results suggest semantic anchoring is a sufficient but not necessary condition: VLMs can also learn to use visual information directly when the training task demands fine-grained visual comparison, and the two pathways (semantic anchoring in Section~\ref{sec:teaching_names} vs.\ direct visual comparison here) represent distinct mechanisms for closing the ``hidden-in-plain-sight'' gap.

%% file: sections/3_related_work.tex
\section{Related Work}
Recent work has shown that current vision-language models still struggle on tasks that require pixel-level or fine-grained visual reasoning. These limitations appear in chart and diagram understanding, optical illusions, and subtle visual discrimination, where success depends on preserving local evidence and comparing small visual differences rather than relying on coarse object semantics \citep{mathverse,illusionvqa,babyvision,vlmsubtlebench,anatomyvlm,glovlms,vilam3,geobenchvlm}. A closely related line of work further shows that the problem is often not the absence of visual information itself. On perception-heavy tasks, the needed signal can still be recovered from internal representations even when the model's final text answer is wrong \citep{fu2025hiddenplainsightvlms,liu2025visual}. Together, these findings suggest that the main bottleneck is not only visual encoding, but also the model's limited ability to preserve and use fine-grained visual evidence during decoding.

\citet{saini2026languageoverwritesvisionoveralignment} show that for the attention mechanism of VLMs to yield mathematically meaningful similarity scores between visual queries and textual keys (or vice versa), both modalities must operate within a shared, dimensionally
consistent coordinate system. Indeed, several studies suggest that visual features are progressively transformed into more language-like states inside the language model. Across layers, image-conditioned hidden states become more interpretable in vocabulary space, some query tokens capture global image information, and fine-grained attributes can still be recovered from spatially localized visual tokens \citep{neo2024towards,kaduri2025s,wang2025understandingknowledgeevolveslarge}. Layer-wise analyses of intermediate representations make it possible to track this transformation more directly and relate internal visual states to later textual predictions \citep{neo2024towards,kaduri2025s,liu2025visual}. Other work shows that the language decoder can compensate for weak or incomplete visual representations, suggesting a division of labor between visual features and language priors \citep{takishita2025llms,merullo2022linearly}. The same mechanism also affects downstream behavior: visual inputs can be harder than textual inputs for factual recall, and even when visual and textual tasks are matched, they can rely on partly different internal circuits \citep{cohen2025performance,nikankin2025same}.

Our work connects these two lines of research. We argue that one reason pixel-level tasks remain difficult is that VLMs perform best when the target can be mapped to a clear semantic anchor in language space. We study this in correspondence-style tasks, where success depends on matching local regions or shapes rather than naming an object category. Across semantic, shape, and face correspondence, we show that the representation-output gap becomes much larger for unnameable entities even when internal features remain sufficient. We further compare two ways of closing this gap: teaching arbitrary names, which strengthens a language-mediated shortcut, and direct task finetuning, which instead improves visual comparison without requiring the same kind of semantic recoding.

%% file: sections/7_conclusion.tex
\section{Conclusion}

We propose a simple explanation for the hidden-in-plain-sight gap: VLMs are much better at using visual information when they can attach a semantic label to it. Across semantic, synthetic, and face correspondence tasks, performance is substantially higher for familiar, nameable entities than for semantically unknown ones, even when representations retain the information needed to solve the task. Logit Lens analyses and naming interventions suggest that semantic labels let models shortcut direct visual comparison by mapping the problem to language. At the same time, semantic labeling is not necessary for strong performance. With direct task finetuning, VLMs can learn more general visual skills that transfer to unseen shapes and families while relying less on semantic anchoring. Together, our results suggest that many current VLM failures reflect a learned shortcut rather than a fundamental limitation of multimodal reasoning.

%% file: sections/8_appendix.tex
\clearpage
{
\centering
\Large\bf Supplementary Material: Appendices \\ [20pt]
}

\section{Qualitative Chain-of-Thought Examples on Named and No-Name Semantic Correspondence}

\begin{figure}[!h]
    \centering
    \includegraphics[width=0.95\linewidth]{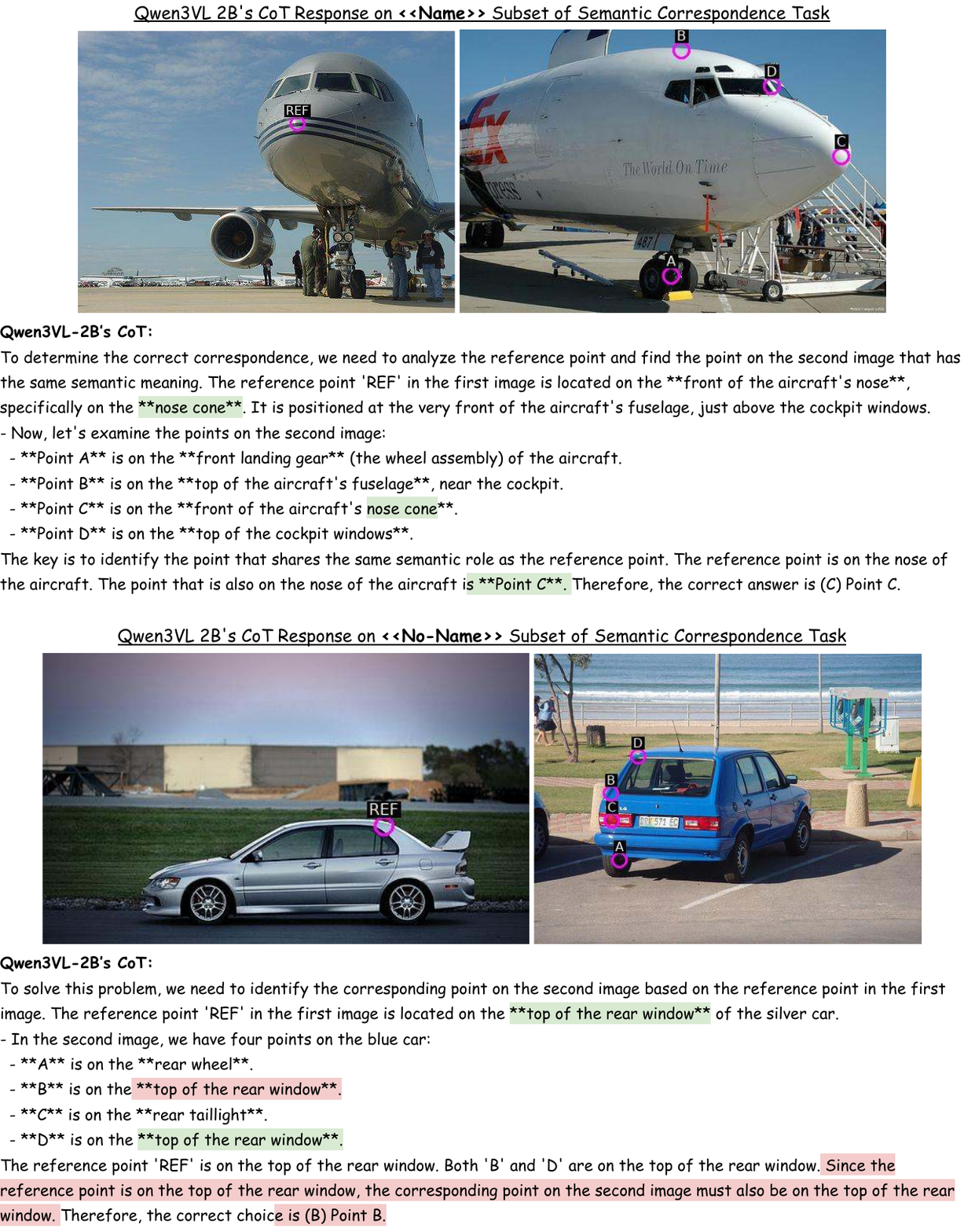}
    \caption{Chain-of-Thought reasoning allows the VLM to recover the semantic label for each point explicitly and essentially convert the task to a verbal problem. However, when the points lack a concrete semantic label, the transcription process becomes harder, leading to hallucinations, as shown by the VLM mislabeling point B as the top of the rear window instead of the bottom.}
    \label{fig:sem_corr_cot_example}
\end{figure}

\clearpage
\section{Representation Probing Performance Per Layer For All Tasks}
\label{app:rep_probe_layers}
\begin{figure}[h]
    \centering
    \begin{subfigure}[b]{0.31\textwidth}
        \centering
        \includegraphics[width=\textwidth]{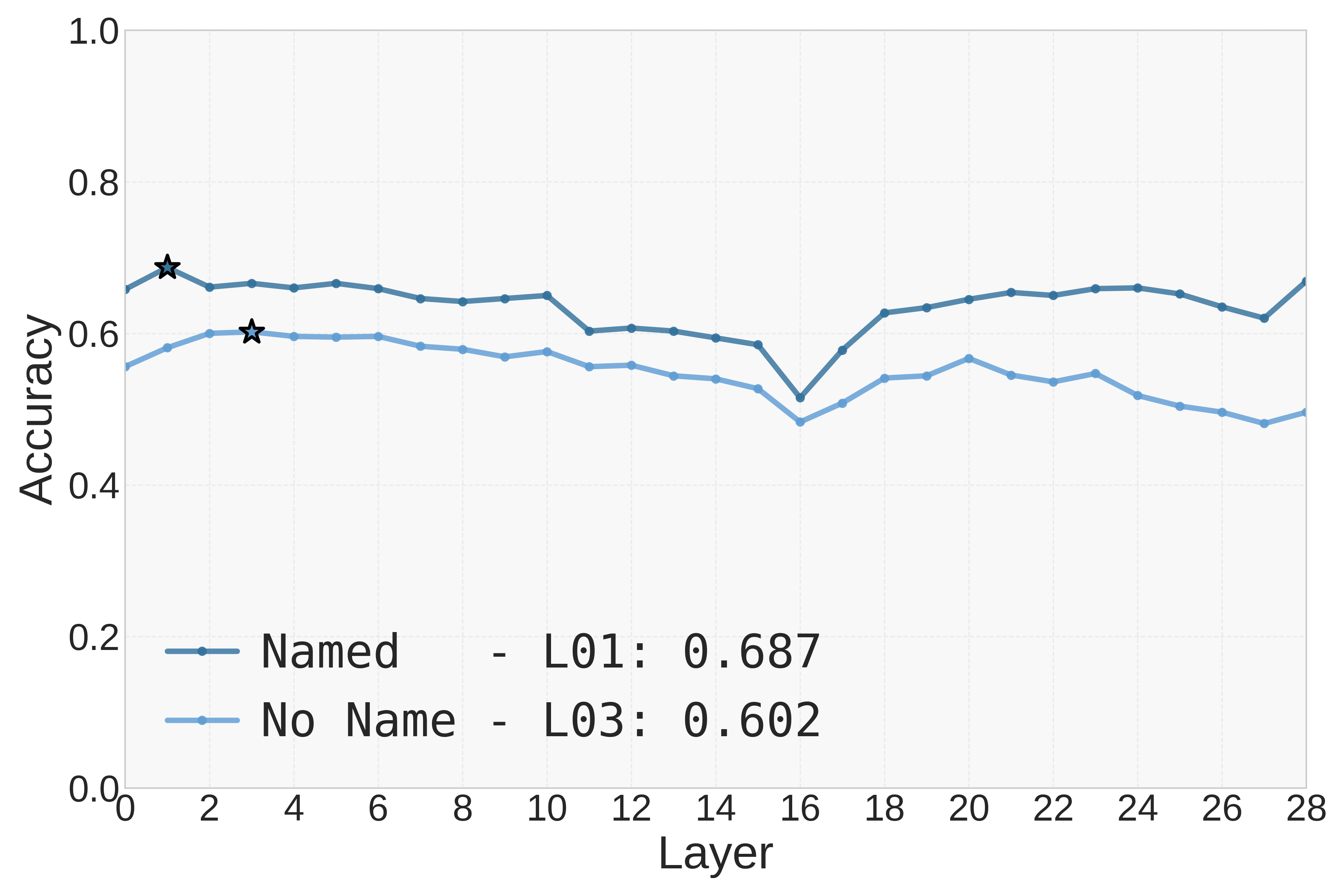}
        \caption{Qwen3VL-2B}
    \end{subfigure}
    \hfill
    \begin{subfigure}[b]{0.31\textwidth}
        \centering
        \includegraphics[width=\textwidth]{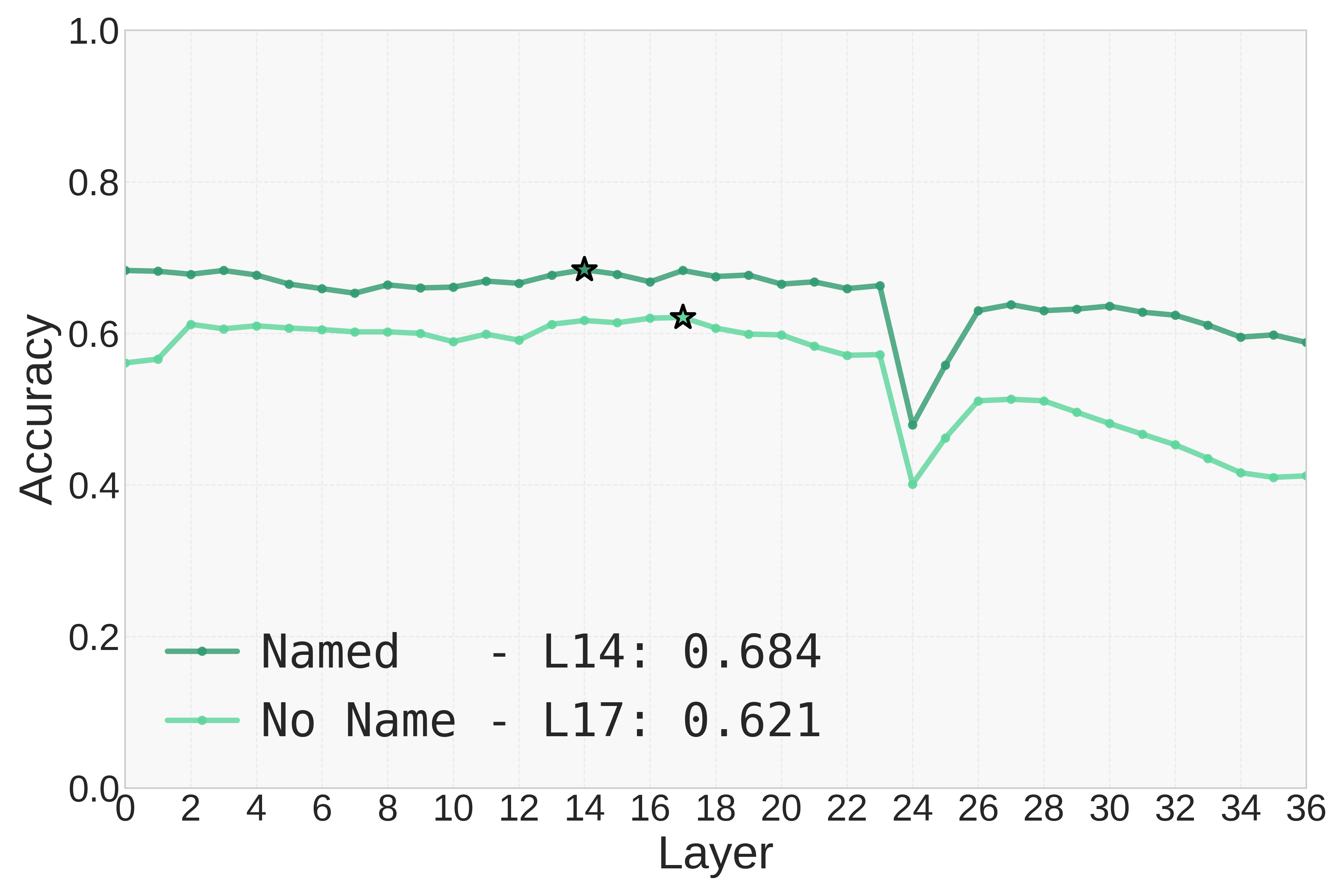}
        \caption{Qwen3VL-4B}
    \end{subfigure}
    \hfill
    \begin{subfigure}[b]{0.31\textwidth}
        \centering
        \includegraphics[width=\textwidth]{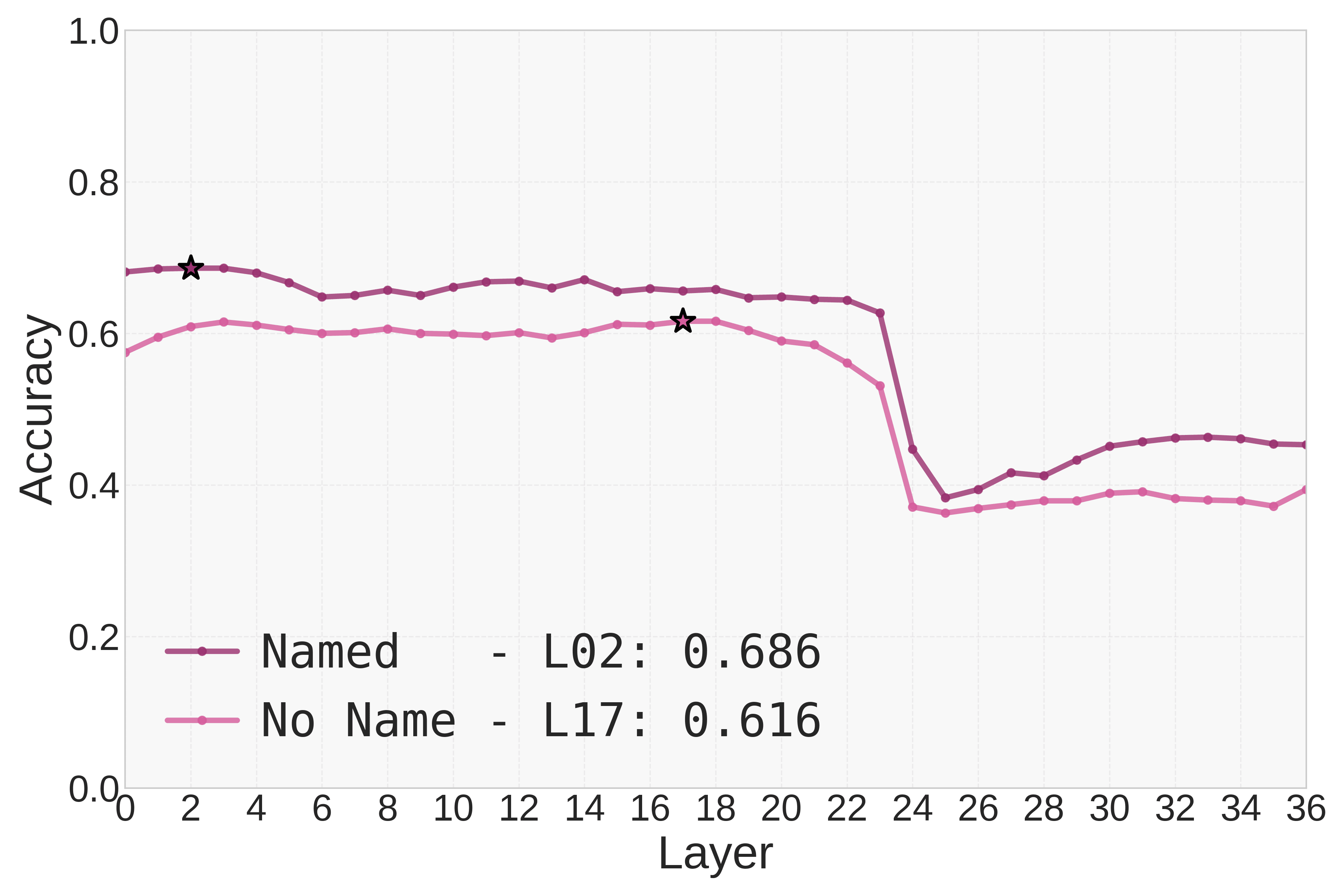}
        \caption{Qwen3VL-8B}
    \end{subfigure}
    
    \begin{subfigure}[b]{0.31\textwidth}
        \centering
        \includegraphics[width=\textwidth]{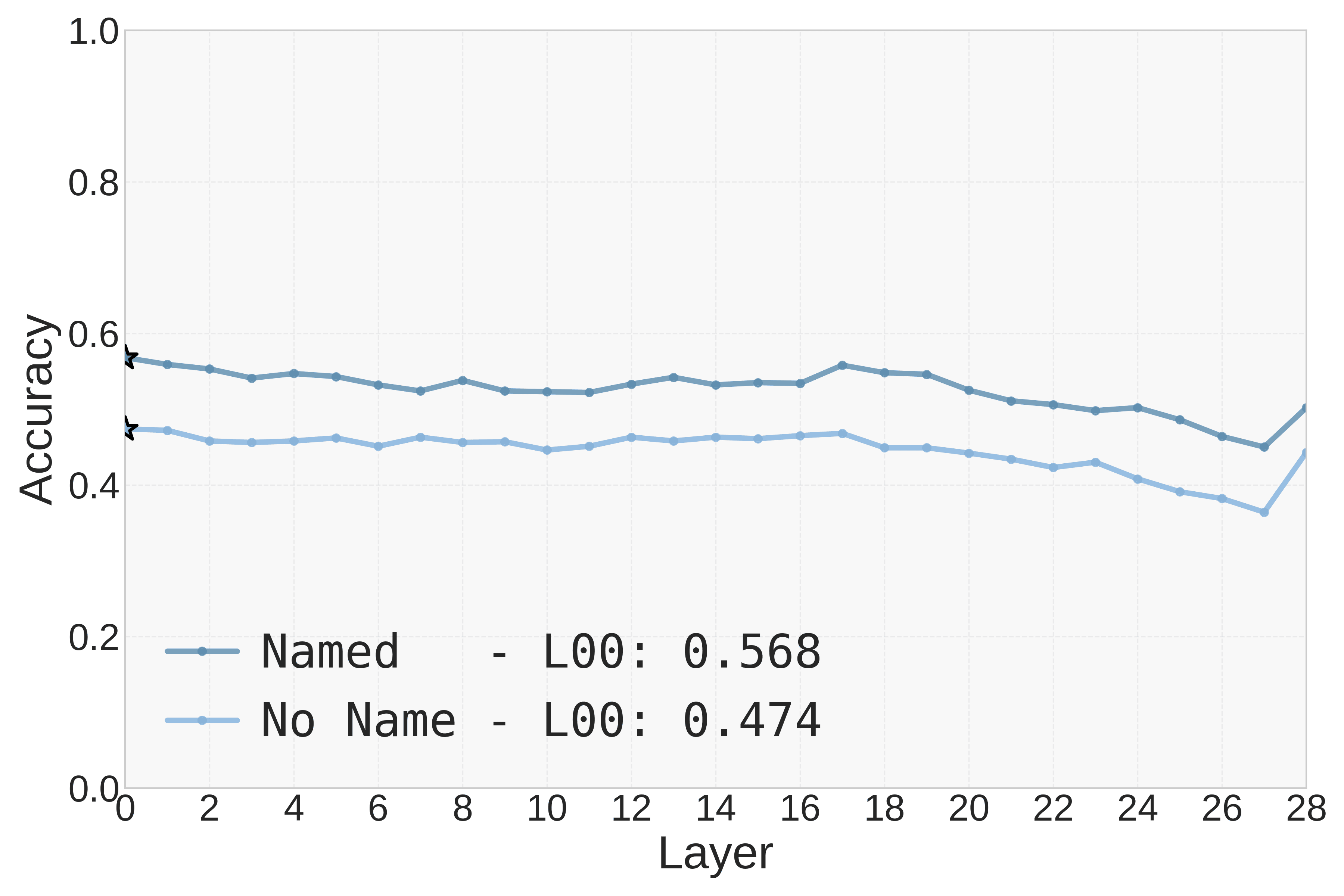}
        \caption{InternVL3.5-2B}
    \end{subfigure}
    \hfill
    \begin{subfigure}[b]{0.31\textwidth}
        \centering
        \includegraphics[width=\textwidth]{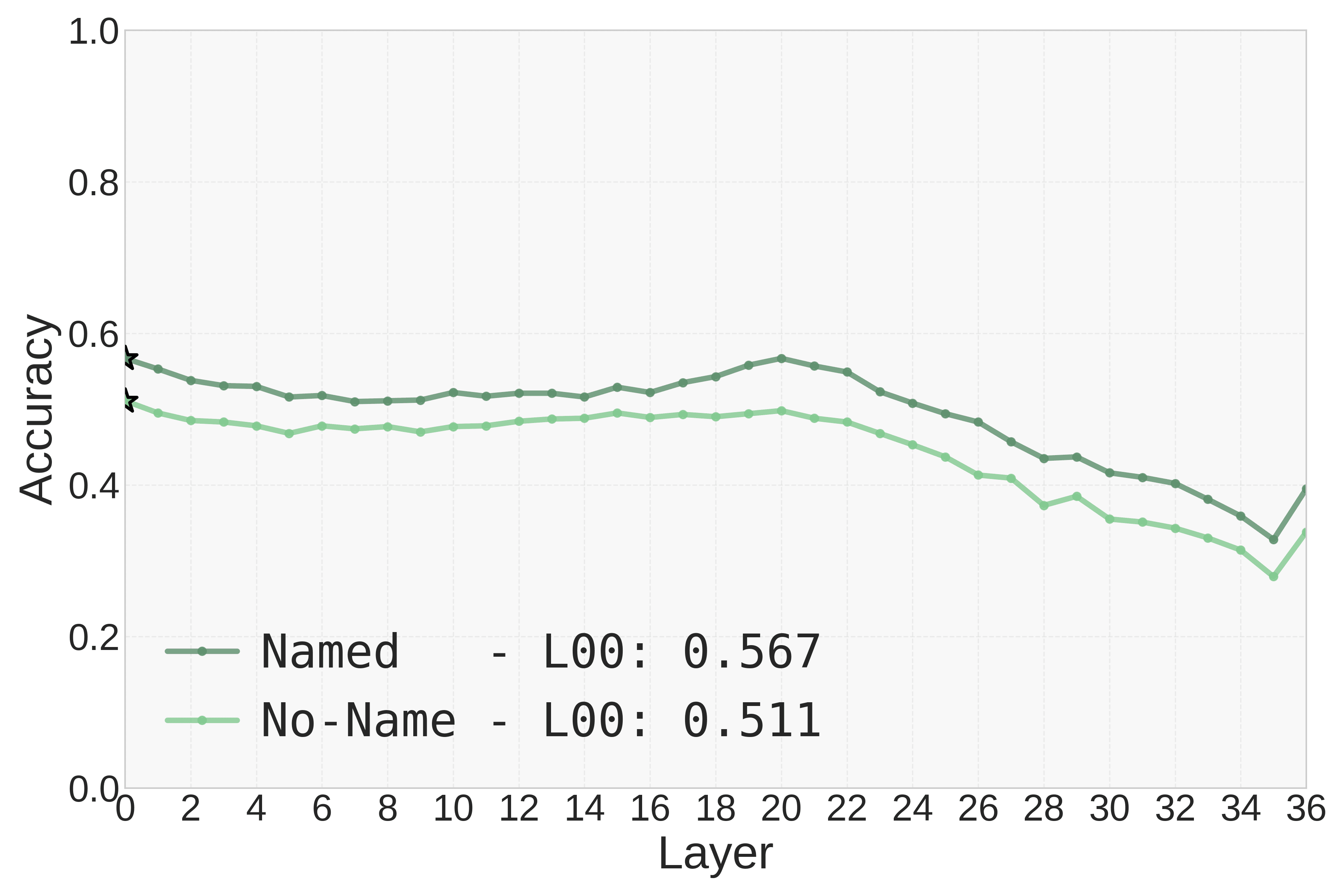}
        \caption{InternVL3.5-8B}
    \end{subfigure}
    \hfill
    \begin{subfigure}[b]{0.31\textwidth}
        \centering
        \includegraphics[width=\textwidth]{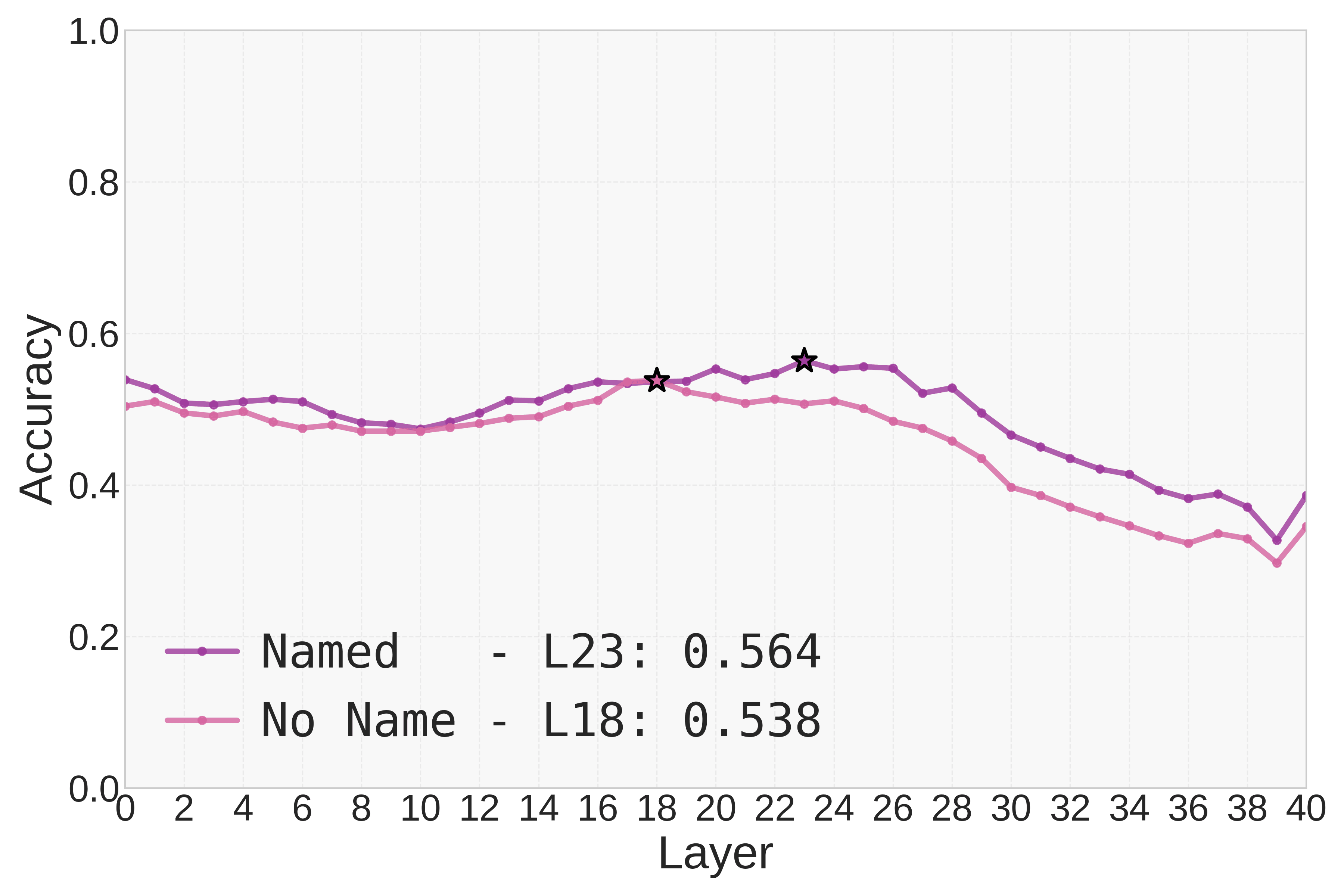}
        \caption{InternVL3.5-14B}
    \end{subfigure}
    \caption{Layer-wise Representation Probing Performance for Nameable and No-Name reference points from the \textbf{Semantic Correspondence} \citep{min2019spair} task.}
    \label{fig:rep_probe_layer_sem_corr}
\end{figure}

\begin{figure}[h]
    \centering
    \begin{subfigure}[b]{0.31\textwidth}
        \centering
        \includegraphics[width=\textwidth]{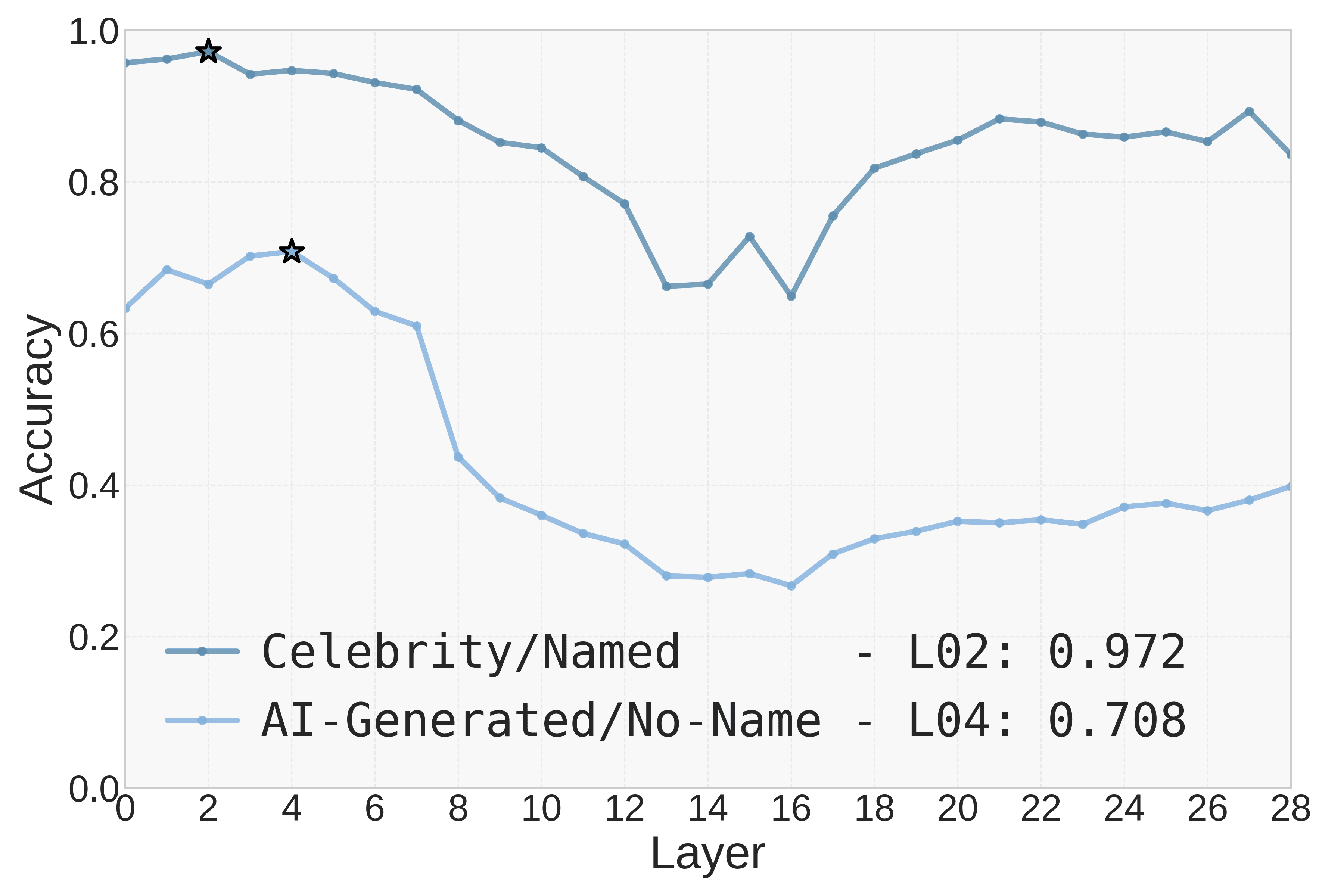}
        \caption{Qwen3VL-2B}
    \end{subfigure}
    \hfill
    \begin{subfigure}[b]{0.31\textwidth}
        \centering
        \includegraphics[width=\textwidth]{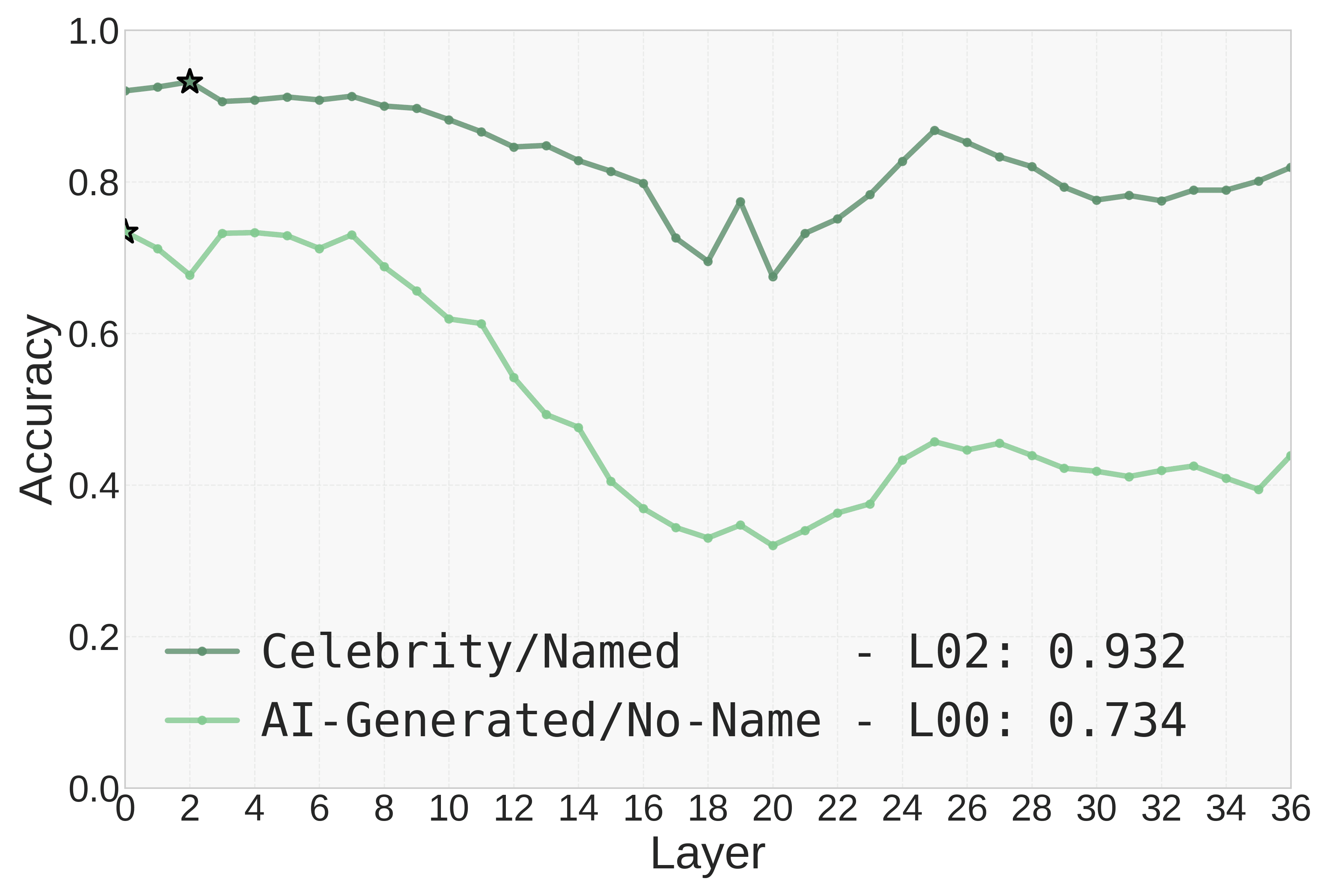}
        \caption{Qwen3VL-4B}
    \end{subfigure}
    \hfill
    \begin{subfigure}[b]{0.31\textwidth}
        \centering
        \includegraphics[width=\textwidth]{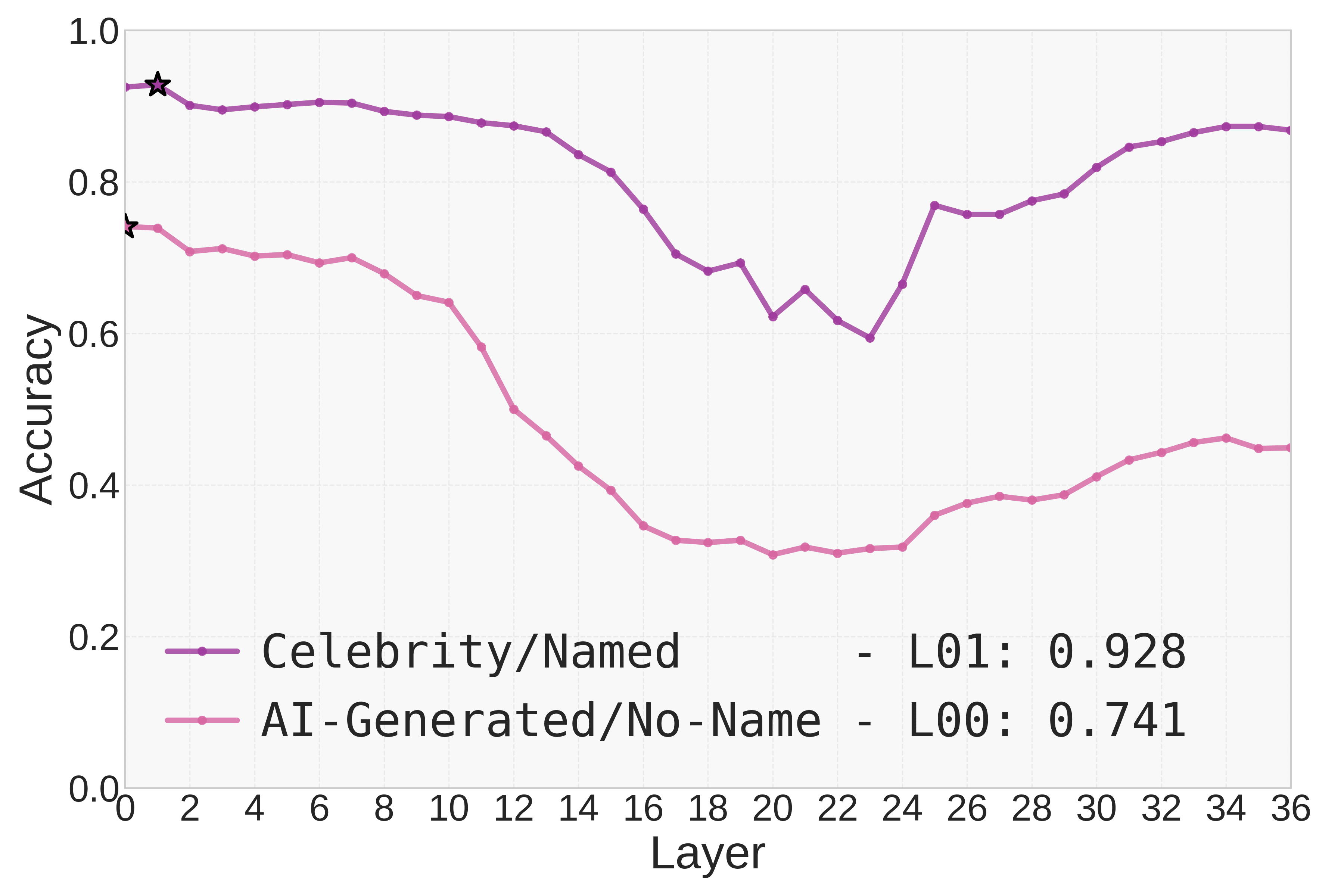}
        \caption{Qwen3VL-8B}
    \end{subfigure}
    \begin{subfigure}[b]{0.31\textwidth}
        \centering
        \includegraphics[width=\textwidth]{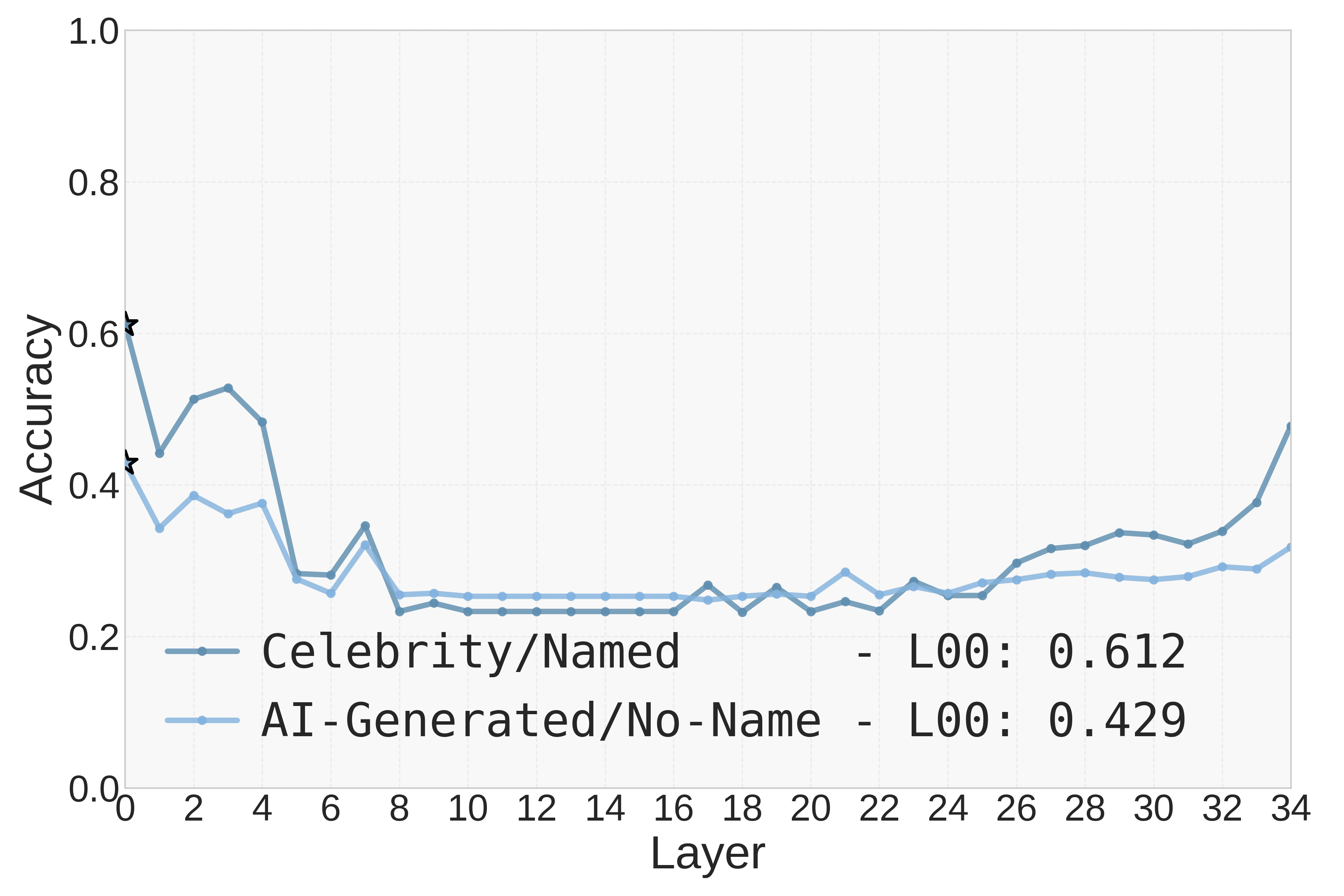}
        \caption{Gemma3-4B}
    \end{subfigure}
    \begin{subfigure}[b]{0.31\textwidth}
        \centering
        \includegraphics[width=\textwidth]{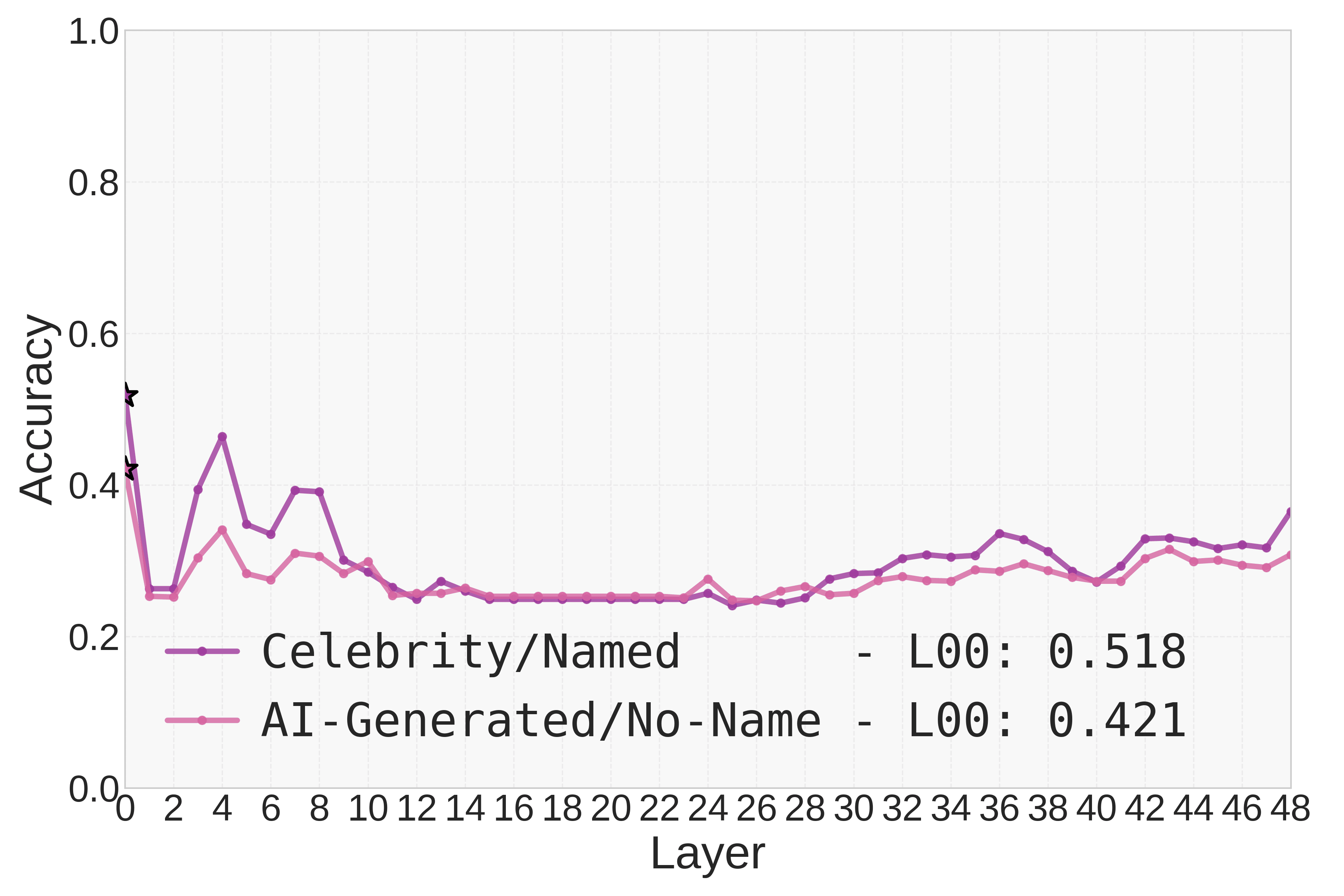}
        \caption{Gemma3-12B}
    \end{subfigure}
    \caption{Layer-wise Representation Probing Performance for \textbf{Known (Celebrity) and Unknown (AI-Generated) Faces}.}
    \label{fig:rep_probe_layer_faces}
\end{figure}

\begin{figure}[h]
    \centering
    \begin{subfigure}[b]{0.31\textwidth}
        \centering
        \includegraphics[width=\textwidth]{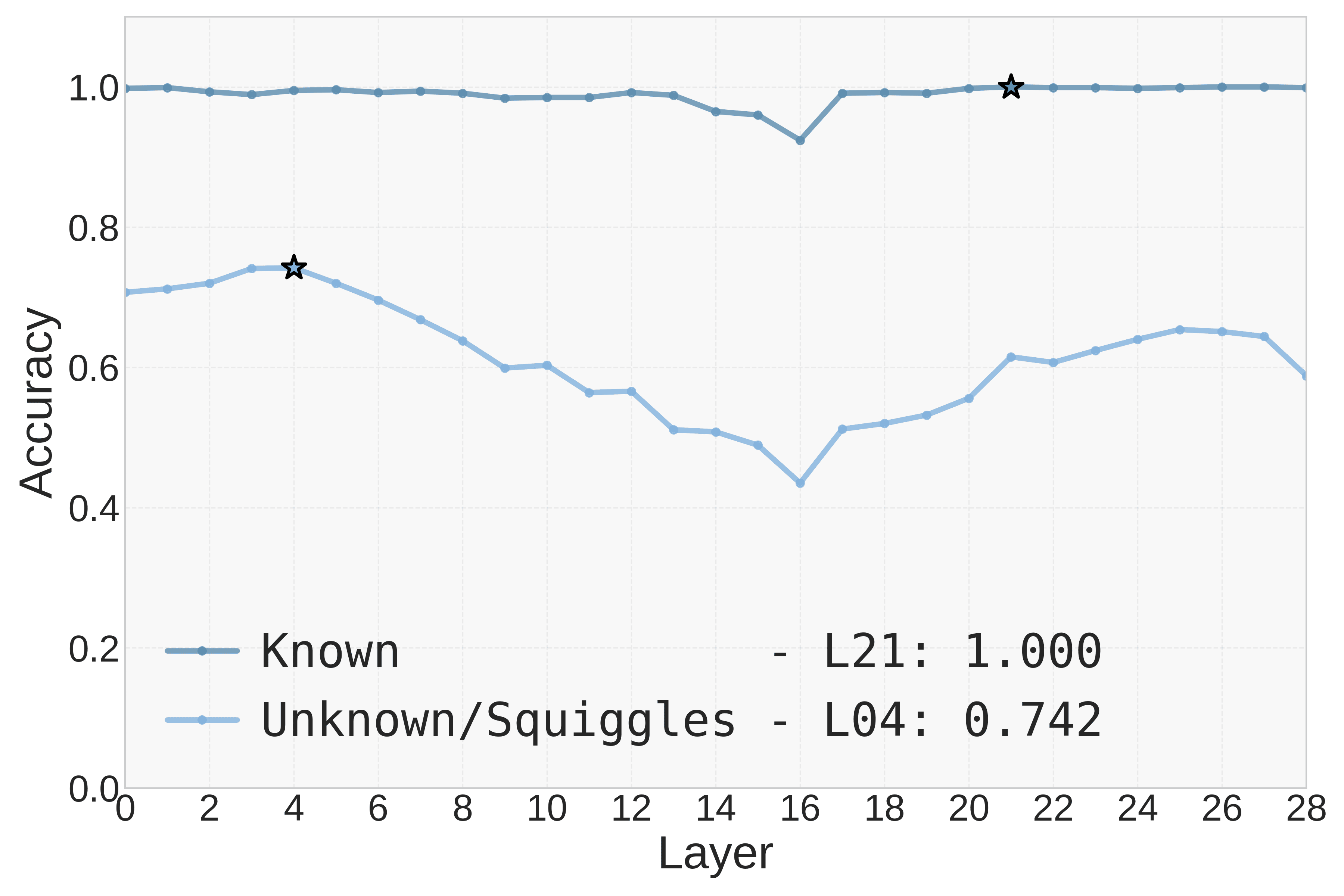}
        \caption{Qwen3VL-2B}
        \label{fig:qwen2b_shapes_rep}
    \end{subfigure}
    \hfill
    \begin{subfigure}[b]{0.31\textwidth}
        \centering
        \includegraphics[width=\textwidth]{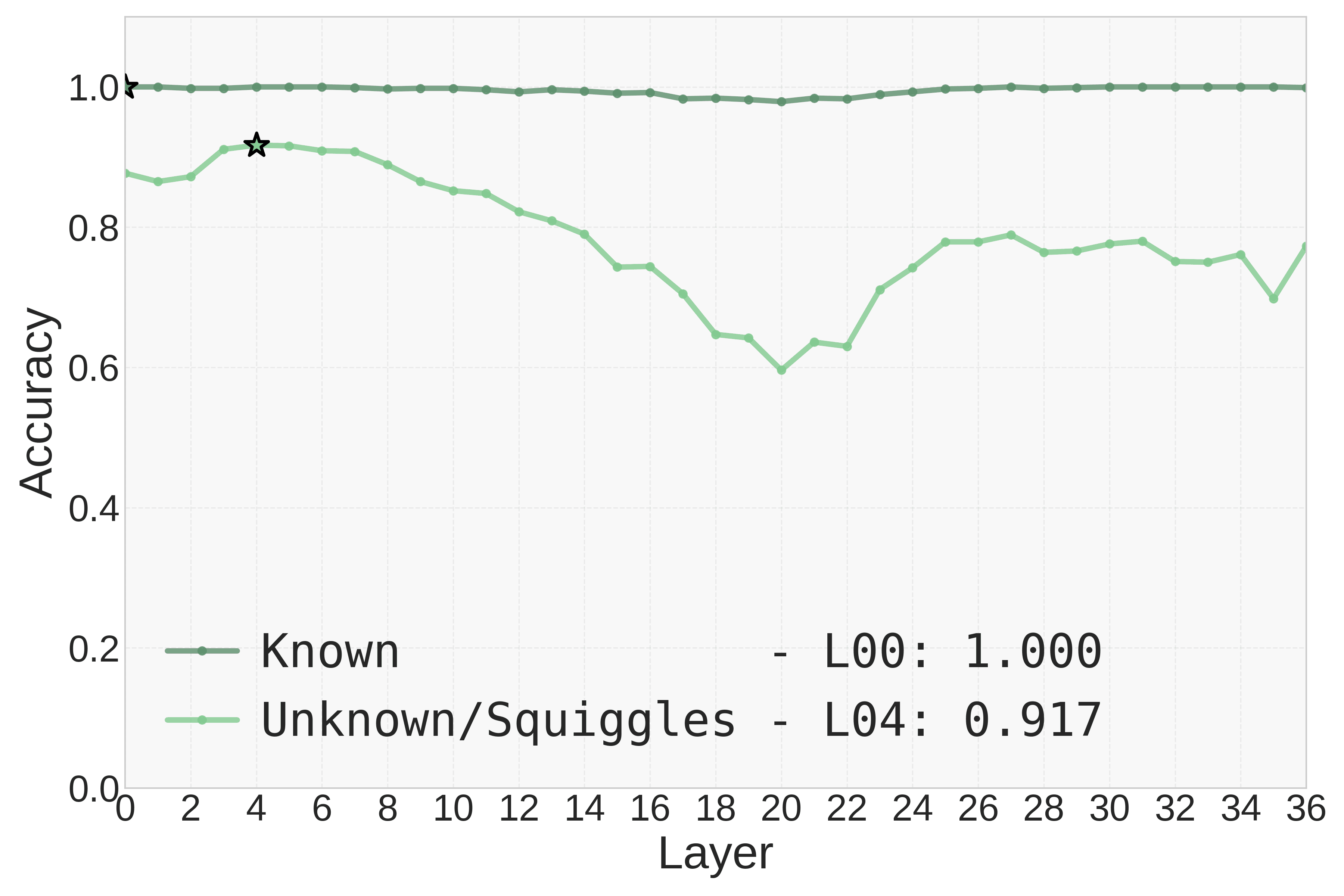}
        \caption{Qwen3VL-4B}
    \end{subfigure}
    \hfill
    \begin{subfigure}[b]{0.31\textwidth}
        \centering
        \includegraphics[width=\textwidth]{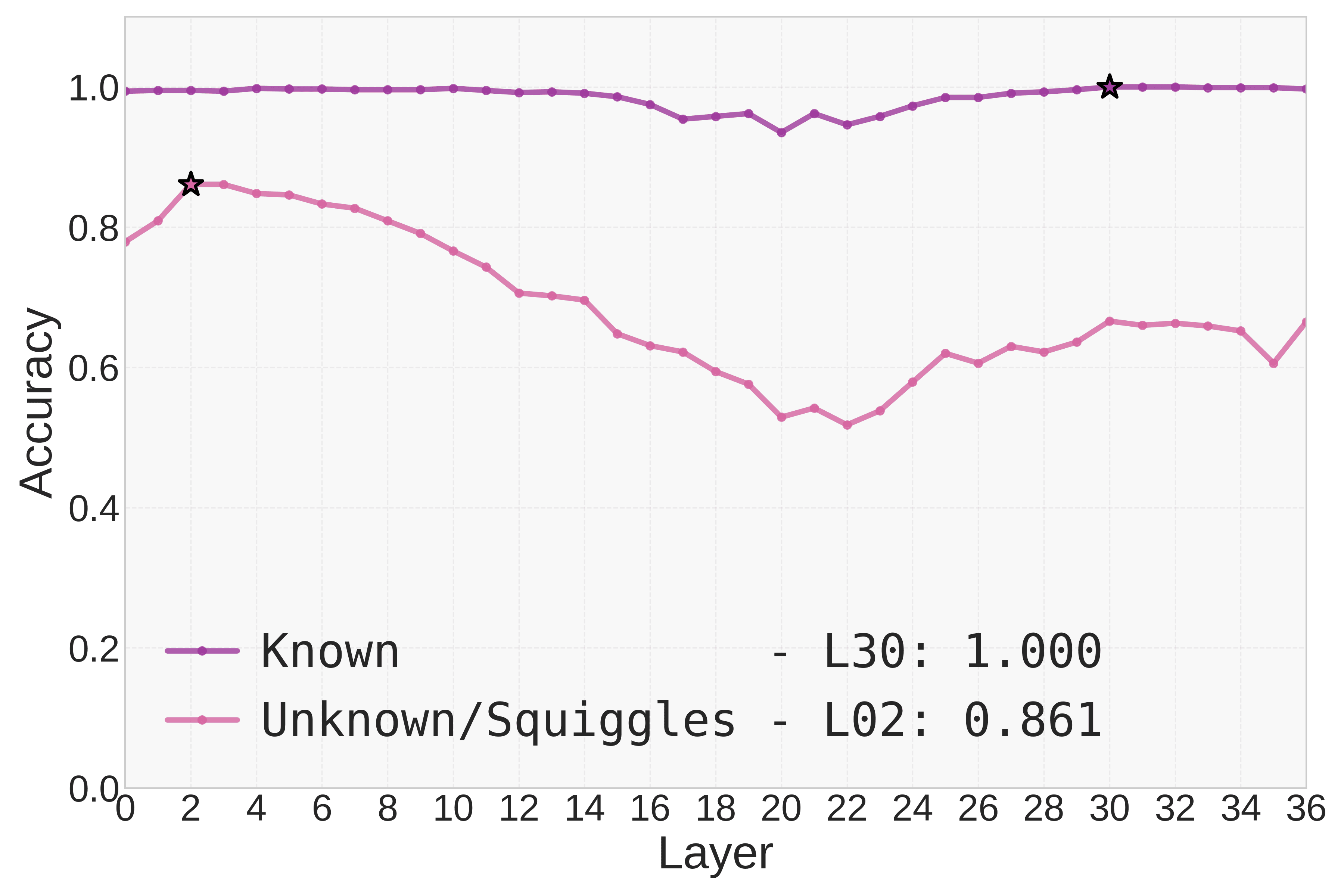}
        \caption{Qwen3VL-8B}
    \end{subfigure}
    \begin{subfigure}[b]{0.31\textwidth}
        \centering
        \includegraphics[width=\textwidth]{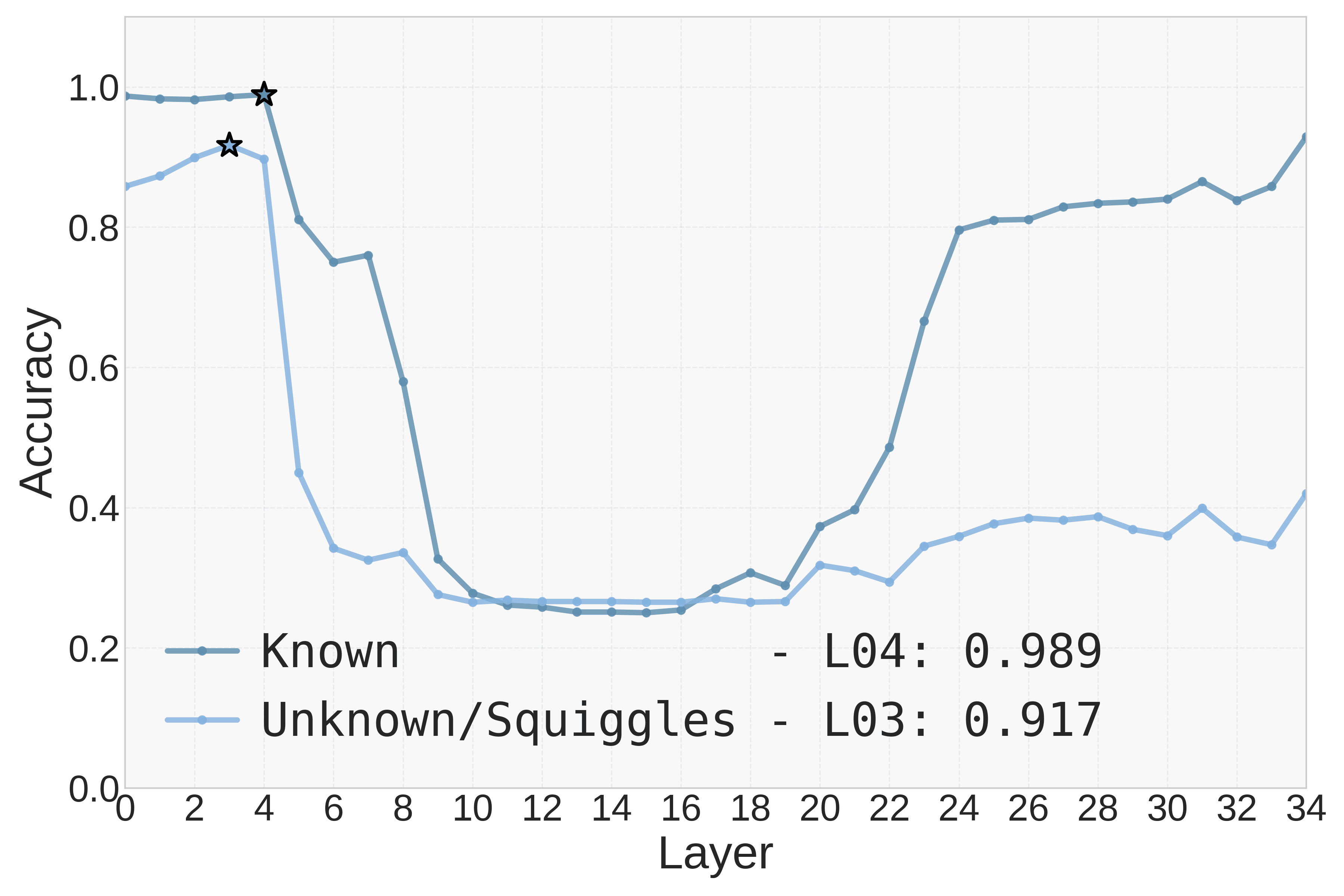}
        \caption{Gemma3-4B}
        \label{fig:gemma4b_shapes_rep}
    \end{subfigure}
    \begin{subfigure}[b]{0.31\textwidth}
        \centering
        \includegraphics[width=\textwidth]{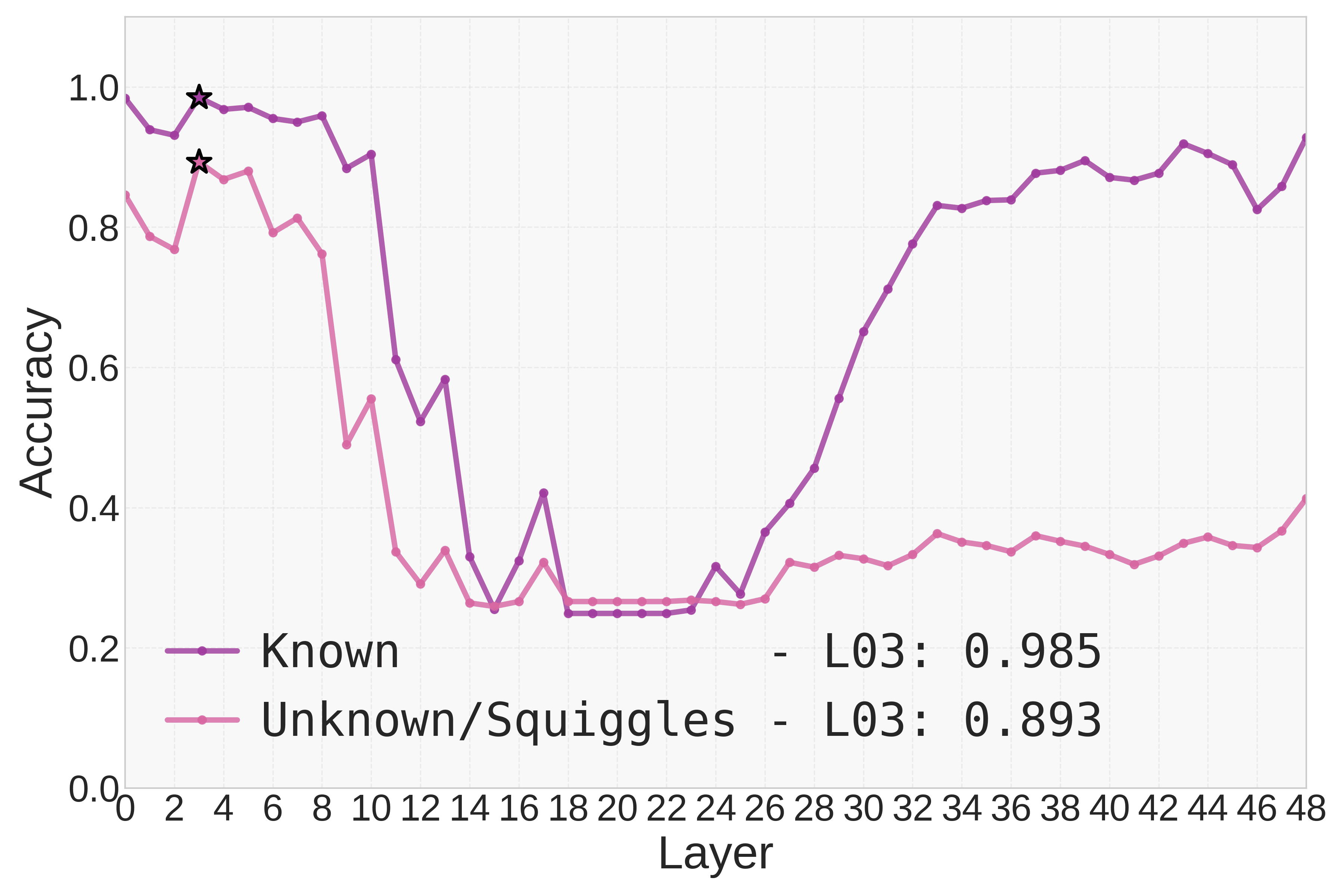}
        \caption{Gemma3-12B}
    \end{subfigure}
    
    \begin{subfigure}[b]{0.31\textwidth}
        \centering
        \includegraphics[width=\textwidth]{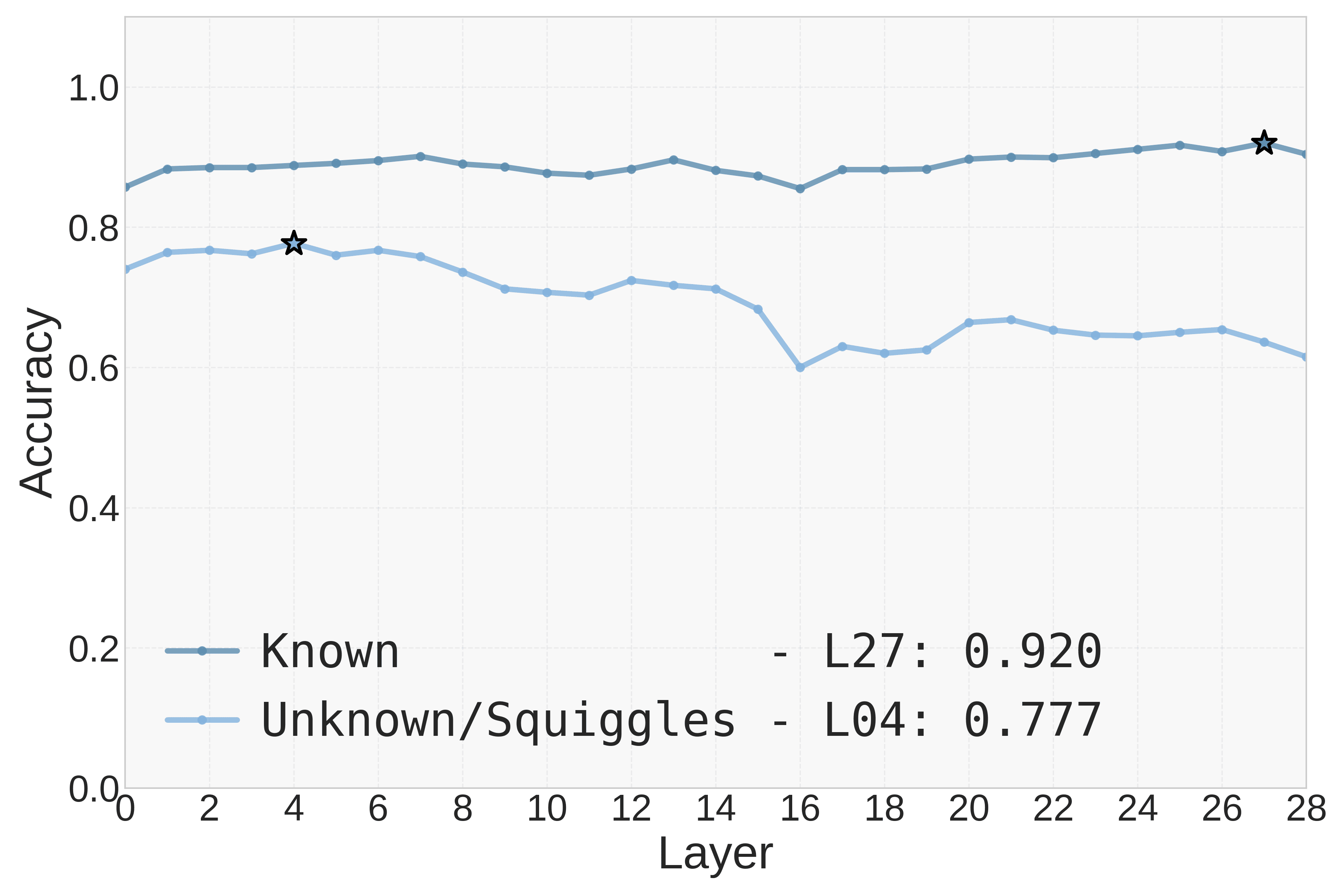}
        \caption{InternVL-3.5-2B}
        \label{fig:internvl352b_shapes_rep}
    \end{subfigure}
    \hfill
    \begin{subfigure}[b]{0.31\textwidth}
        \centering
        \includegraphics[width=\textwidth]{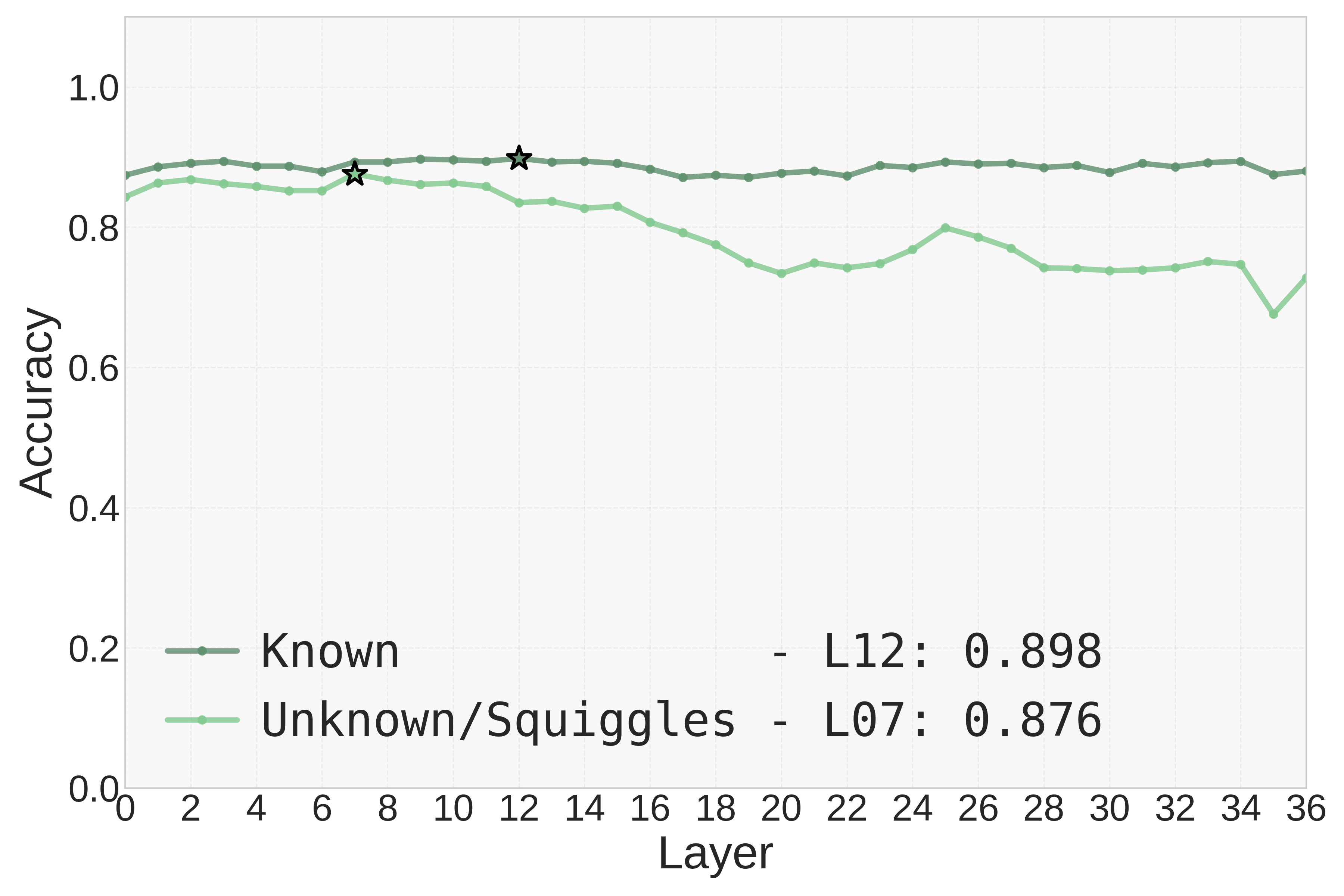}
        \caption{InternVL-3.5-8B}
    \end{subfigure}
    \hfill
    \begin{subfigure}[b]{0.31\textwidth}
        \centering
        \includegraphics[width=\textwidth]{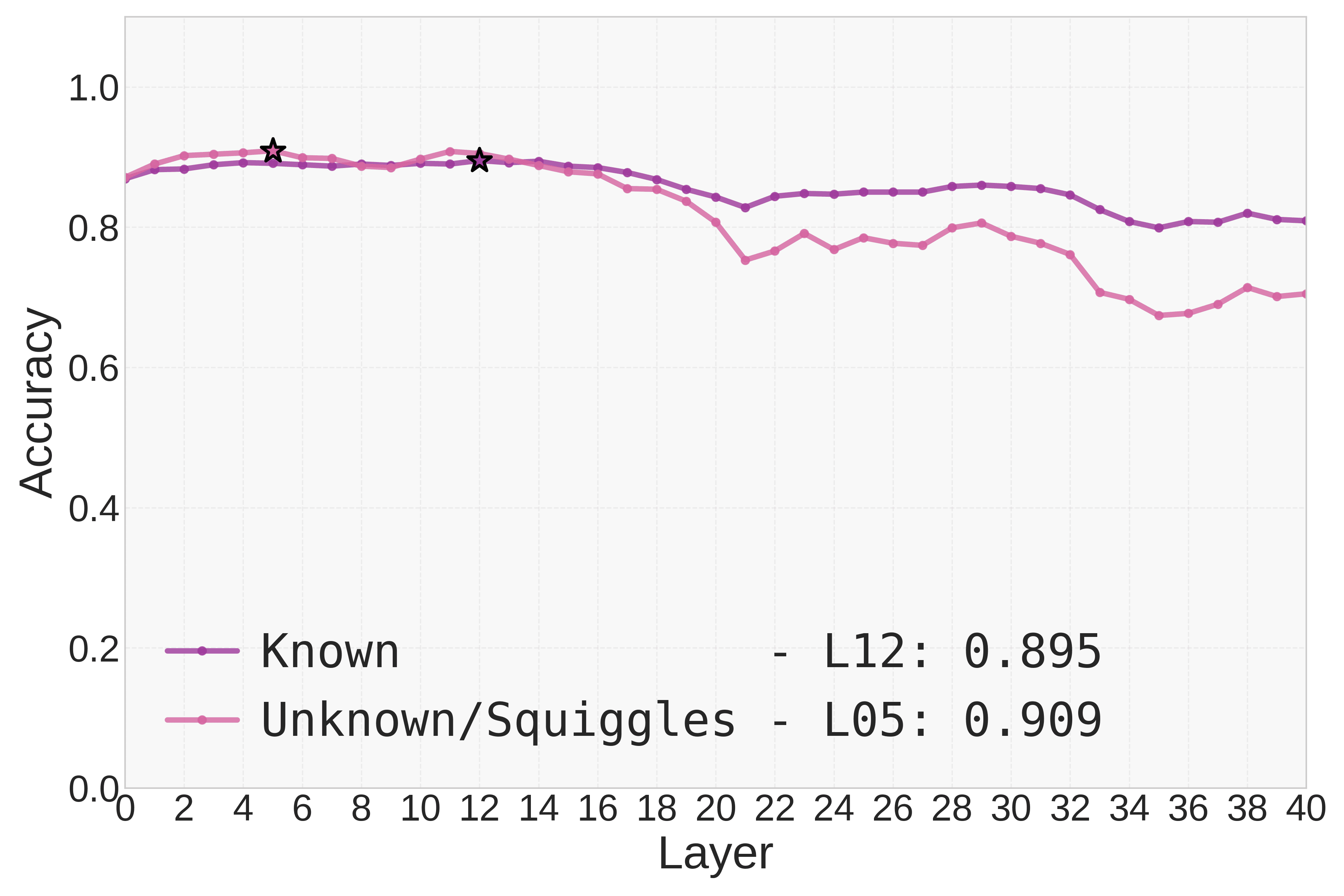}
        \caption{InternVL-3.5-14B}
    \end{subfigure}
    \caption{Layer-wise Representation Probing Performance for \textbf{2D Known and Procedurally Generated (Squiggle) Shapes}.}
    \label{fig:rep_probe_layer_shapes}
\end{figure}

\clearpage
\section{Faces Dataset Experimental Setup}
\label{app:faces_dataset}

\begin{figure}[h]
    \centering   
    \begin{subfigure}[c]{0.18\textwidth}
        \centering
        \includegraphics[width=\textwidth]{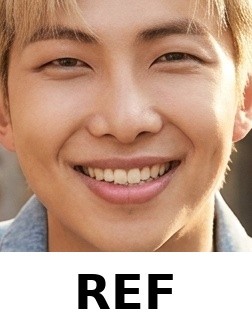}
        \caption{Celebrity Ref.}
    \end{subfigure}
    \hfill
    \begin{subfigure}[c]{0.23\textwidth}
        \centering
        \includegraphics[width=\textwidth]{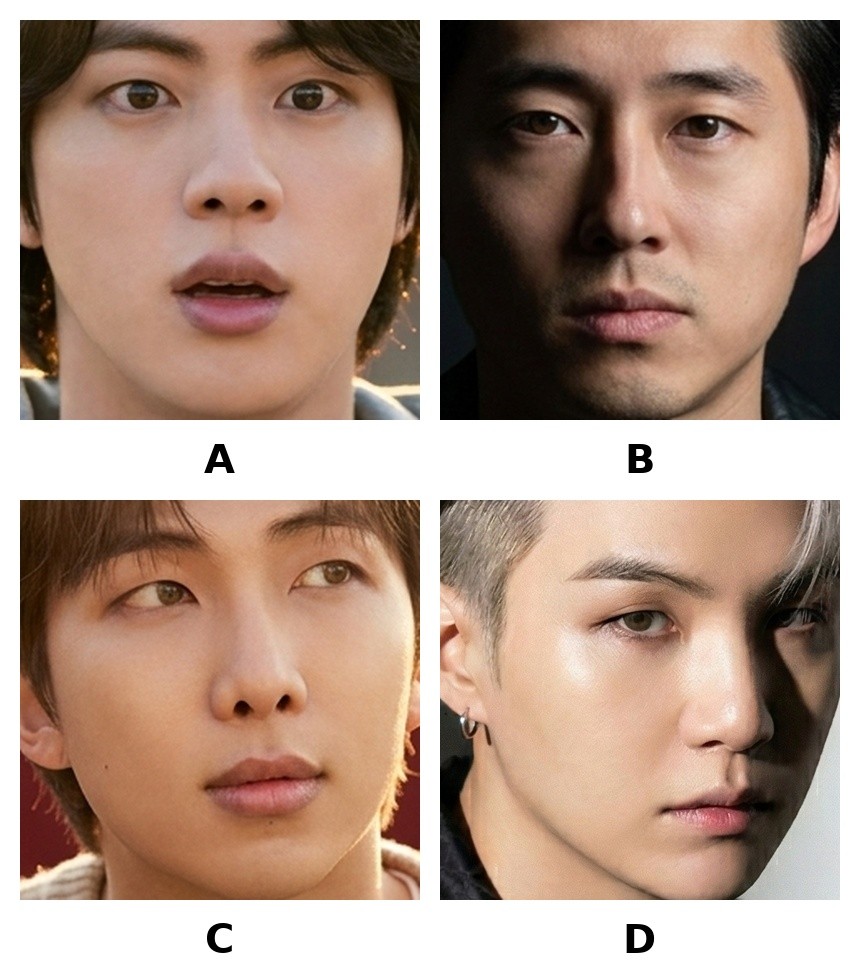}
        \caption{Celebrity Tgt.}
    \end{subfigure}
    \hfill
    \begin{subfigure}[c]{0.17\textwidth}
        \centering
        \includegraphics[width=\textwidth]{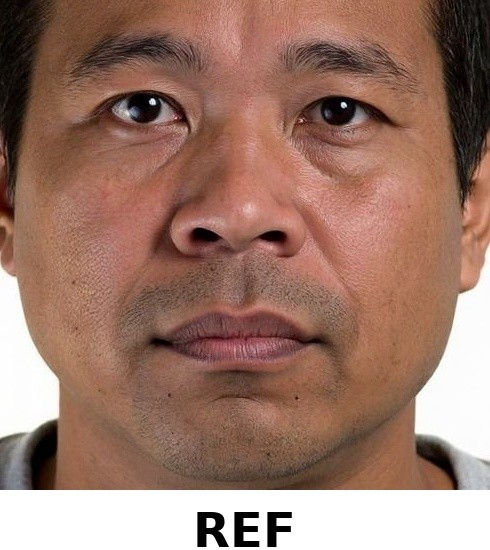}
        \caption{AI-Gen. Ref.}
    \end{subfigure}
    \hfill
    \begin{subfigure}[c]{0.23\textwidth}
        \centering
        \includegraphics[width=\textwidth]{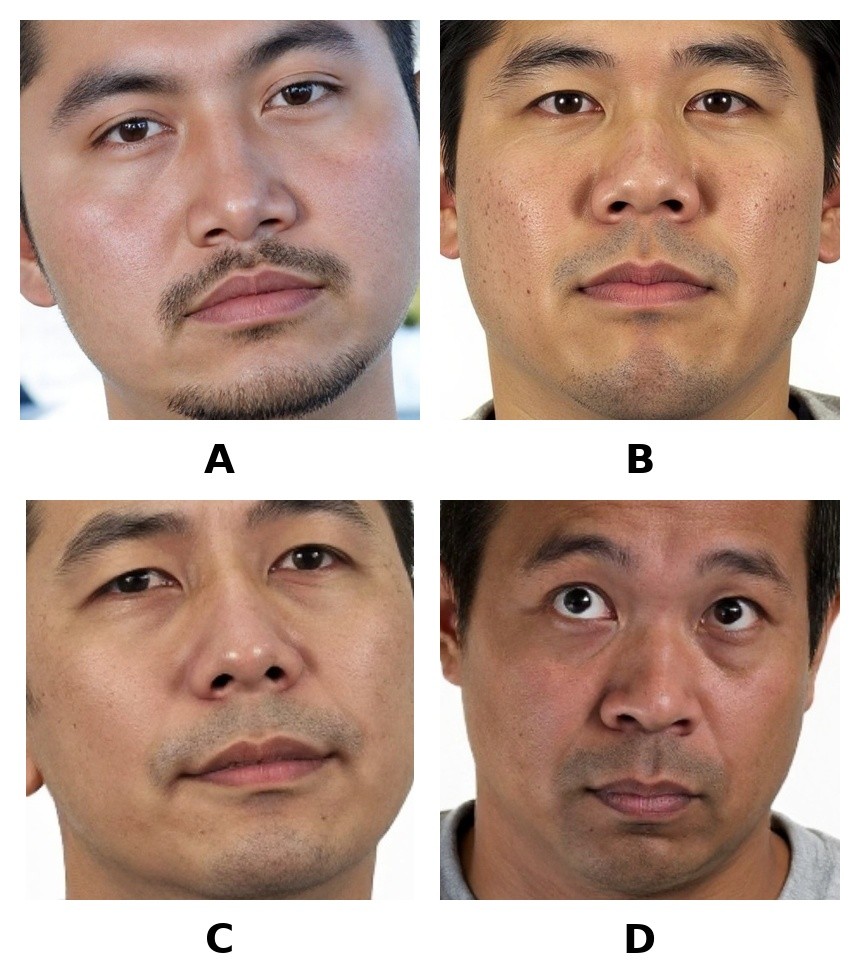}
        \caption{AI-Gen. Tgt.}
    \end{subfigure}

    \caption{Face correspondence task. Known celebrities (left) and AI-generated faces (right).}
    \label{fig:face_corr_task}
\end{figure}

To build our face correspondence dataset, we collect an initial set of celebrities and use Nano-Banana-2 \citep{nano_banana} to generate four images per person, ensuring uniform resolution across all samples and avoiding artifacts that may arise from real images scraped from the internet. We construct model-specific splits by probing each VLM's knowledge: given an image, we prompt the model with "Who is the person in this image?" and designate a person as known to the model if at least one of the four generated images is identified correctly. Table \ref{tab:recognized-people} shows the initial set and model-specific subsets of celebrities.

\begin{table}[!h]
\centering
\begin{tabular}{p{0.18\linewidth} p{0.75\linewidth}}
\toprule
\textbf{VLM} & \textbf{Recognized Faces} \\
\midrule
All & Bruce Lee, BTS J-Hope, BTS Jimin, BTS Jin, BTS Jungkook, BTS RM, BTS Suga, BTS V, Jack Ma, Jackie Chan, Masayoshi Son, Song Heungmin, Steven Yeun, Yao Ming \\
\midrule
Qwen-2B & BTS Jimin, BTS Jin, BTS Jungkook, BTS Suga, BTS V, Bruce Lee, Jack Ma, Jackie Chan, Song Heungmin, Yao Ming \\
\midrule
Qwen-4B & BTS J-Hope, BTS Jimin, BTS Jin, BTS RM, BTS Suga, BTS V, Bruce Lee, Jack Ma, Jackie Chan, Steven Yeun, Yao Ming \\
\midrule
Qwen-8B & BTS Jimin, BTS Jin, BTS Jungkook, BTS Suga, BTS V, Bruce Lee, Jack Ma, Jackie Chan, Steven Yeun, Yao Ming \\
\midrule
Gemma-4B & BTS Jimin, BTS Jin, BTS Jungkook, BTS RM, BTS V, Bruce Lee, Jack Ma, Jackie Chan, Masayoshi Son, Steven Yeun \\
\midrule
Gemma-12B & BTS Jimin, BTS Jin, BTS Jungkook, BTS RM, BTS Suga, BTS V, Jackie Chan, Steven Yeun \\
\bottomrule
\end{tabular}
\caption{VLM-specific Face Correspondence dataset.}
\label{tab:recognized-people}
\end{table}

\paragraph{Nano Banana 2 Prompt}

\texttt{\{Reference Image\} Generate four images of this person with varying expressions, lighting, and poses. Also, change his clothes and hairstyles. Ensure the images are square, close-ups of his face. The person should not look too far to one side. The face should be almost fully visible. There should be no visible hands in the image.}

\clearpage
\section{InternVL3.5's Facial Recognition Failure and Results on Shape Correspondence}
\label{app:internvl_appendix}

\paragraph{Facial Recognition Failures}
Our pilot experiments revealed that the InternVL3.5 model family fails to recognize the images of most celebrities. We probe the VLM with the prompt \textit{``Who is the person in this image? Respond with the format: 'The person in the image is <name>.' with no other text."}. Table \ref{tab:internvl_face_probe_results} shows that InternVL-3.5-14B fails to recognize all but two celebrities. InternVL-3.5-2B and 8B are similarly limited. As such, we did not evaluate InternVl-3.5 on the facial correspondence task. 

\begin{table}[h]
\centering
\renewcommand{\arraystretch}{1.2}
\begin{tabular}{p{3cm}p{8cm}c}
\hline
\textbf{Ground Truth} & \textbf{InternVl3.5 14B Predictions (Images 1--4)} & \textbf{Correct} \\
\hline
Barack Obama      & Barack Obama, Barack Obama, Barack Obama, Barack Obama                           & \correct{4/4} \\
Cristiano Ronaldo & \textit{IDK}, Cristiano Ronaldo, Paul Walker, \textit{IDK}                      & \partialcolor{1/4} \\
Donald Trump      & Donald Trump, Donald Trump, Donald Trump, Donald Trump                          & \correct{4/4} \\
Lionel Messi      & Cristiano Ronaldo, Cristiano Ronaldo, Chris Hemsworth, Aaron Hernandez          & \wrong{0/4}   \\
\hline
BTS J-Hope        & Park Yohan, Park Yoo-chun, Jungkook, Lee Min-ho                                 & \wrong{0/4} \\
BTS Jimin         & Lee Seung-gi, Yohan Lemoine, Park Yohan, Park Yohan                             & \wrong{0/4} \\
BTS Jin           & Park Yoo-chun, Park Bo-gum, Park Yohan, Park Yoo-chun                           & \wrong{0/4} \\
BTS Jungkook      & Lee Seung-gi, Park Yoo-chun, Park Yohan, Park Yoo-chun                          & \wrong{0/4} \\
BTS RM            & Park Yohan, Lee Hong-gi, Lee Jong-suk, Lee Min-ho                               & \wrong{0/4} \\
BTS Suga          & Park Yohan, Park Yoo-chun, Park Yohan, Lee Jong-suk                             & \wrong{0/4} \\
BTS V             & Park Yohan, Lee Seung-gi, Park Yohan, Lee Min-ho                                & \wrong{0/4} \\
Bruce Lee         & Hu Ge, Tony Leung, Lee Min-ho, Tony Leung                                       & \wrong{0/4} \\
Jack Ma           & \textit{IDK}, \textit{IDK}, Chen Songlian, \textit{IDK}                         & \wrong{0/4} \\
Jackie Chan       & \textit{IDK}, \textit{IDK}, \textit{IDK}, Chen Kaige                            & \wrong{0/4} \\
Jet Li            & Tony Leung, \textit{IDK}, Tony Leung, \textit{IDK}                              & \wrong{0/4} \\
Masayoshi Son     & Chen Songlian, Masahiro Hamaguchi, Tetsuro Yamada, Chen Songlian                & \wrong{0/4} \\
Shigeru Miyamoto  & Tetsuro Ishida, Masahiro Miki, Chen Kaige, Tetsu Komai                          & \wrong{0/4} \\
Shohei Ohtani     & Aaron Yonda, Lee Jong-suk, Aaron Kwok, Aaron Yonda                              & \wrong{0/4} \\
Son Heung-min     & Lee Min-ho, Jun, Lee Jong-suk, Chen Kun                                         & \wrong{0/4} \\
Steven Yeun       & Aaron Kwok, Chen Liang, Lee Byung-hun, Shuichi Saihara                          & \wrong{0/4} \\
Yao Ming          & Xiaoyu Wang, Aaron Kwok, Joonas Suotamo, Wang Zhi                               & \wrong{0/4} \\
\hline
\end{tabular}
\caption{InternVL 3.5 14B face recognition probing results. \textit{IDK} means the VLM responded that it was unable to identify the person. We test additional faces of very famous celebrities to confirm InternVL-3.5 failure in face recognition.}
\label{tab:internvl_face_probe_results}
\end{table}

\paragraph{InternVL-3.5 on Shape Correspondence}

We evaluated the InternVL-3.5 model family on shape correspondence since they recognize common geometric shapes (``square", ``circle", ``star", etc). Table \ref{tab:intern_squiggles} shows that InternVL-3.5 has a larger gap between representation probing and verbal output for unknown shapes. Furthermore, Chain-of-Thought reasoning is only beneficial when the shapes are known, completely corroborating our findings for Gemma3 and Qwen3VL in Table \ref{tab:shape_face_corr}.

\begin{table}[h]
\centering
\small
\setlength{\tabcolsep}{4pt}
\renewcommand{\arraystretch}{1.15}

\begin{tabular}{@{} l c c c c c c c @{}}
\toprule
\multicolumn{2}{c}{\textbf{Model}} & \textbf{Known}
  & \multicolumn{5}{c@{}}{\textbf{2D Shapes}} \\
\cmidrule(lr){4-8}
 & & 
  & \makecell{\textbf{Direct}\\\textbf{(D)}}
  & \makecell{\textbf{CoT}\\\textbf{(C)}}
  & \makecell{\textbf{C$-$D}\\\textbf{$\Delta$}}
  & \makecell{\textbf{Rep. Probe}\\\textbf{(R)}}
  & \makecell{\textbf{R$-$max(D,C)}\\\textbf{$\Delta$}} \\
\midrule

\multirow{6}{*}{\rotatebox[origin=c]{90}{InternVL 3.5}}
  & \multirow{2}{*}{2B}  & \checkmark
    & 46.1& 76& 29.9& 92& 16\\
  &                      & $\times$  
    & 26.1& 27.7& 1.6& 77.7& \textbf{50}\\
  \cmidrule(lr){2-8}
  & \multirow{2}{*}{8B}  & \checkmark
    & 66.4& 89.9& 23.5& 89.8& -0.1\\
  &                      & $\times$  
    & 27.7& 27.6& -0.1& 86.8& \textbf{59.1}\\
  \cmidrule(lr){2-8}
  & \multirow{2}{*}{14B}  & \checkmark
    & 55.1& 90.9& 35.8& 89.5& -1.4\\
  &                      & $\times$  
    & 32.6& 31.4& -1.2& 90.4& \textbf{57.8}\\
\bottomrule
\end{tabular}%

\caption{Comparison of Direct, Chain-of-Thought, and Representational Probe accuracy for InternVL-3.5. R denotes the representation-probing accuracy from the \textbf{best-performing layer} of the language model. We report layer-wise performance in Appendix Fig. \ref{fig:rep_probe_layer_shapes}. $\Delta$ columns show the gain of CoT over Direct (C$-$D) and of the representational probe over the stronger of the two baselines (R$-$max(D,C)). Bold entries indicate that the probe substantially exceeds the best prompted baseline for Unknown 2D-shapes.}
\label{tab:intern_squiggles}
\end{table}
\section{Gemma-3 Performance on Semantic Correspondence}
\label{app:Gemma-3-Sem-Corr}

\begin{table}[h]
\centering
\resizebox{0.85\linewidth}{!}{%
\begin{tabular}{lllccc|cc}
\toprule
\textbf{Model} & \textbf{Size} & \textbf{Subset}
  & \makecell{\textbf{Direct}\\\textbf{(D)}}
  & \makecell{\textbf{CoT}\\\textbf{(C)}}
  & \makecell{\textbf{C$-$D}\\\textbf{$\Delta$}}
  & \makecell{\textbf{Rep. Probe}\\\textbf{(R)}}
  & \makecell{\textbf{R$-$max(D,C)}\\\textbf{$\Delta$}} \\
\midrule
\multirow{6}{*}{Gemma-3}
& \multirow{2}{*}{4B}  & Named   & 25.7 & 26.3 & 0.6  & 38.6 & \textbf{12.3} \\
&                      & No-Name & 24.8 & 26.2 & \textbf{1.4}           & 34.6 & 8.4 \\
\cmidrule{2-8}
& \multirow{2}{*}{12B}  & Named   & 31.2 & 35.9 & \textbf{4.7} & 39.4 & 3.5 \\
&                      & No-Name & 27.5 & 28.4 & 0.9           & 33.5 & \textbf{5.1} \\
\bottomrule
\end{tabular}
}

\caption{Semantic correspondence results for Gemma-3 on SPairs71k. Gemma-3-4B performs near chance and therefore does not exhibit the trends observed in more capable VLMs. In contrast, Gemma-3-12B shows a larger gap between Representation Probing and the best textual strategy, \textbf{R$-$max(D, C)}, on No-Name keypoints, consistent with the pattern observed for the Qwen3-VL and InternVL3.5 families in Table \ref{tab:sem_corr}.}
\label{tab:sem_corr_gemma}

\end{table}

\clearpage
\section{Additional Logit Lens Results}
\label{app:logit_lens_extra}

\begin{figure}[h]
    \centering   
    \begin{subfigure}[c]{0.4\textwidth}
        \centering
        \includegraphics[width=\textwidth]{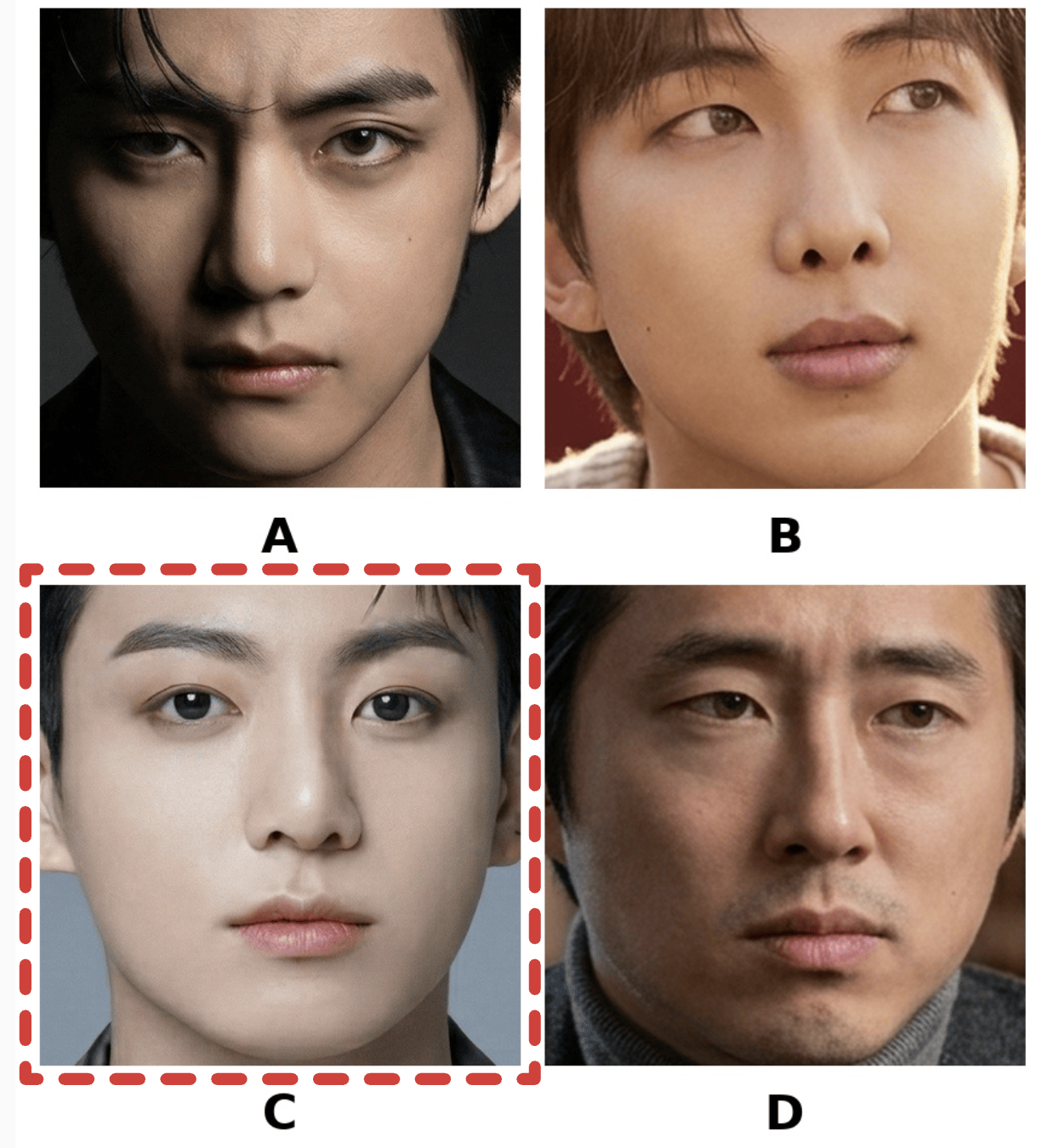}
        \caption{}
    \end{subfigure}
    \hfill
    \begin{subfigure}[c]{0.49\textwidth}
        \centering
        \includegraphics[width=\textwidth]{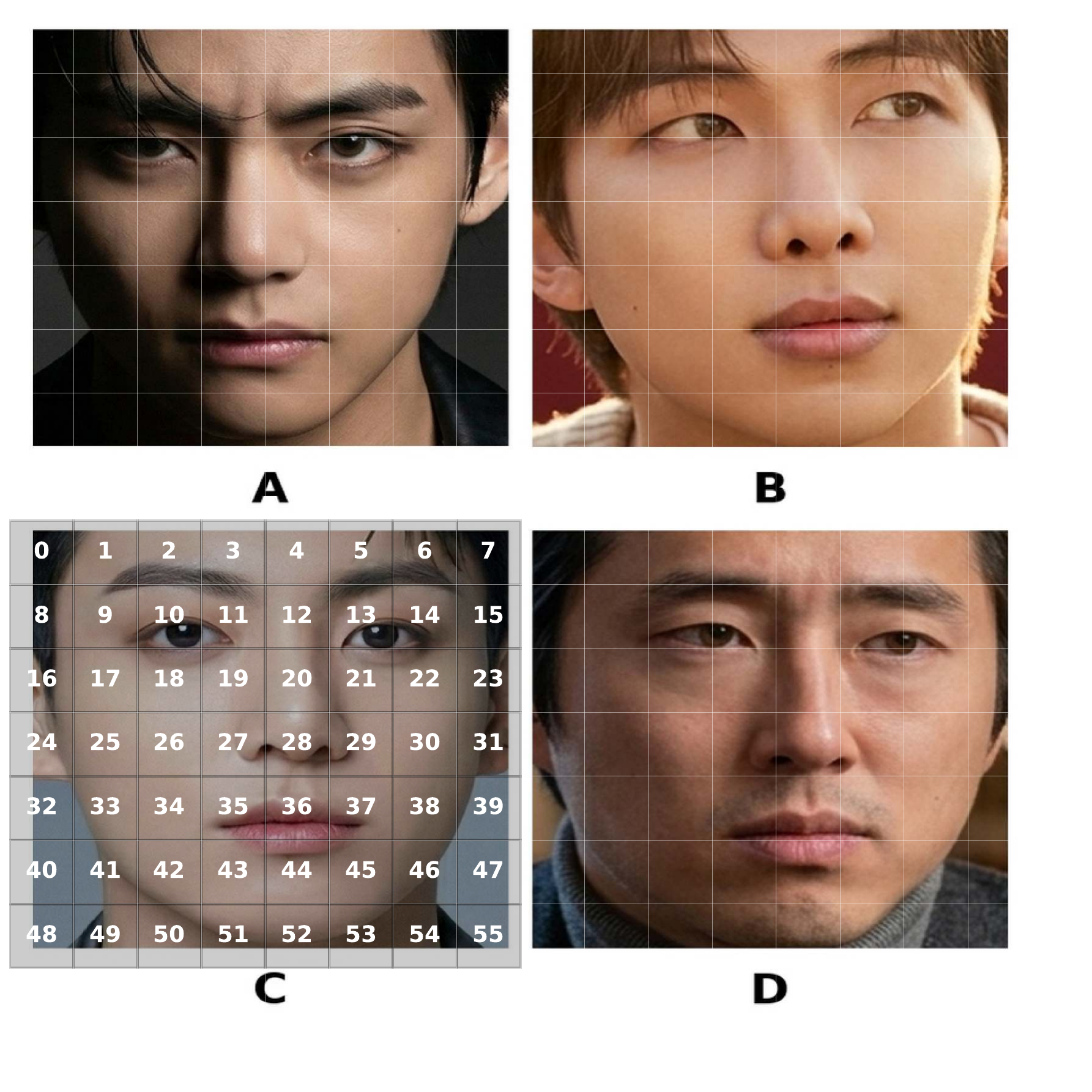}
        \caption{}
    \end{subfigure}
    
    \begin{subfigure}[b]{0.49\textwidth}
        \centering
        \includegraphics[width=\textwidth]{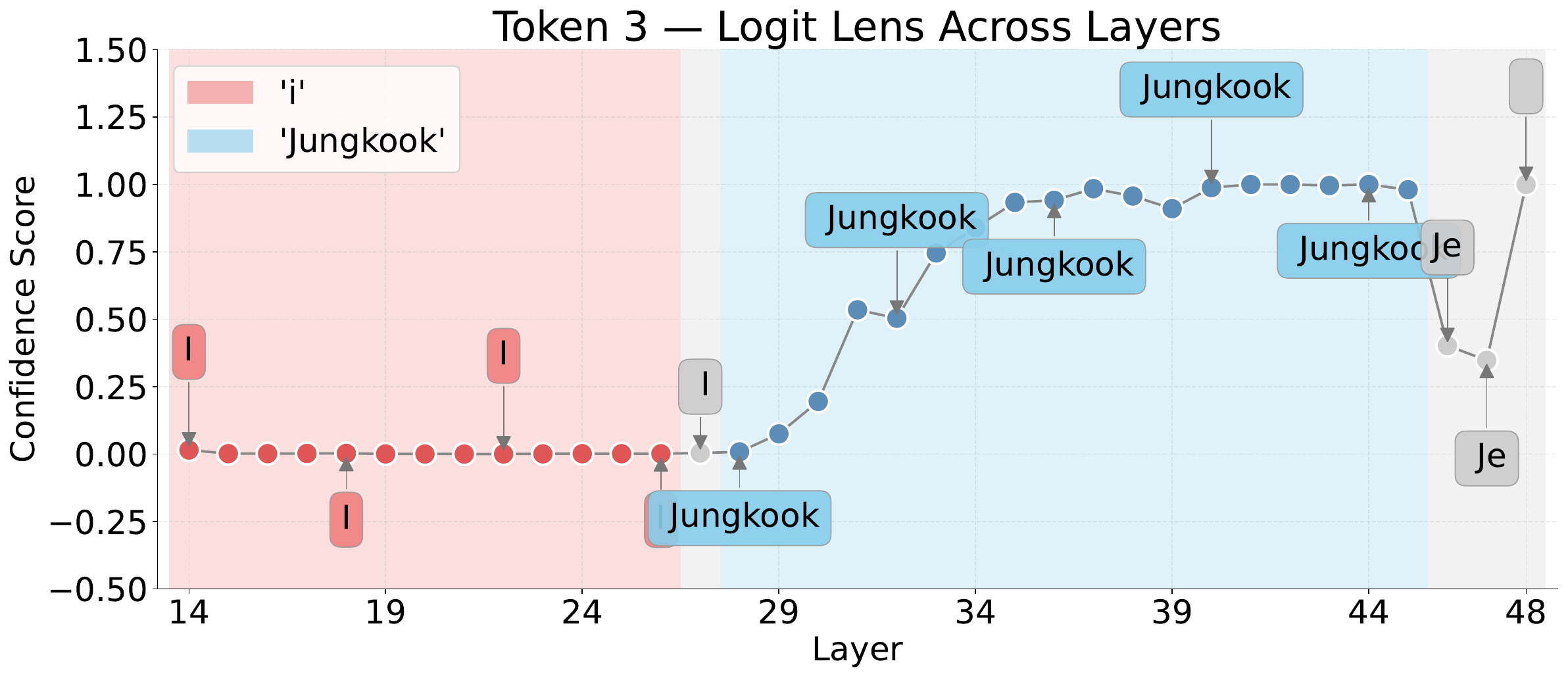}
        \caption{}
    \end{subfigure}
    \hfill
    \begin{subfigure}[b]{0.49\textwidth}
        \centering
        \includegraphics[width=\textwidth]{images/logit_lens_qual/Jungkook_idx34_token5.pdf}
        \caption{}
    \end{subfigure}
    
    \begin{subfigure}[b]{0.49\textwidth}
        \centering
        \includegraphics[width=\textwidth]{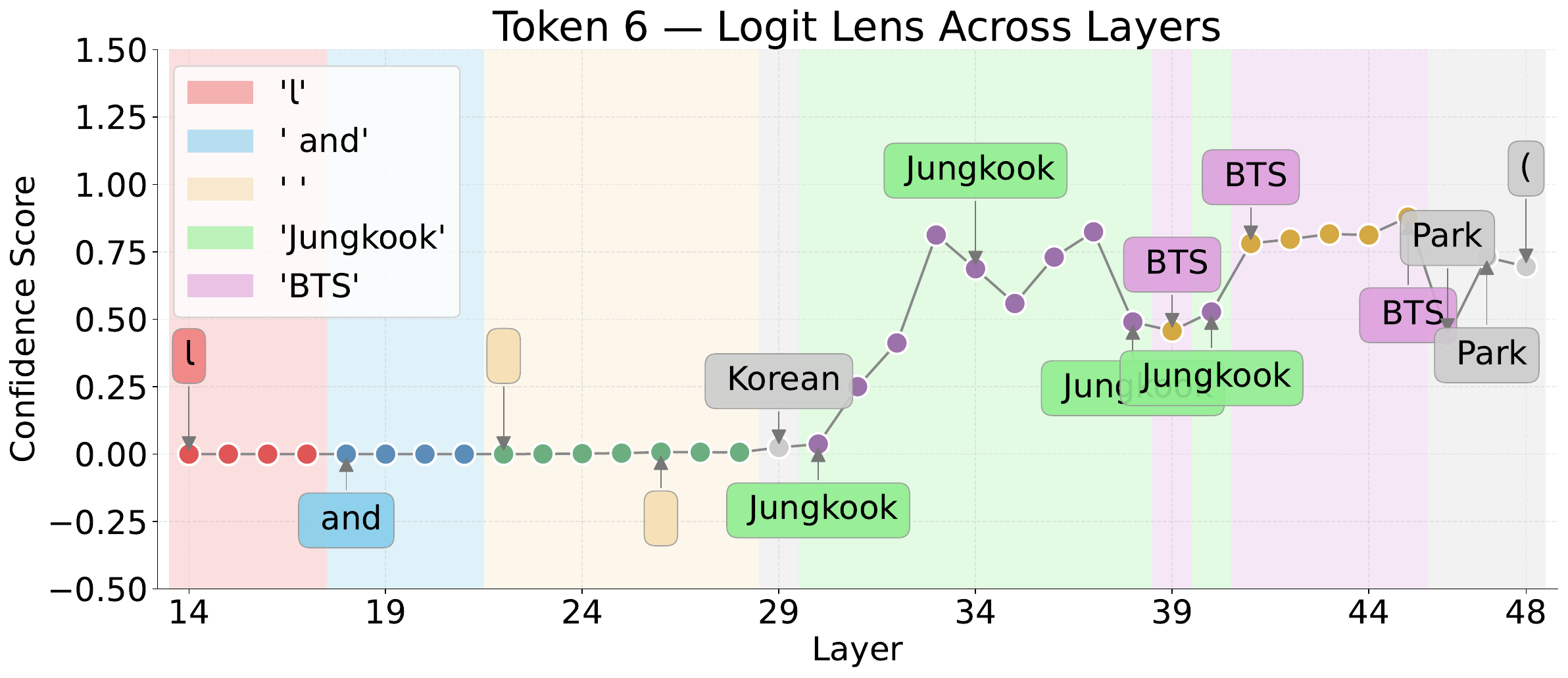}
        \caption{}
    \end{subfigure}
    \hfill
    \begin{subfigure}[b]{0.49\textwidth}
        \centering
        \includegraphics[width=\textwidth]{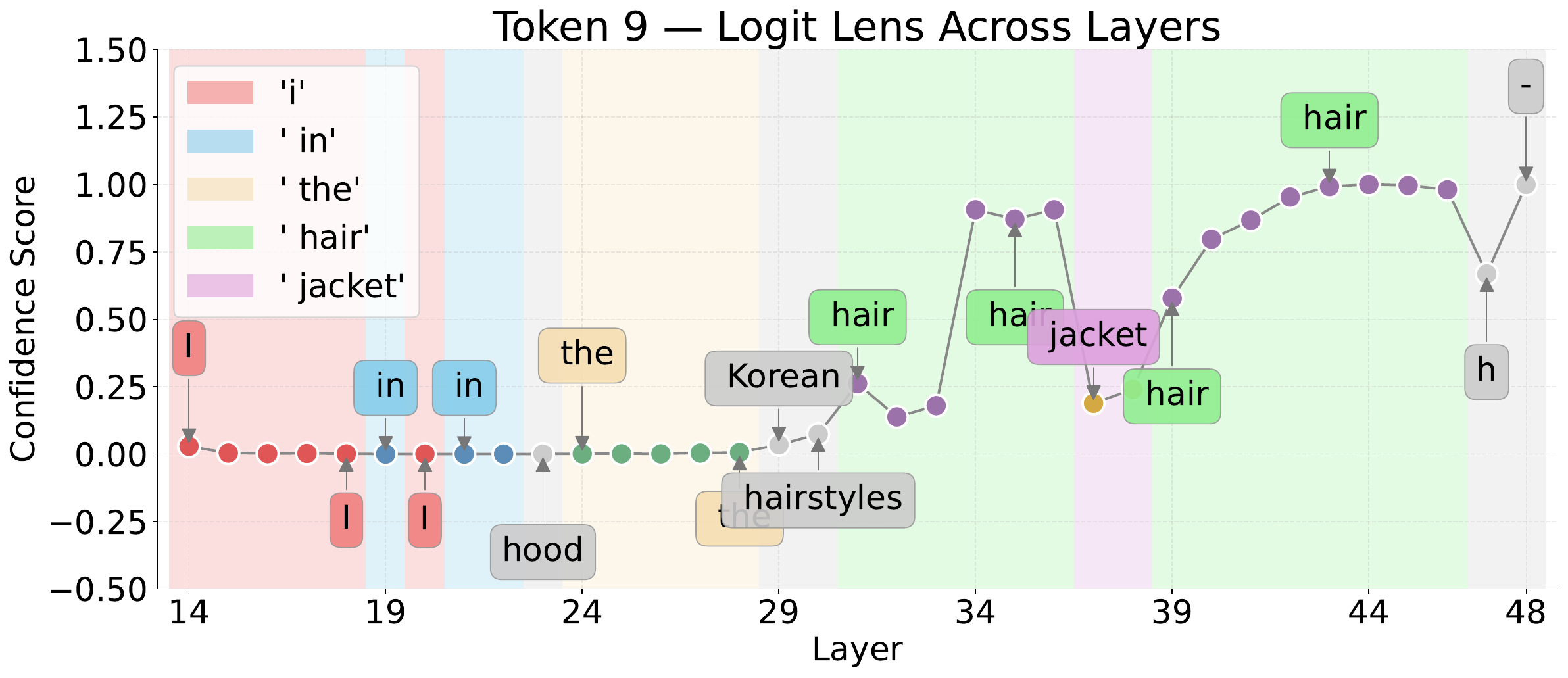}
        \caption{}
        \label{fig:logit_lens_jk_hair}
    \end{subfigure}
    
    \begin{subfigure}[b]{0.49\textwidth}
        \centering
        \includegraphics[width=\textwidth]{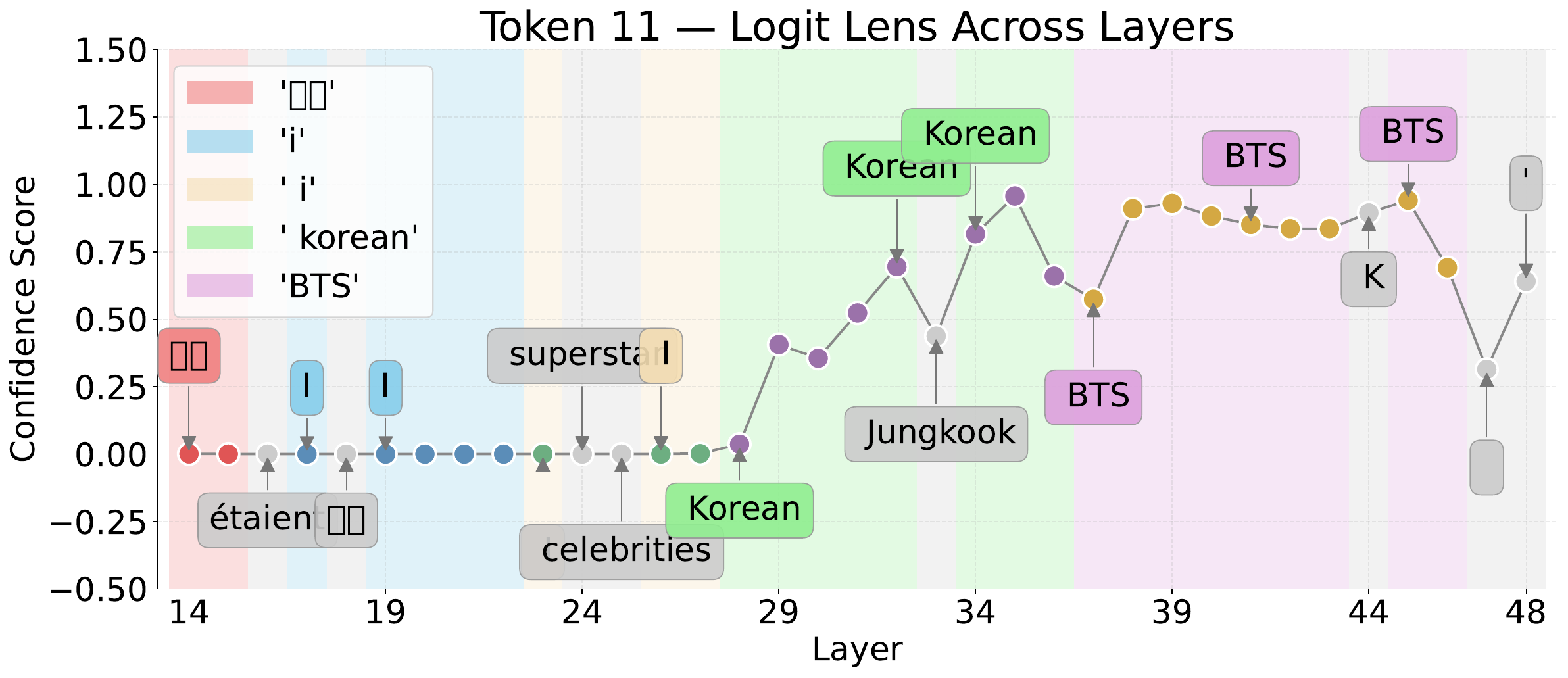}
        \caption{}
    \end{subfigure}
    \hfill
    \begin{subfigure}[b]{0.49\textwidth}
        \centering
        \includegraphics[width=\textwidth]{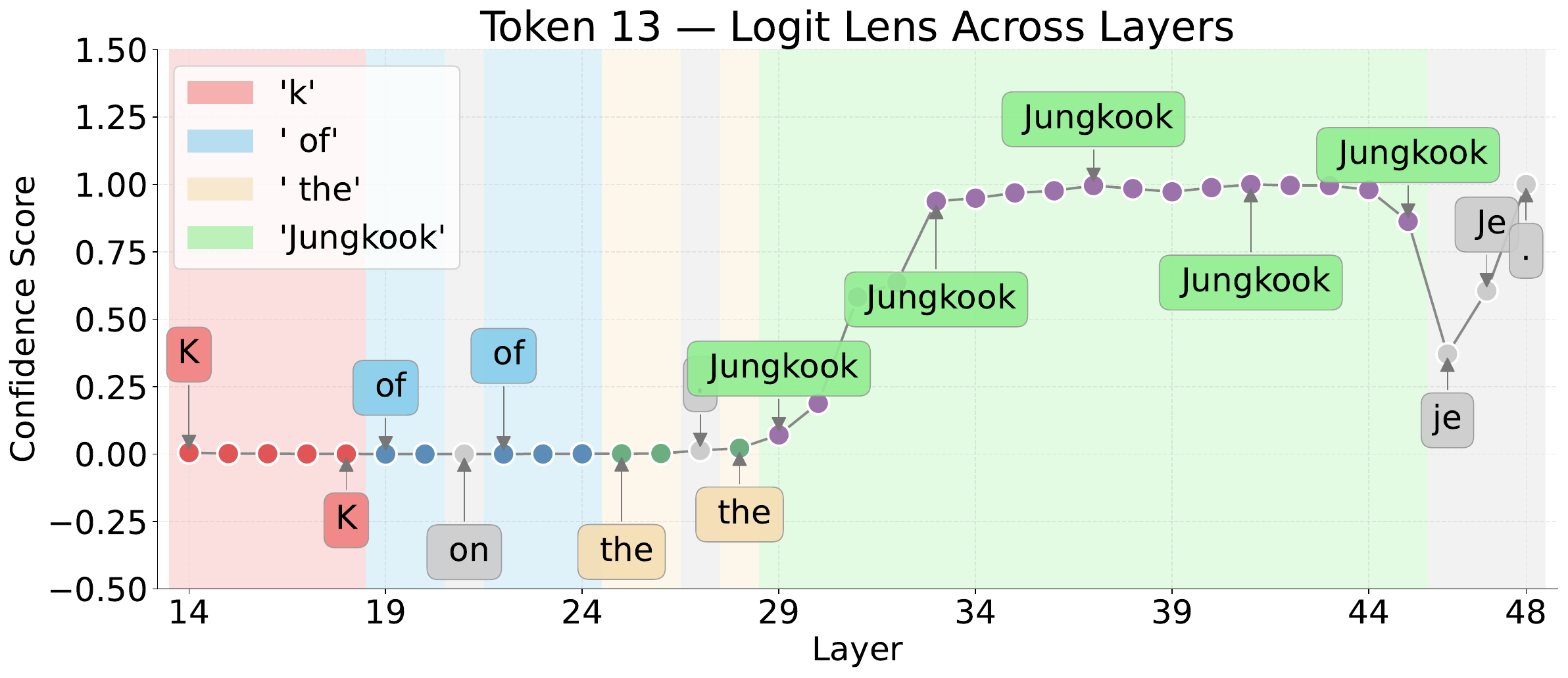}
        \caption{}
    \end{subfigure}
    \caption{Qualitative Logit Lens example from Gemma-3-12B. The results show that some decoded tokens are highly informative of the person's identity and affiliations, while Fig. \ref{fig:logit_lens_jk_hair} shows an example of a visual token that encodes generic semantic information. }
\end{figure}

\begin{figure}[h]
    \centering   
    \begin{subfigure}[b]{0.33\textwidth}
        \centering
        \includegraphics[width=\textwidth]{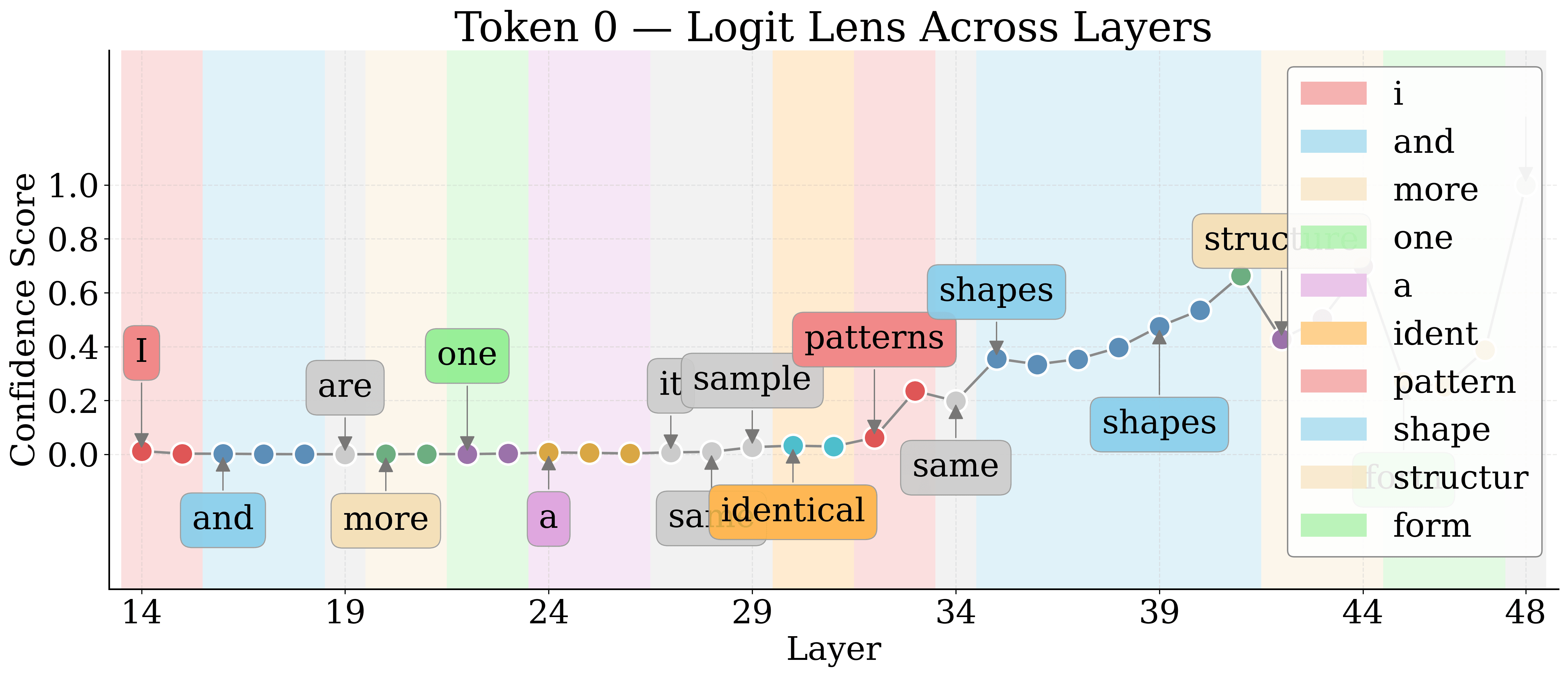}
    \end{subfigure}
    \hfill
    \begin{subfigure}[b]{0.32\textwidth}
        \centering
        \includegraphics[width=\textwidth]{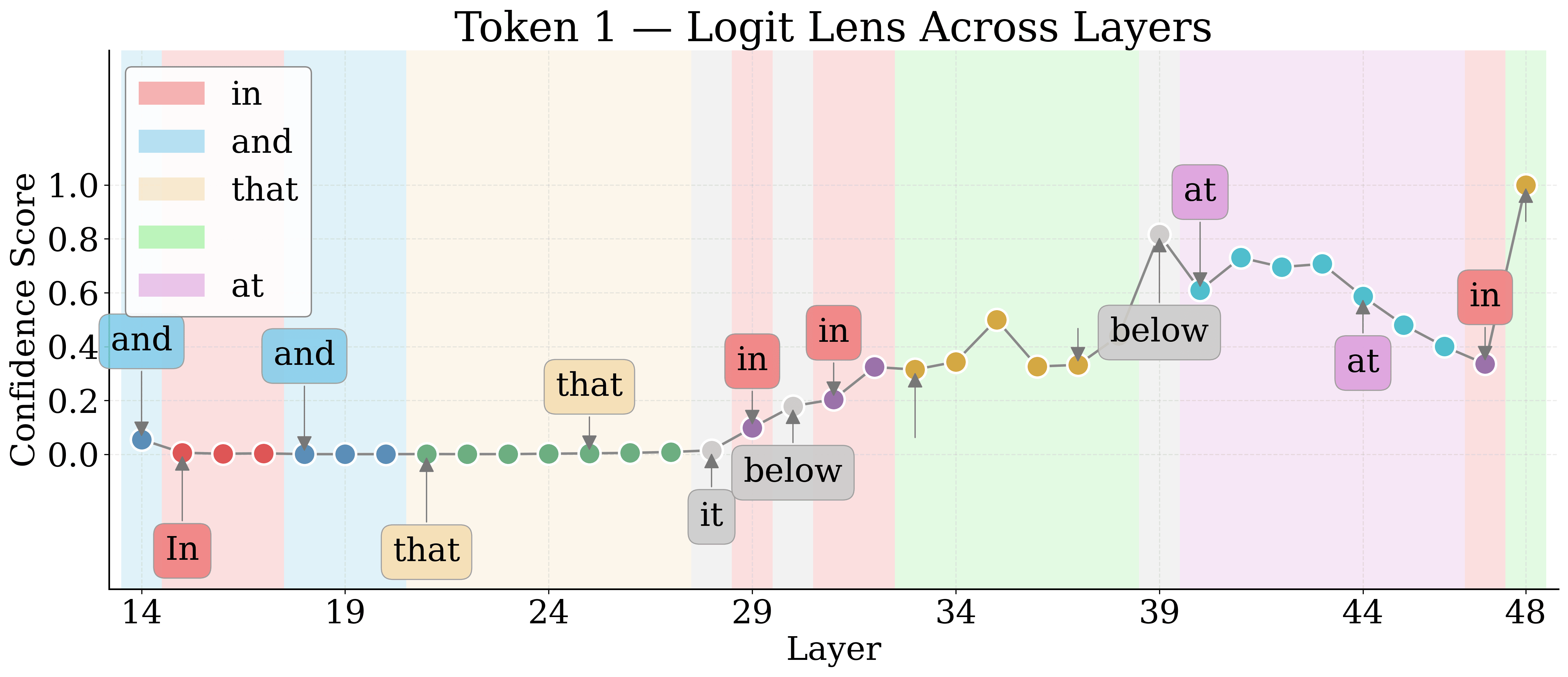}
    \end{subfigure}
    \hfill
    \begin{subfigure}[b]{0.33\textwidth}
        \centering
        \includegraphics[width=\textwidth]{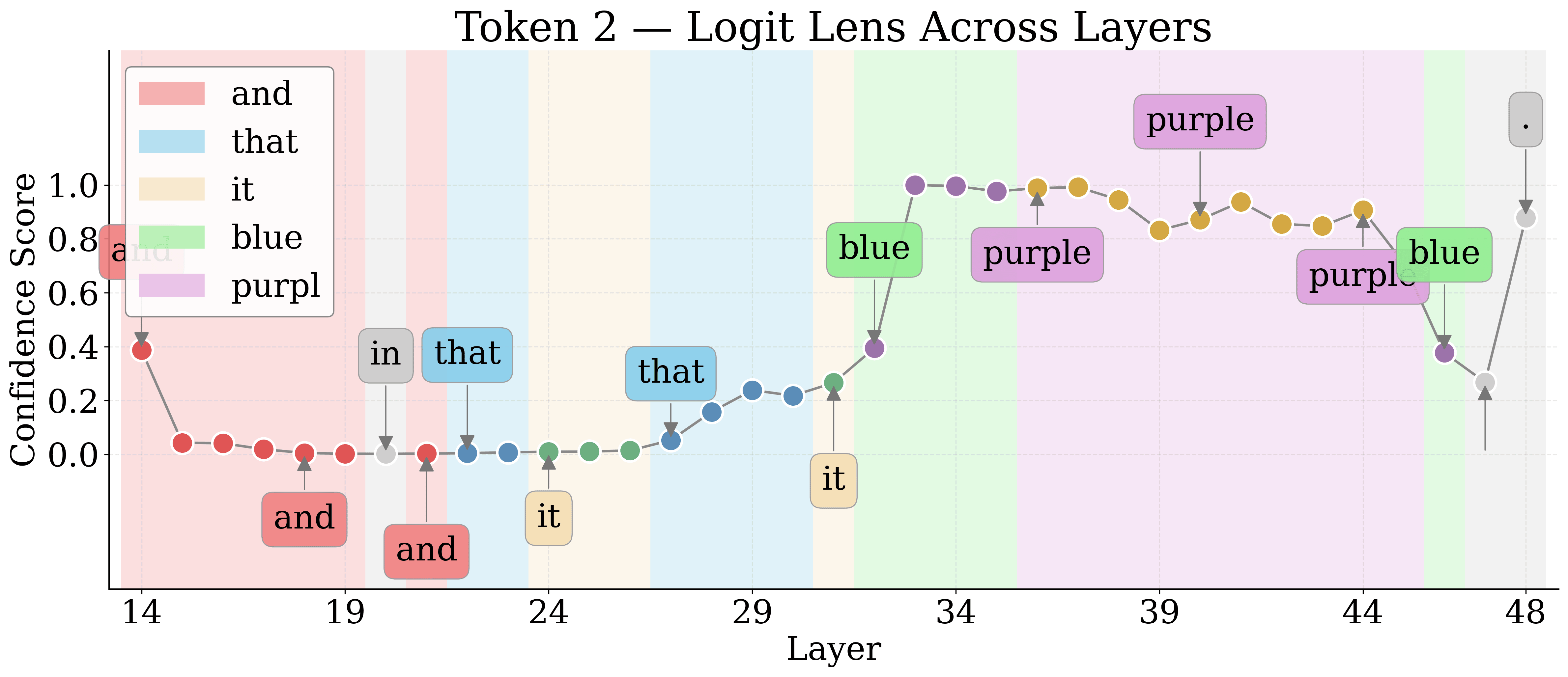}
    \end{subfigure}

    \begin{subfigure}[b]{0.33\textwidth}
        \centering
        \includegraphics[width=\textwidth]{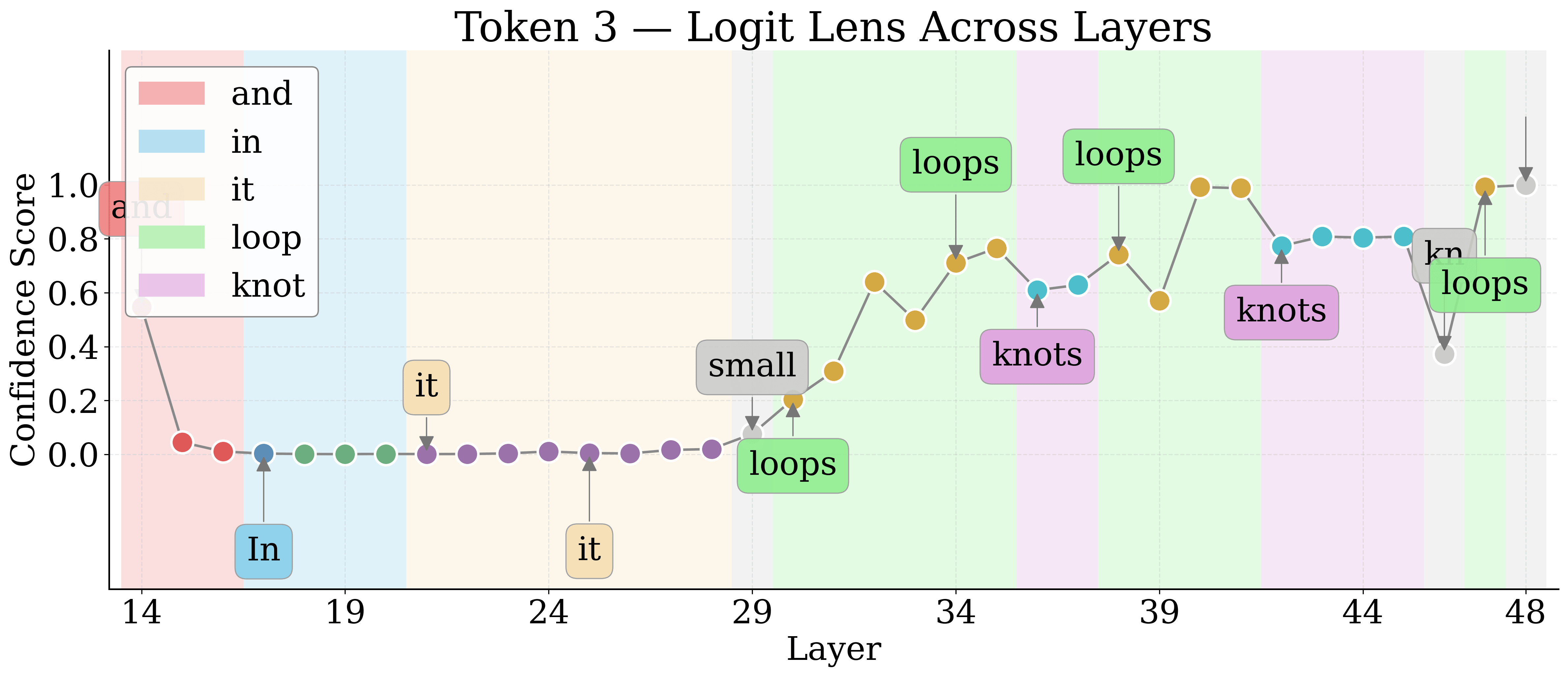}
    \end{subfigure}
    \hfill
    \begin{subfigure}[b]{0.32\textwidth}
        \centering
        \includegraphics[width=\textwidth]{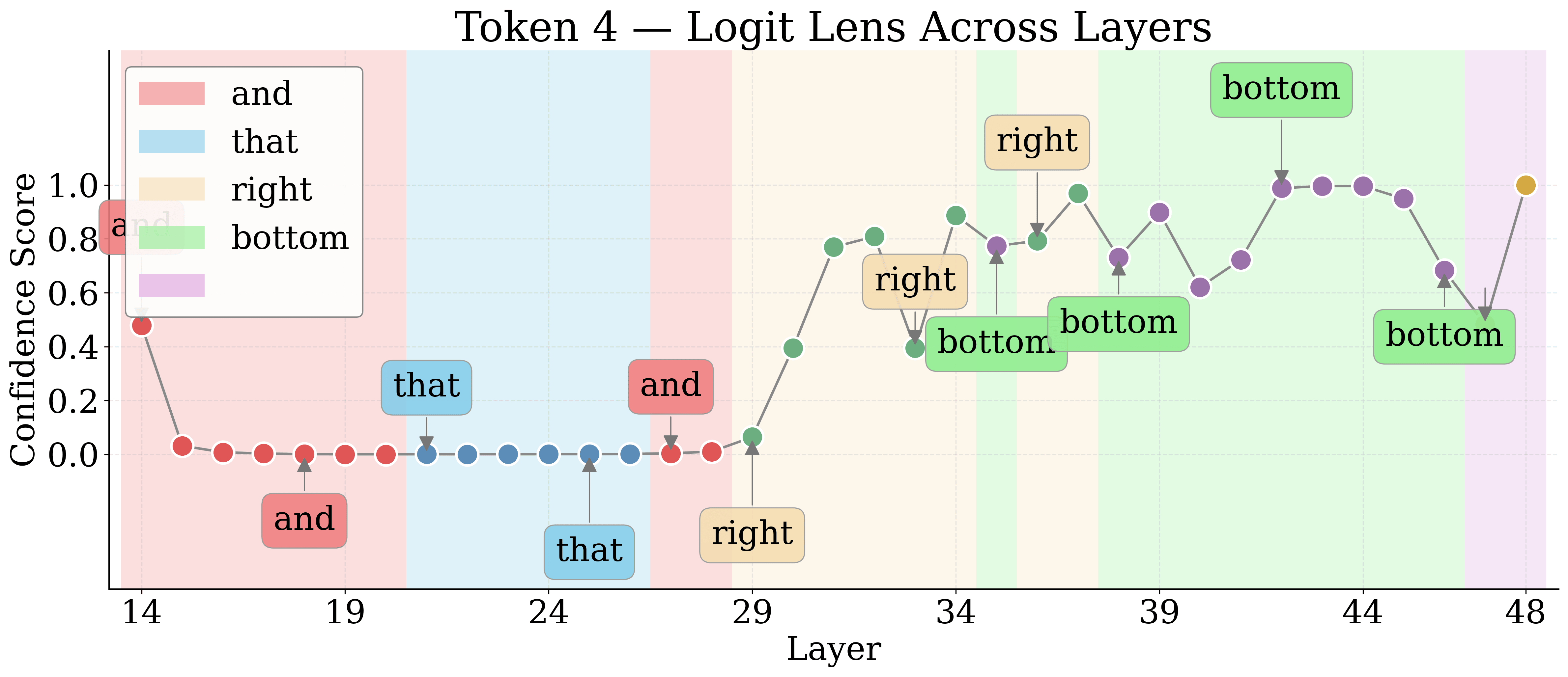}
    \end{subfigure}
    \hfill
    \begin{subfigure}[b]{0.33\textwidth}
        \centering
        \includegraphics[width=\textwidth]{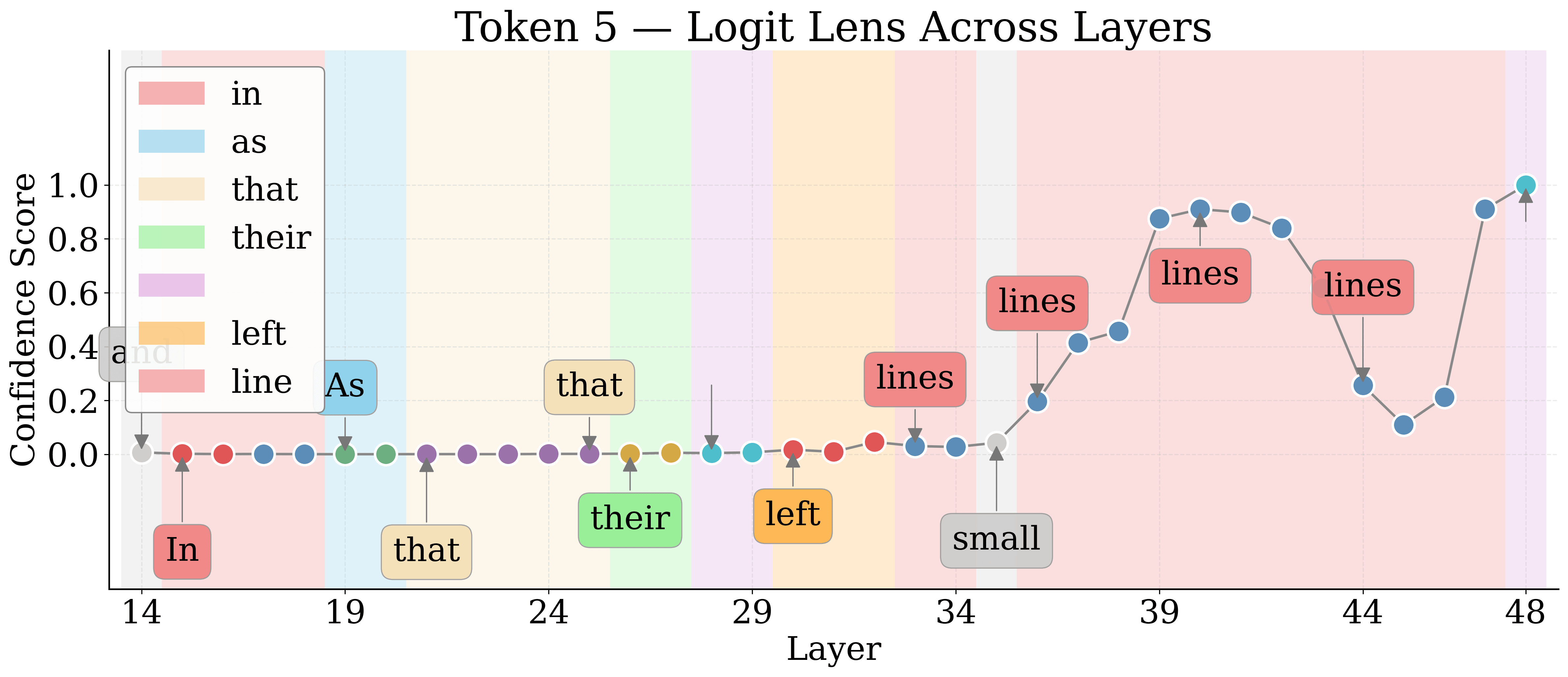}
    \end{subfigure}

        \begin{subfigure}[b]{0.33\textwidth}
        \centering
        \includegraphics[width=\textwidth]{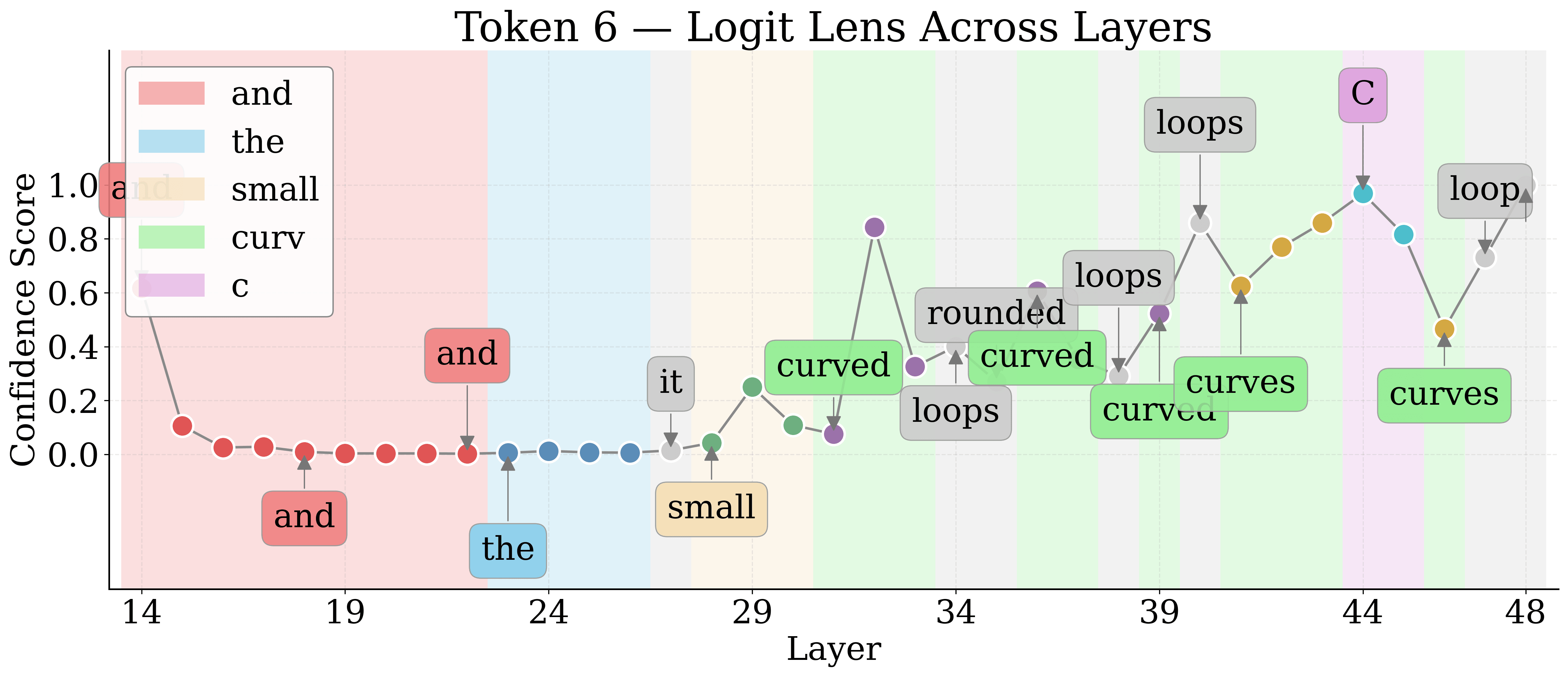}
    \end{subfigure}
    \hfill
    \begin{subfigure}[b]{0.32\textwidth}
        \centering
        \includegraphics[width=\textwidth]{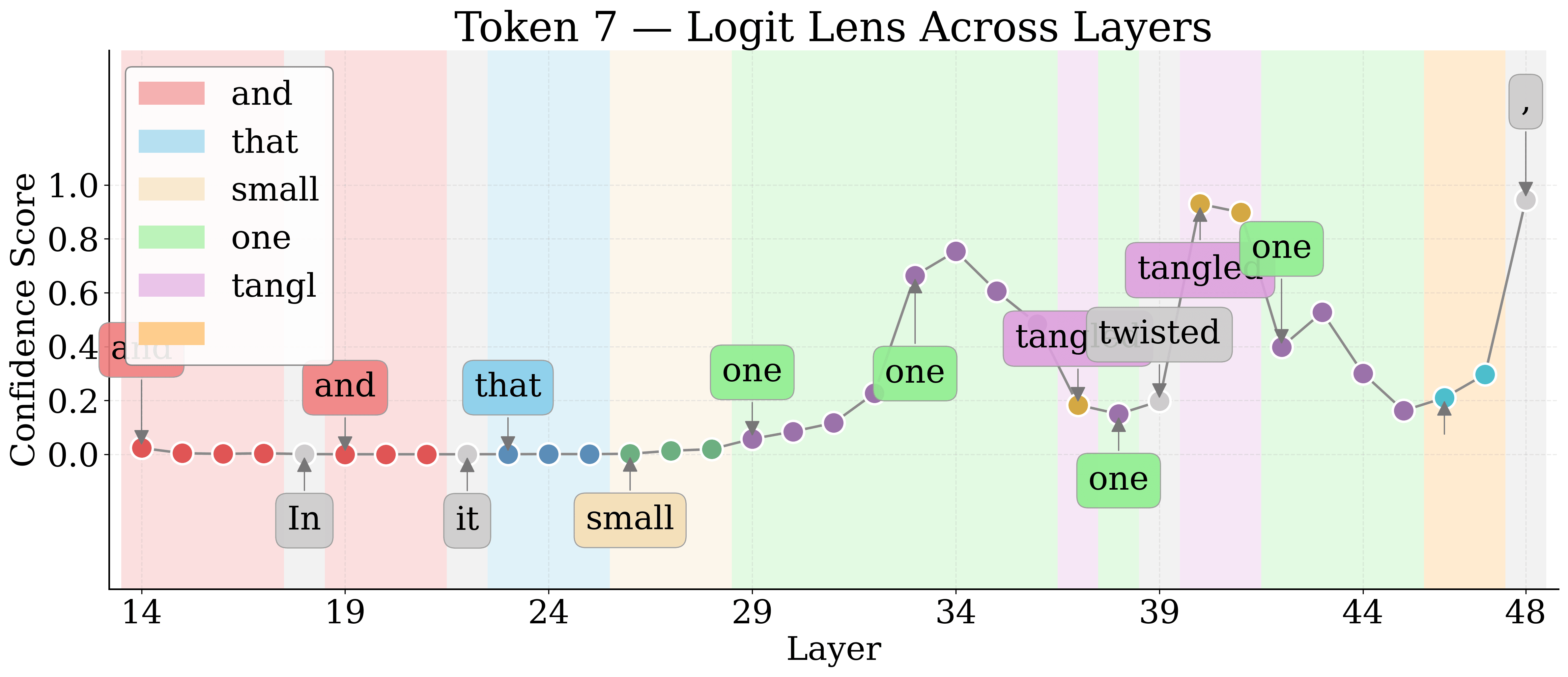}
    \end{subfigure}
    \hfill
    \begin{subfigure}[b]{0.33\textwidth}
        \centering
        \includegraphics[width=\textwidth]{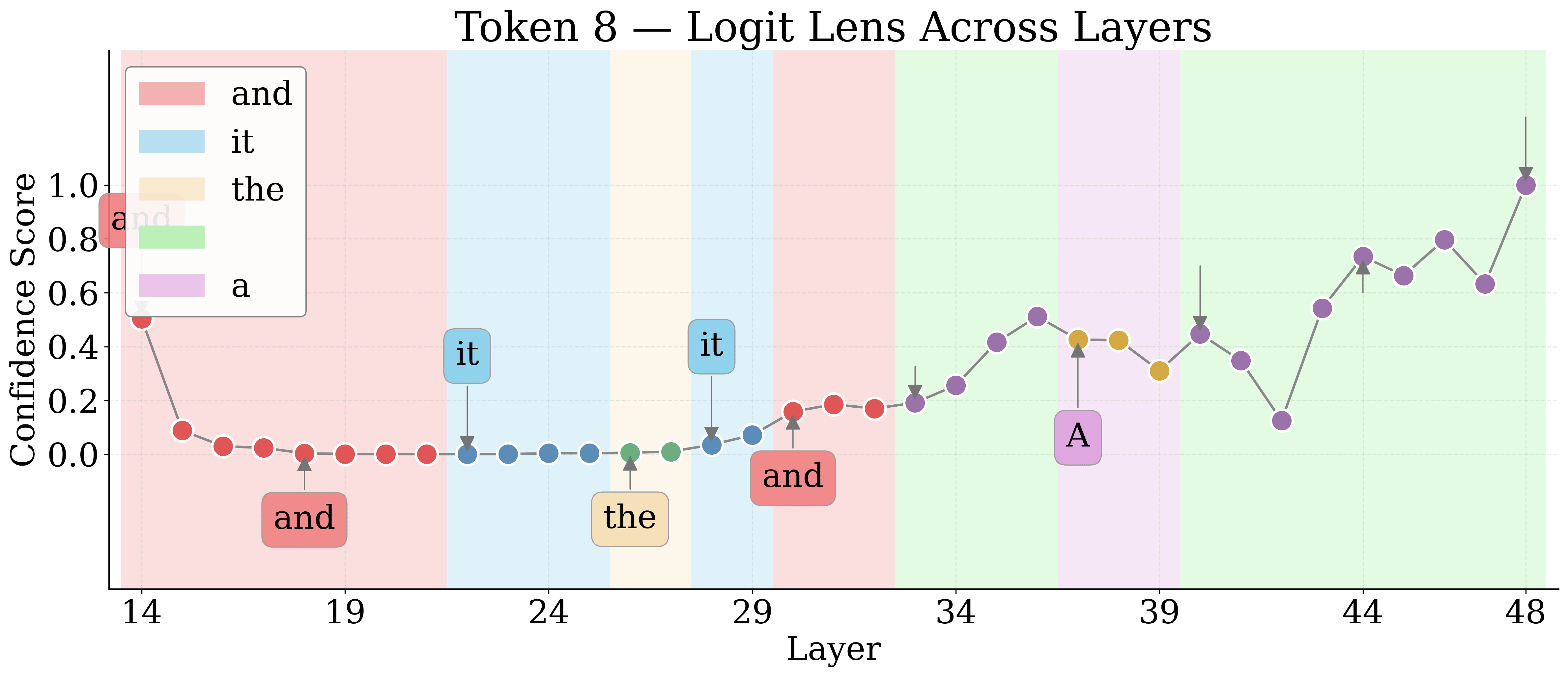}
    \end{subfigure}
        \begin{subfigure}[b]{0.33\textwidth}
        \centering
        \includegraphics[width=\textwidth]{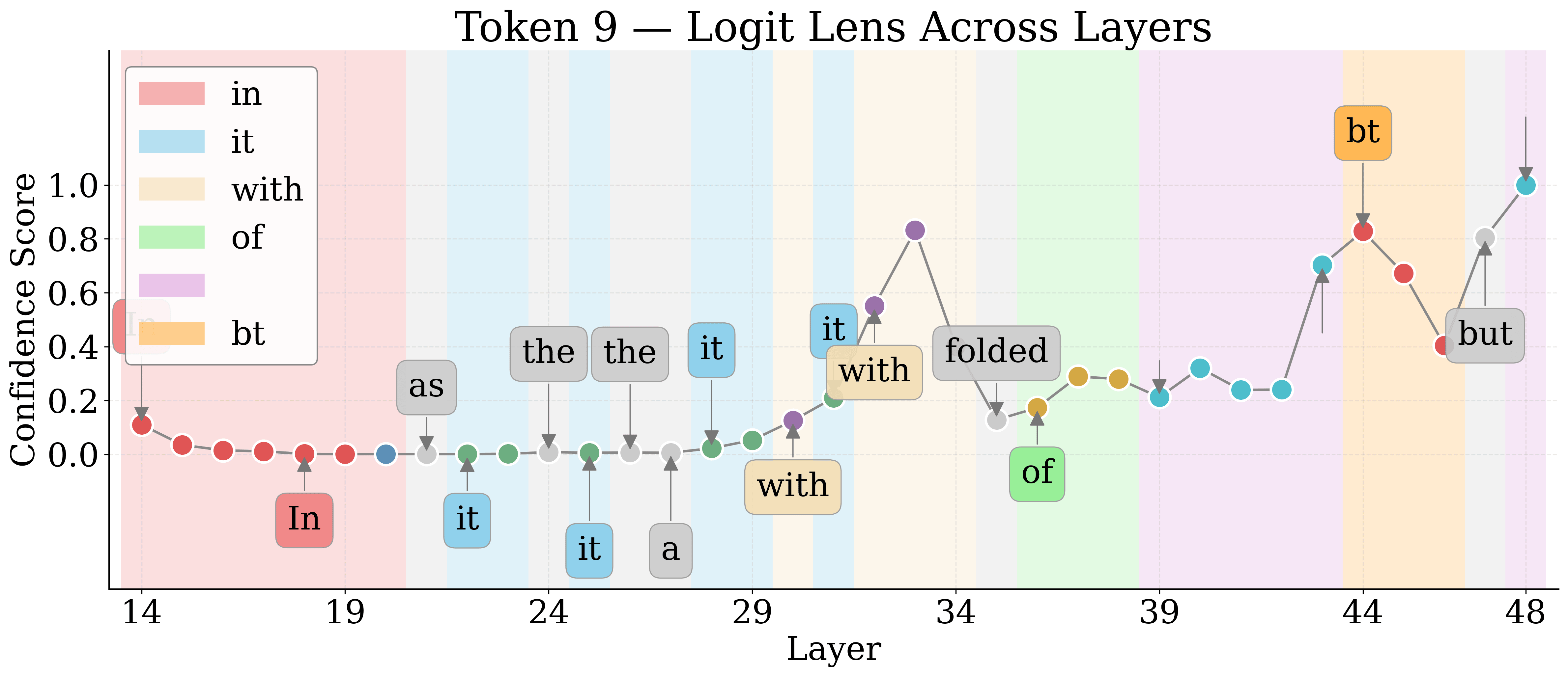}
    \end{subfigure}
    \hfill
    \begin{subfigure}[b]{0.32\textwidth}
        \centering
        \includegraphics[width=\textwidth]{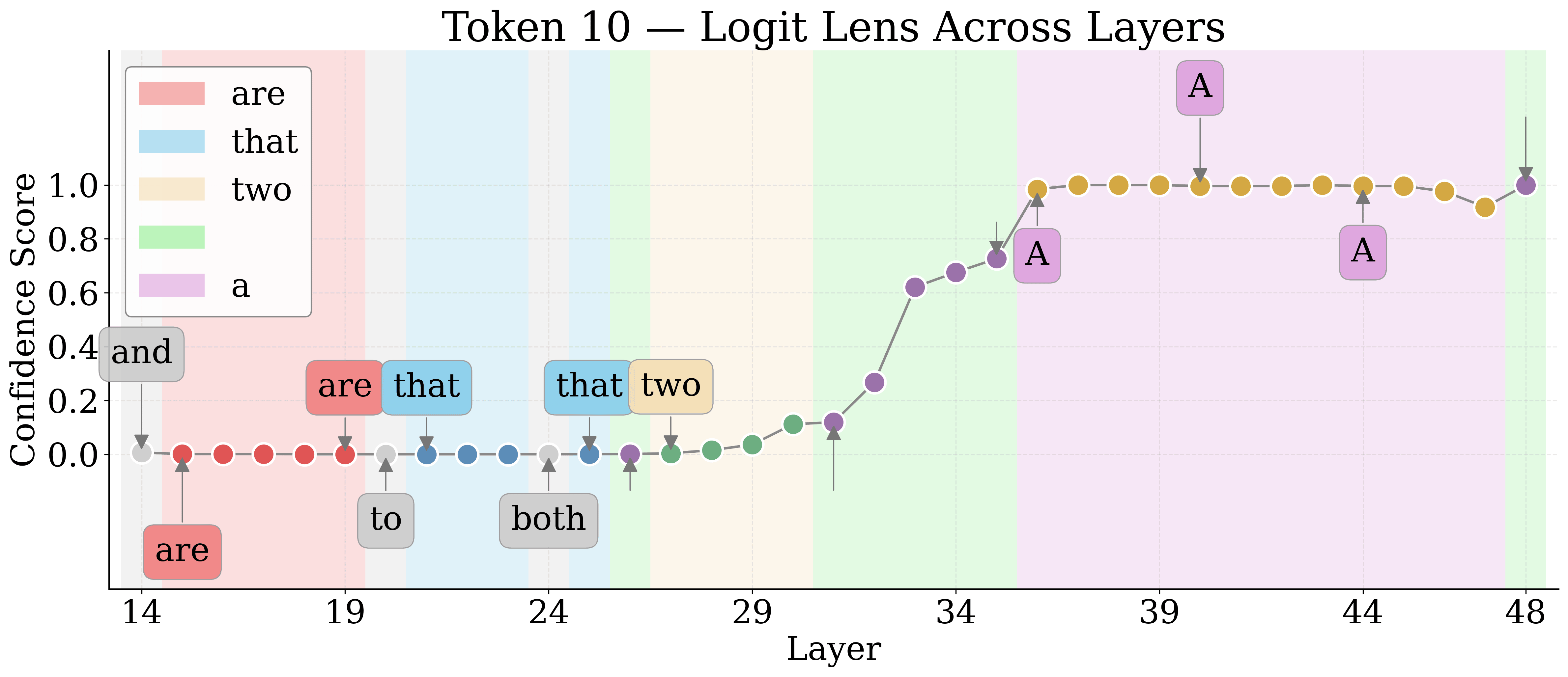}
    \end{subfigure}
    \hfill
    \begin{subfigure}[b]{0.33\textwidth}
        \centering
        \includegraphics[width=\textwidth]{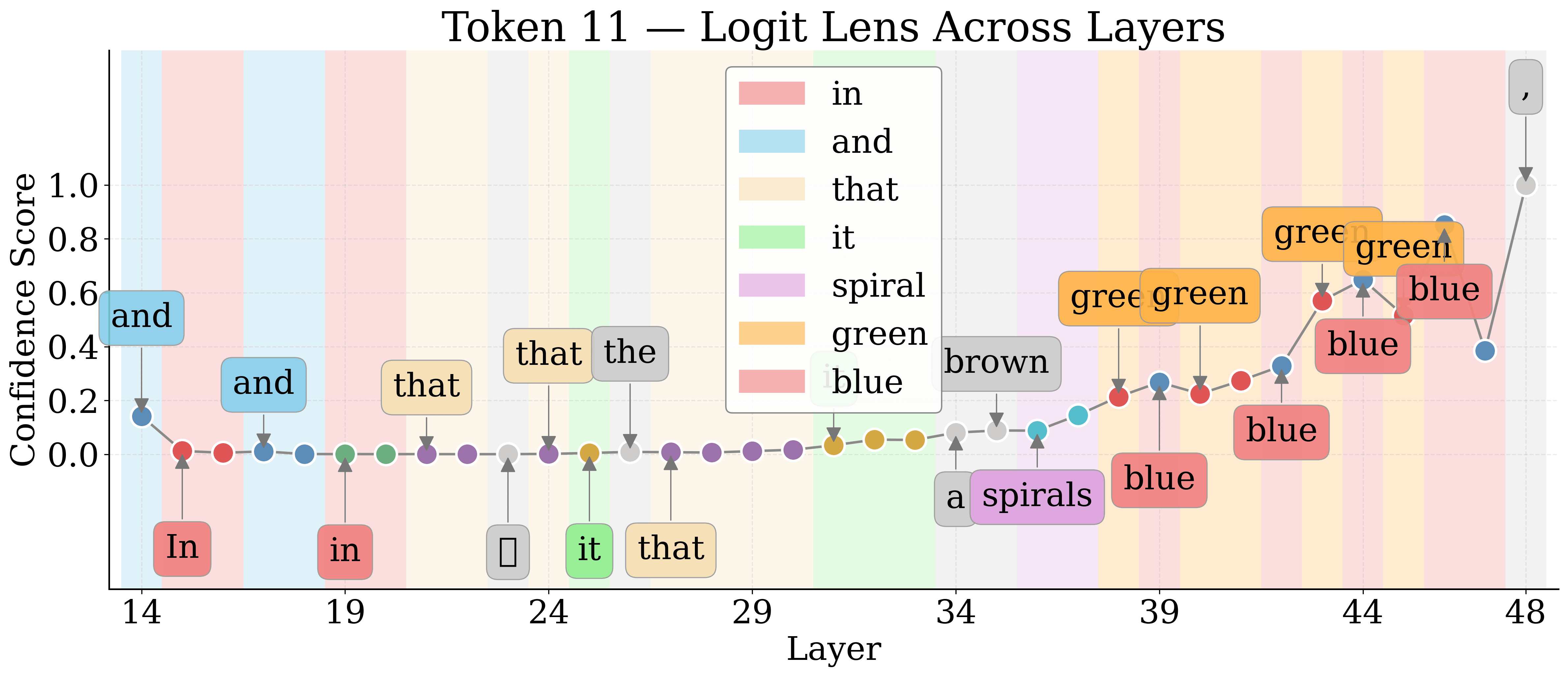}
    \end{subfigure}

        \begin{subfigure}[b]{0.33\textwidth}
        \centering
        \includegraphics[width=\textwidth]{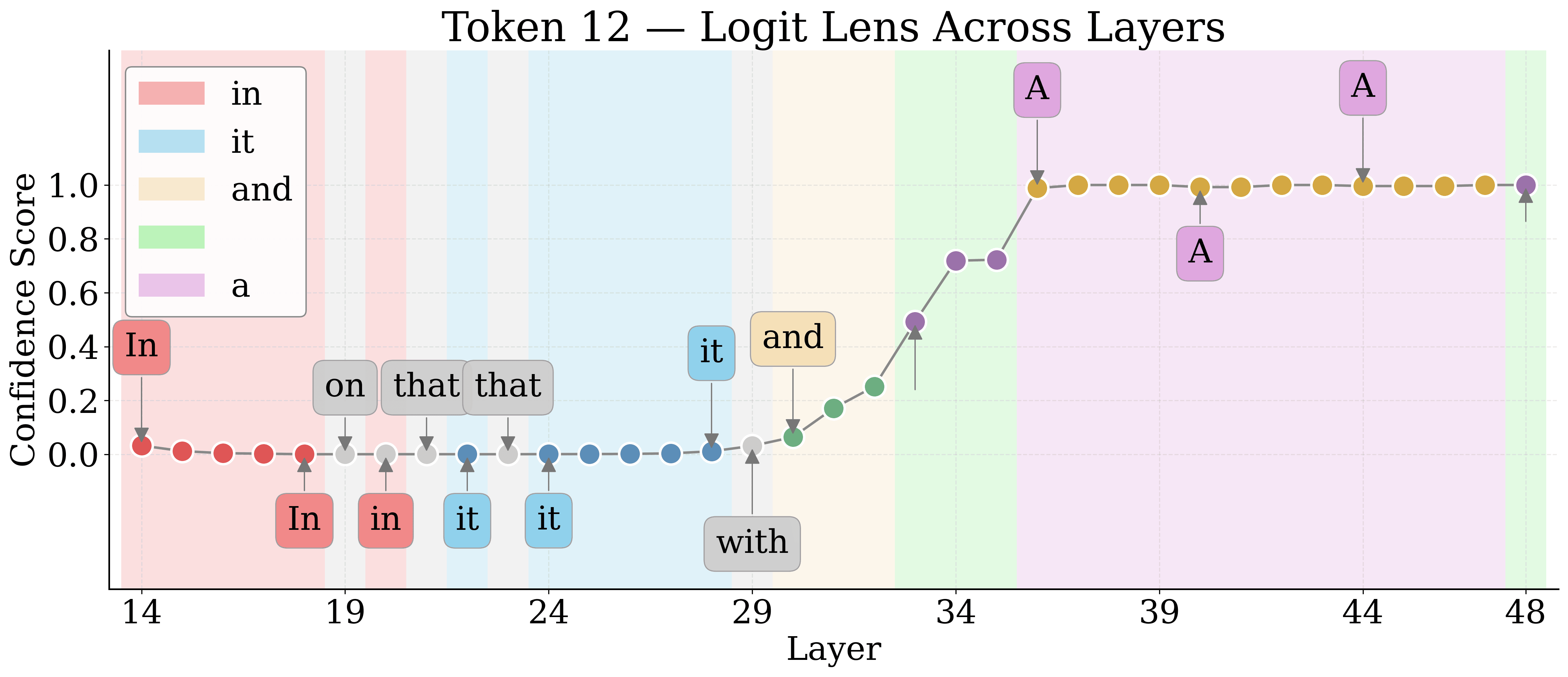}
    \end{subfigure}
    \hfill
    \begin{subfigure}[b]{0.32\textwidth}
        \centering
        \includegraphics[width=\textwidth]{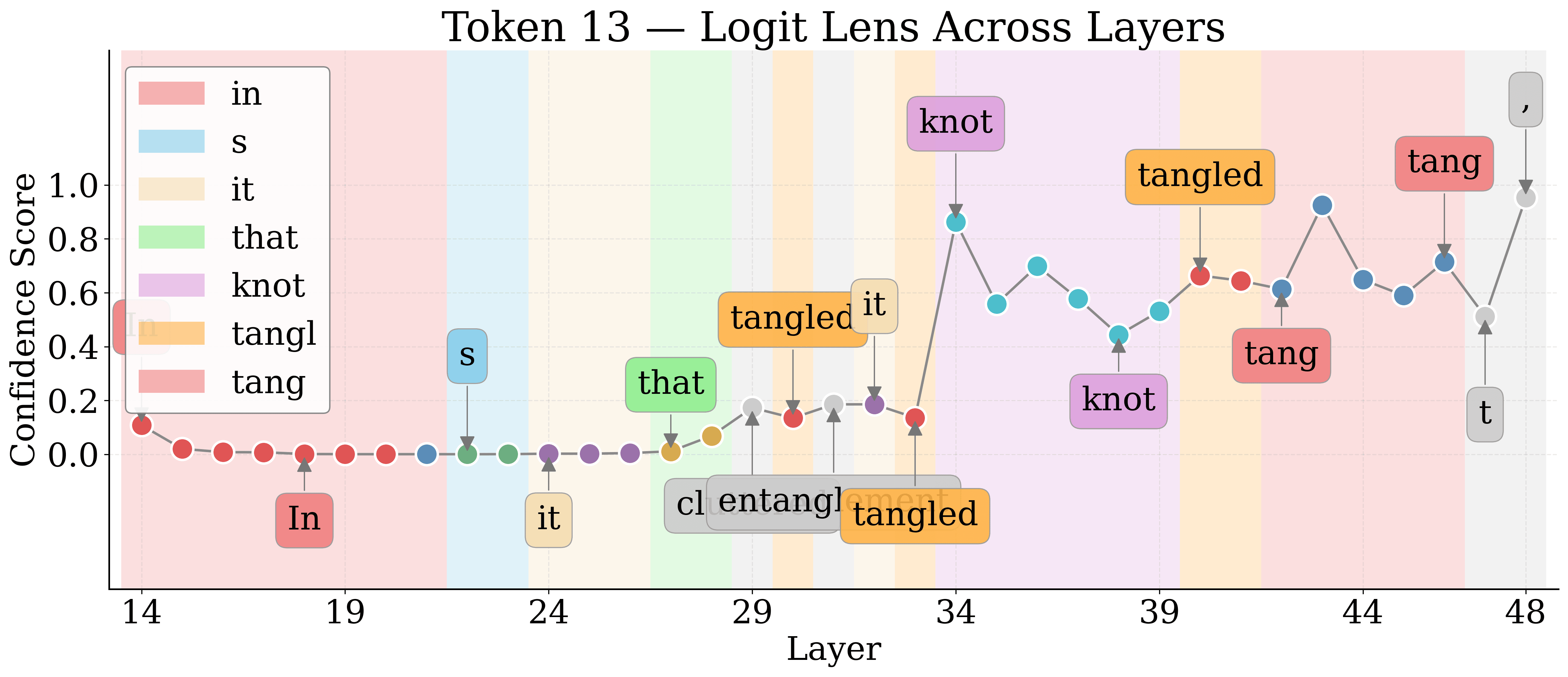}
    \end{subfigure}
    \hfill
    \begin{subfigure}[b]{0.33\textwidth}
        \centering
        \includegraphics[width=\textwidth]{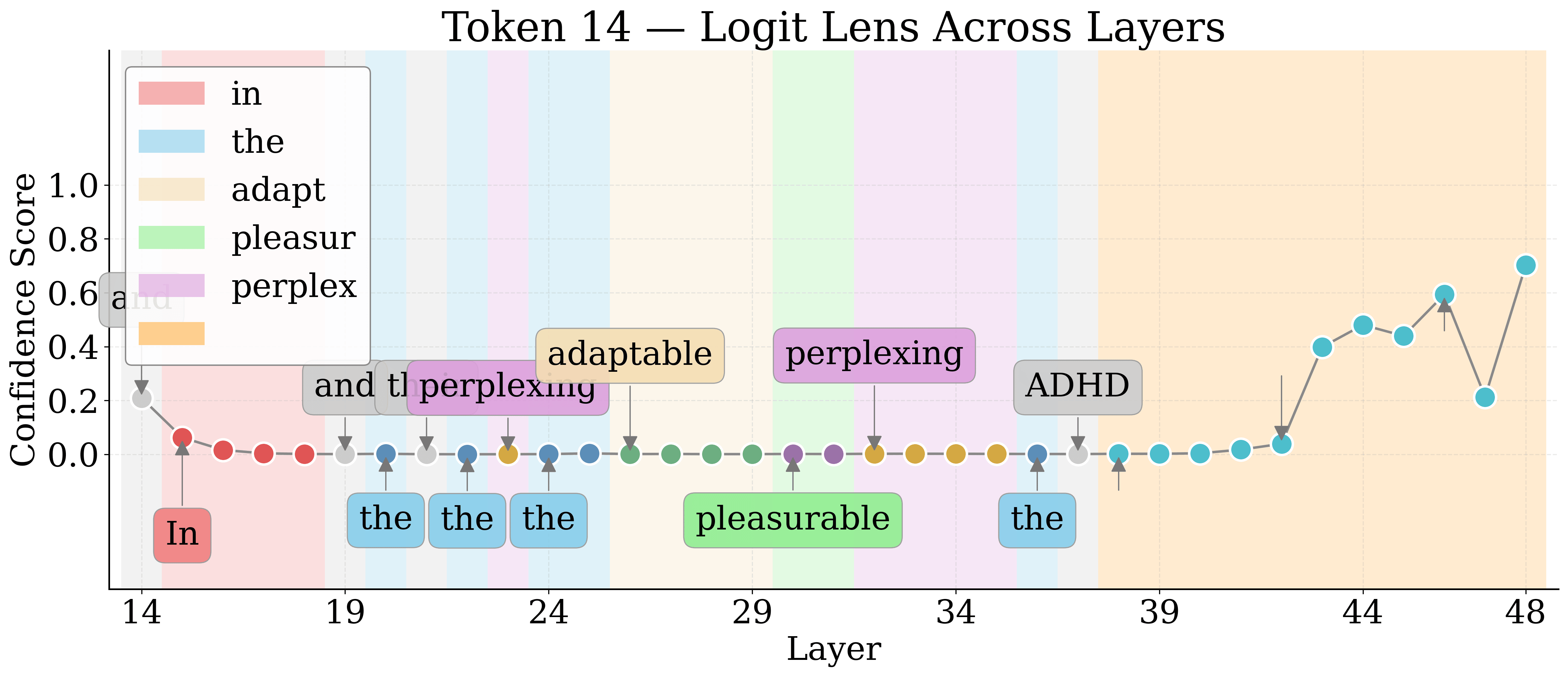}
    \end{subfigure}
    \caption{All Gemma3-12B Logit Lens tokens for  \protect\squiggleO (Unknown Shape 1).}
\end{figure}

\begin{figure}[h]
    \centering   
    \begin{subfigure}[b]{0.33\textwidth}
        \centering
        \includegraphics[width=\textwidth]{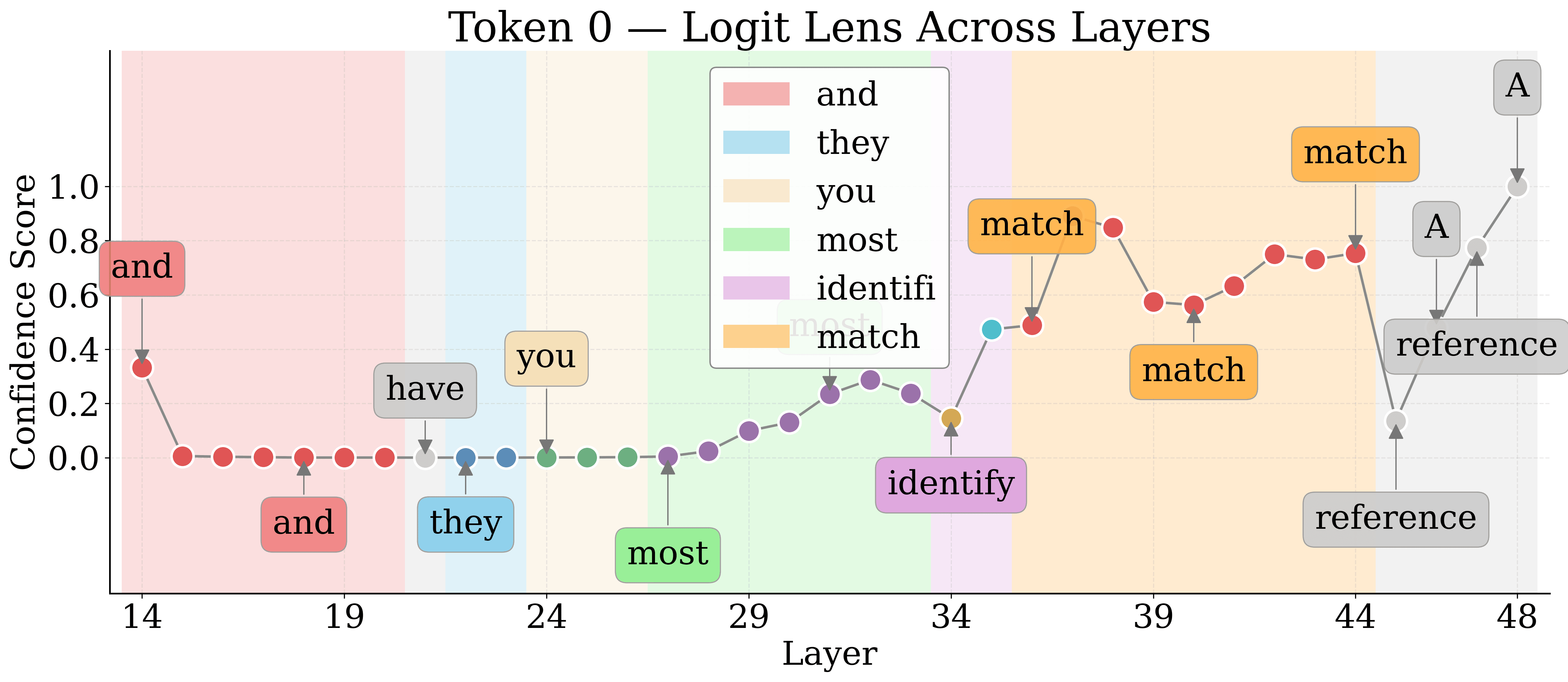}
    \end{subfigure}
    \hfill
    \begin{subfigure}[b]{0.32\textwidth}
        \centering
        \includegraphics[width=\textwidth]{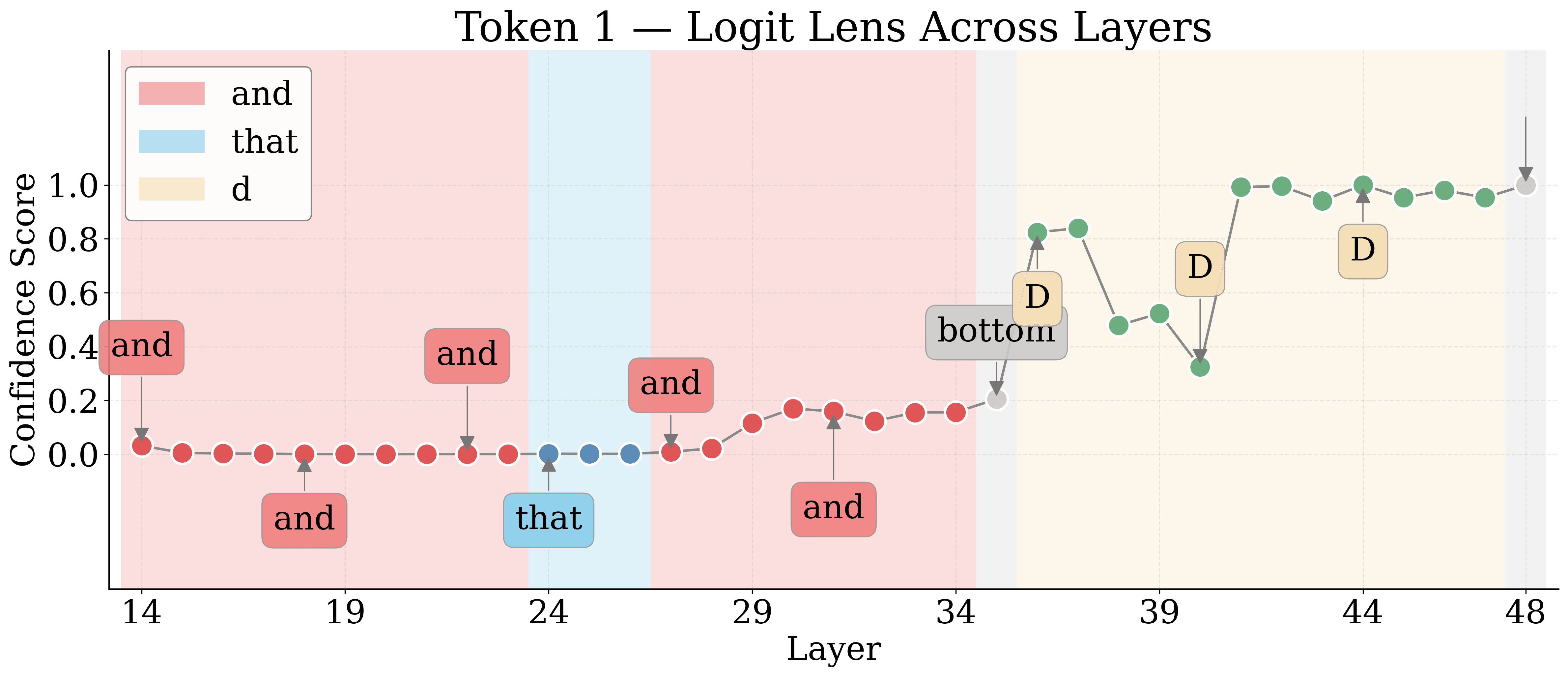}
    \end{subfigure}
    \hfill
    \begin{subfigure}[b]{0.33\textwidth}
        \centering
        \includegraphics[width=\textwidth]{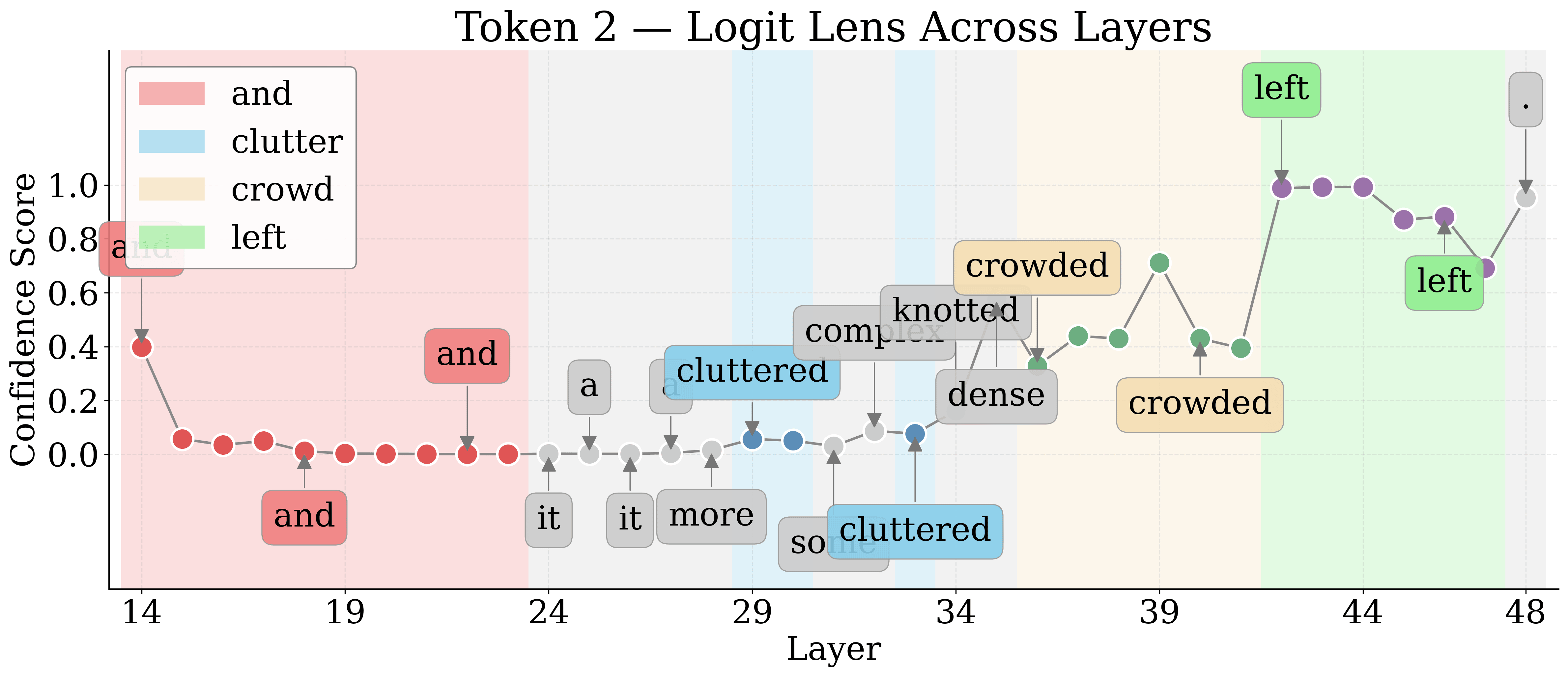}
    \end{subfigure}

    \begin{subfigure}[b]{0.33\textwidth}
        \centering
        \includegraphics[width=\textwidth]{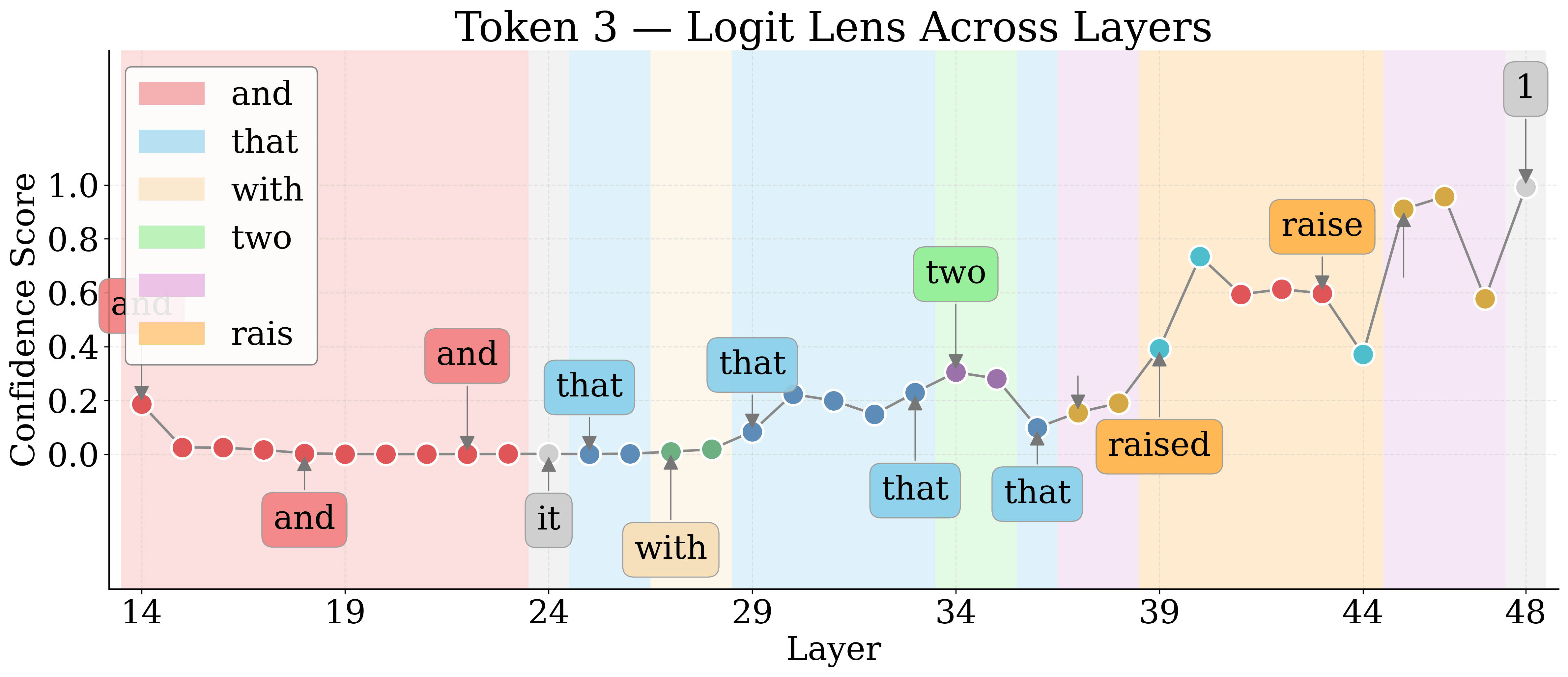}
    \end{subfigure}
    \hfill
    \begin{subfigure}[b]{0.32\textwidth}
        \centering
        \includegraphics[width=\textwidth]{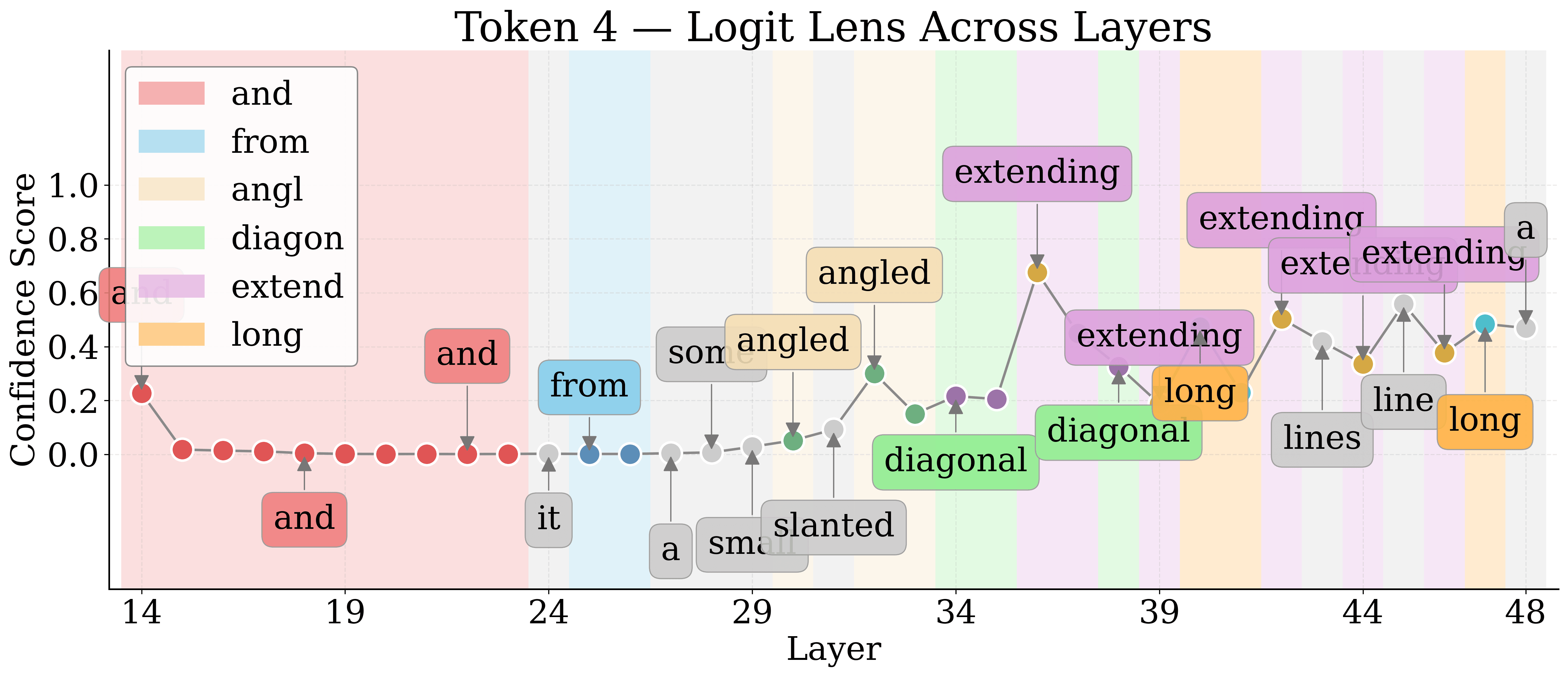}
    \end{subfigure}
    \hfill
    \begin{subfigure}[b]{0.33\textwidth}
        \centering
        \includegraphics[width=\textwidth]{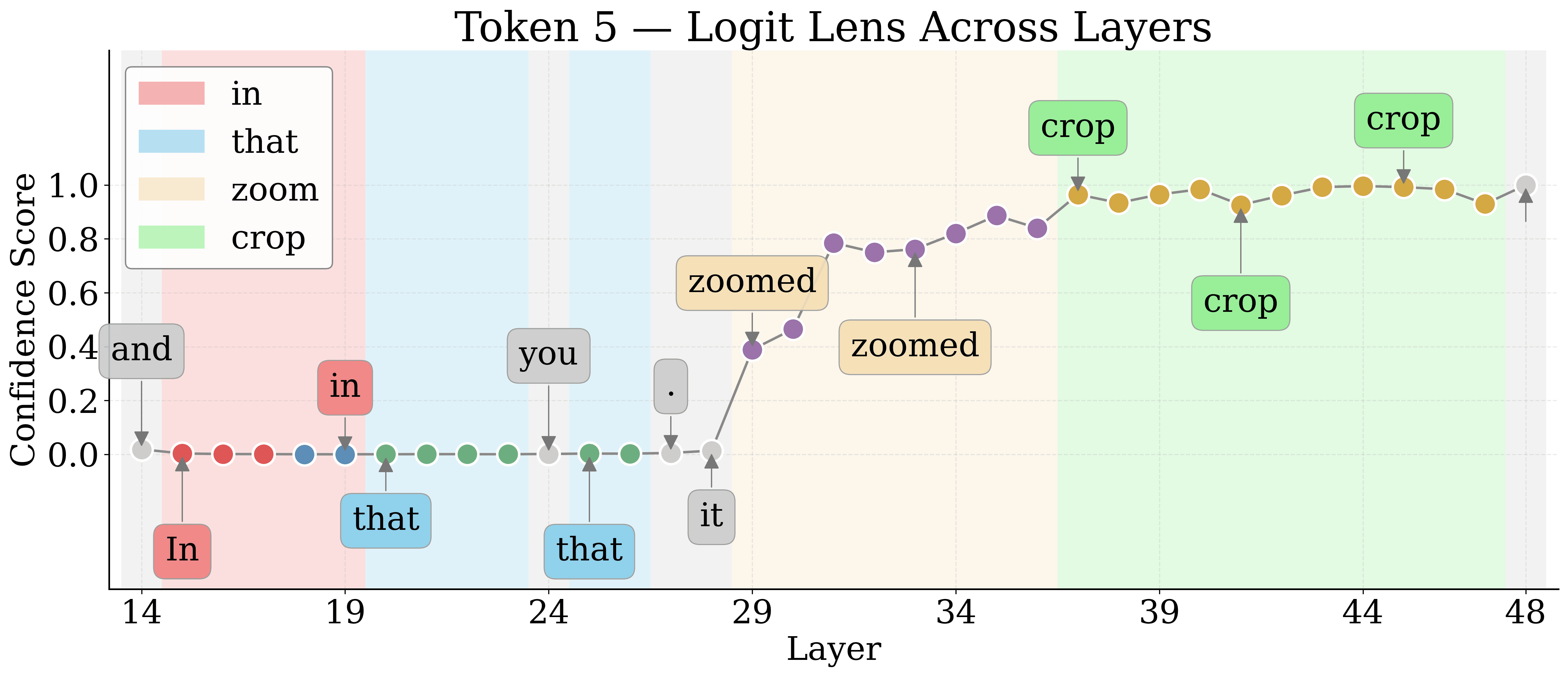}
    \end{subfigure}
        \begin{subfigure}[b]{0.33\textwidth}
        \centering
        \includegraphics[width=\textwidth]{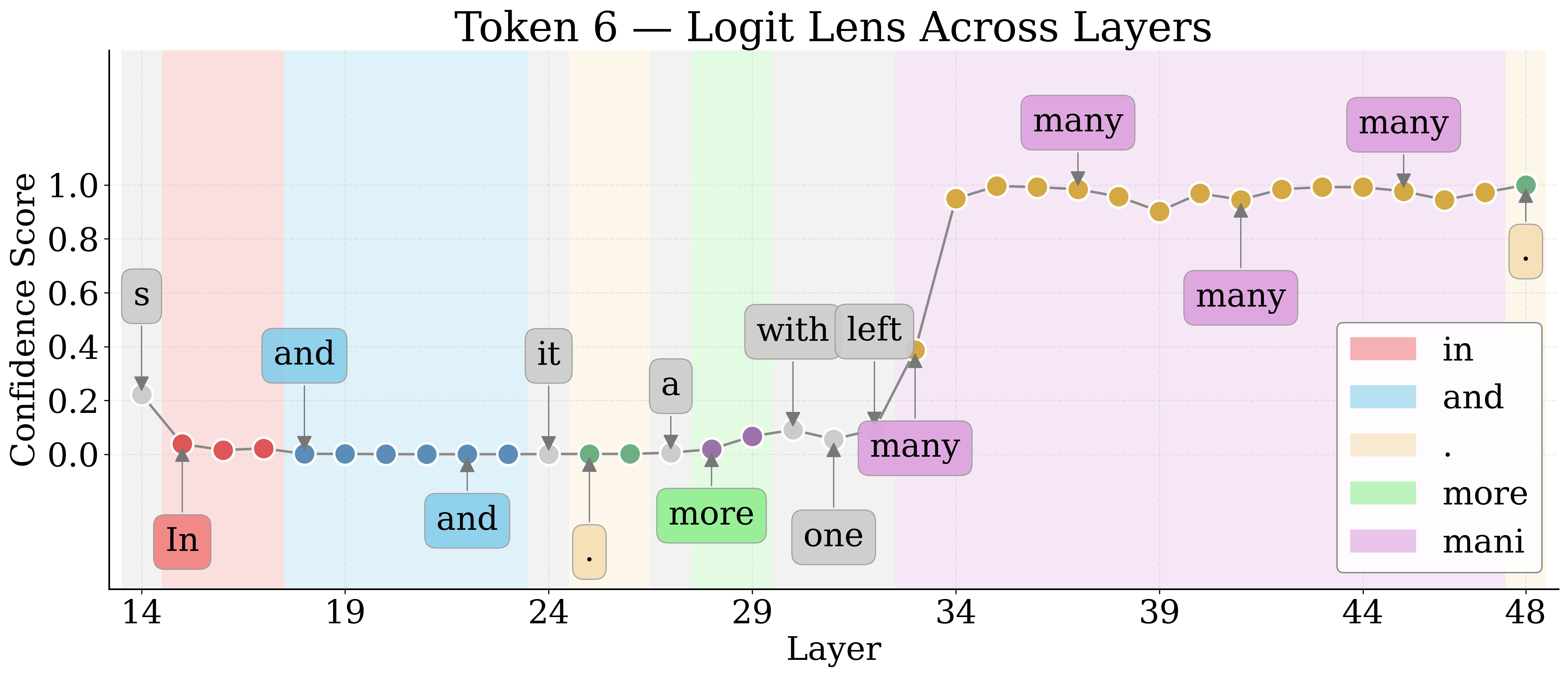}
    \end{subfigure}
    \hfill
    \begin{subfigure}[b]{0.32\textwidth}
        \centering
        \includegraphics[width=\textwidth]{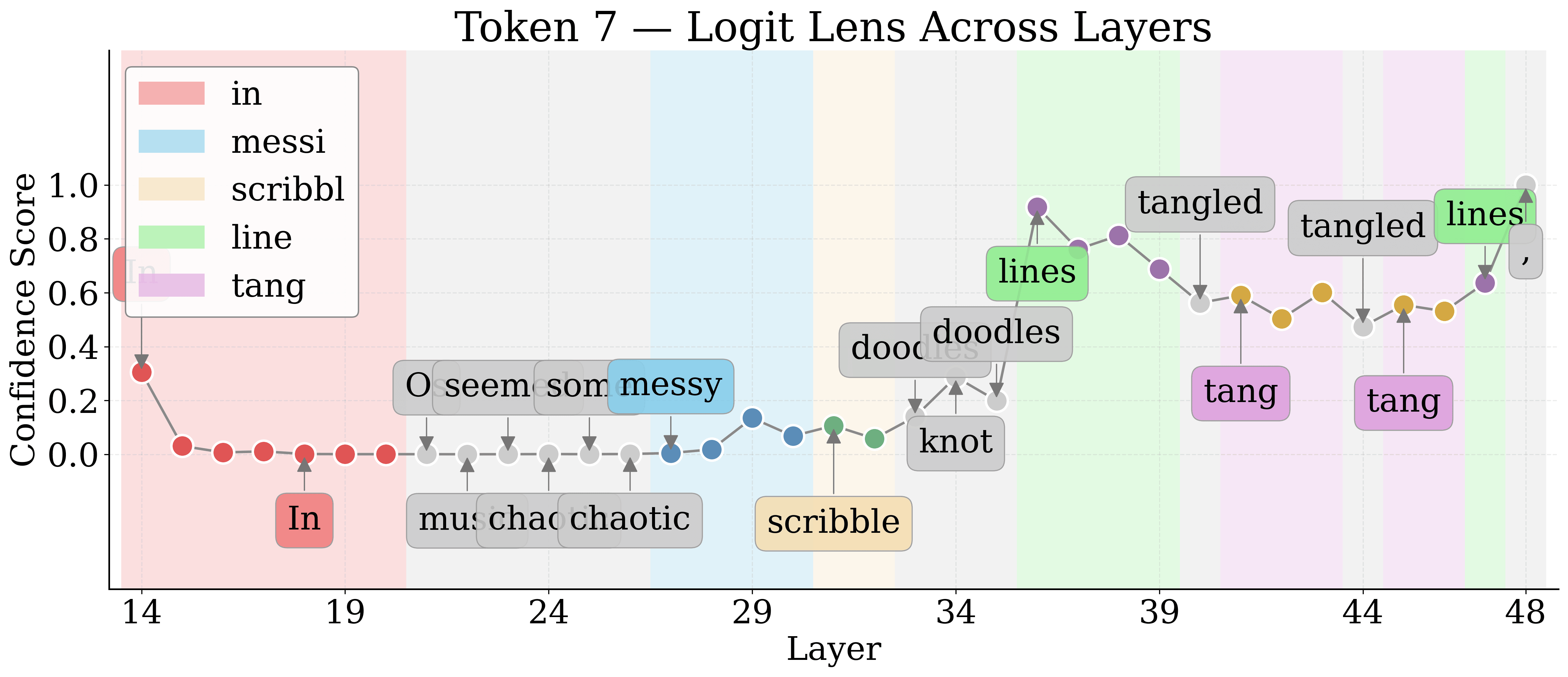}
    \end{subfigure}
    \hfill
    \begin{subfigure}[b]{0.33\textwidth}
        \centering
        \includegraphics[width=\textwidth]{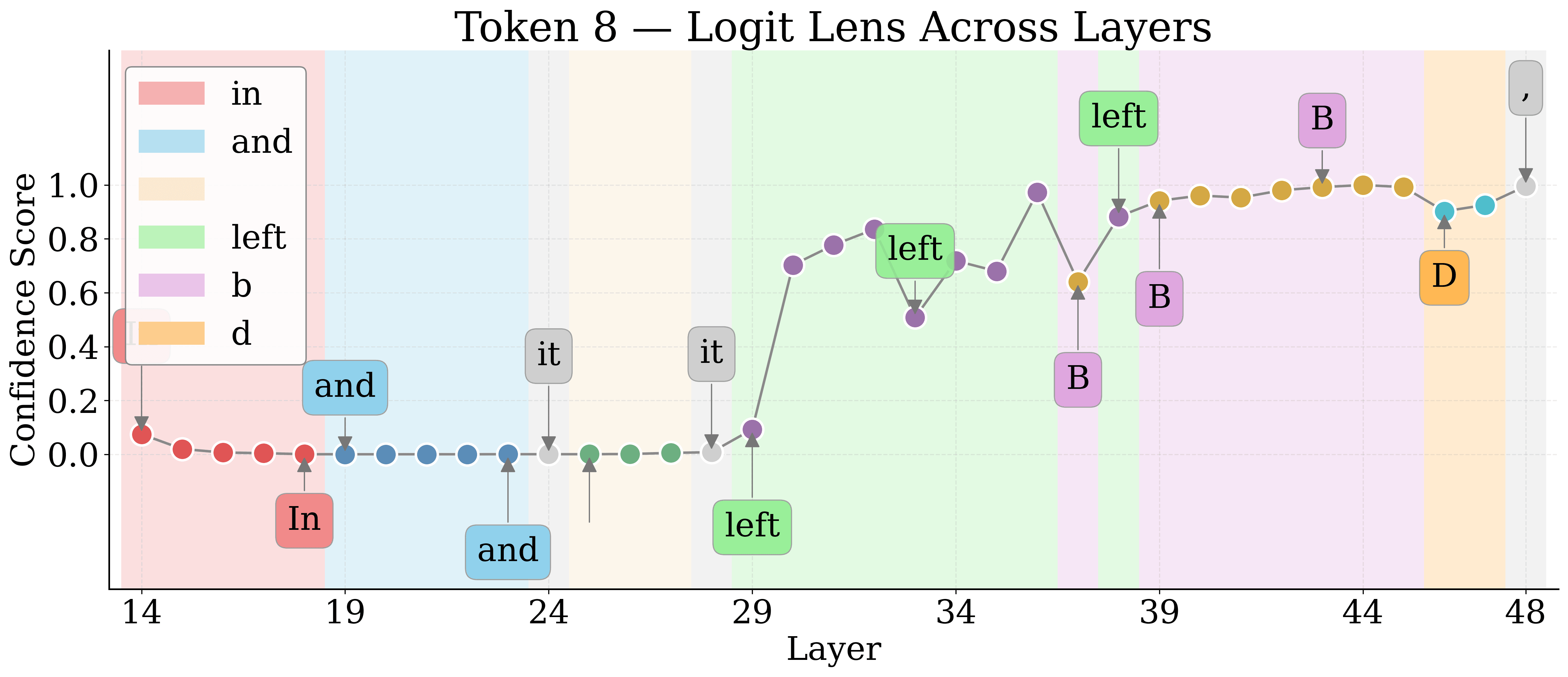}
    \end{subfigure}
        \begin{subfigure}[b]{0.33\textwidth}
        \centering
        \includegraphics[width=\textwidth]{images/UNKNOWN_LOGIT_LENS/squiggle_1_idx34_token9.png}
    \end{subfigure}
    \hfill
    \begin{subfigure}[b]{0.32\textwidth}
        \centering
        \includegraphics[width=\textwidth]{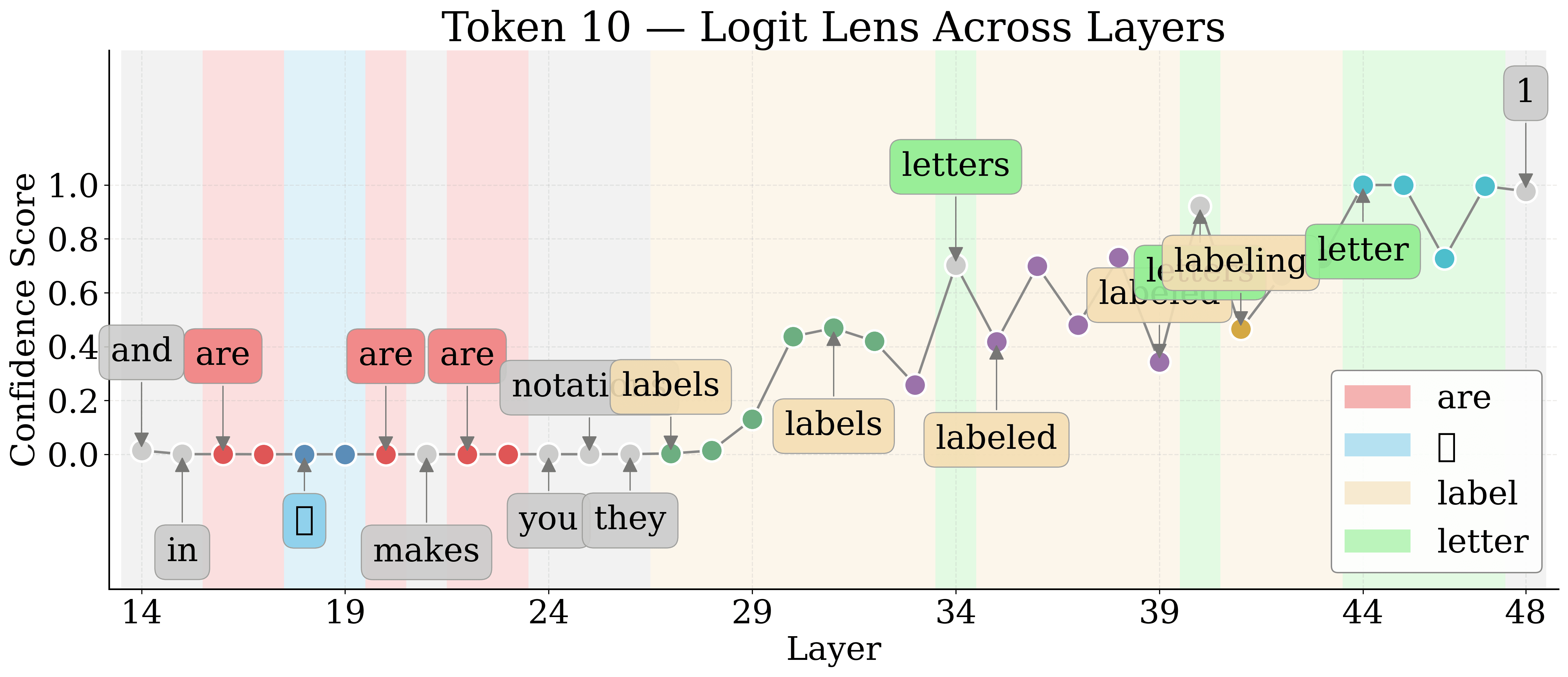}
    \end{subfigure}
    \hfill
    \begin{subfigure}[b]{0.33\textwidth}
        \centering
        \includegraphics[width=\textwidth]{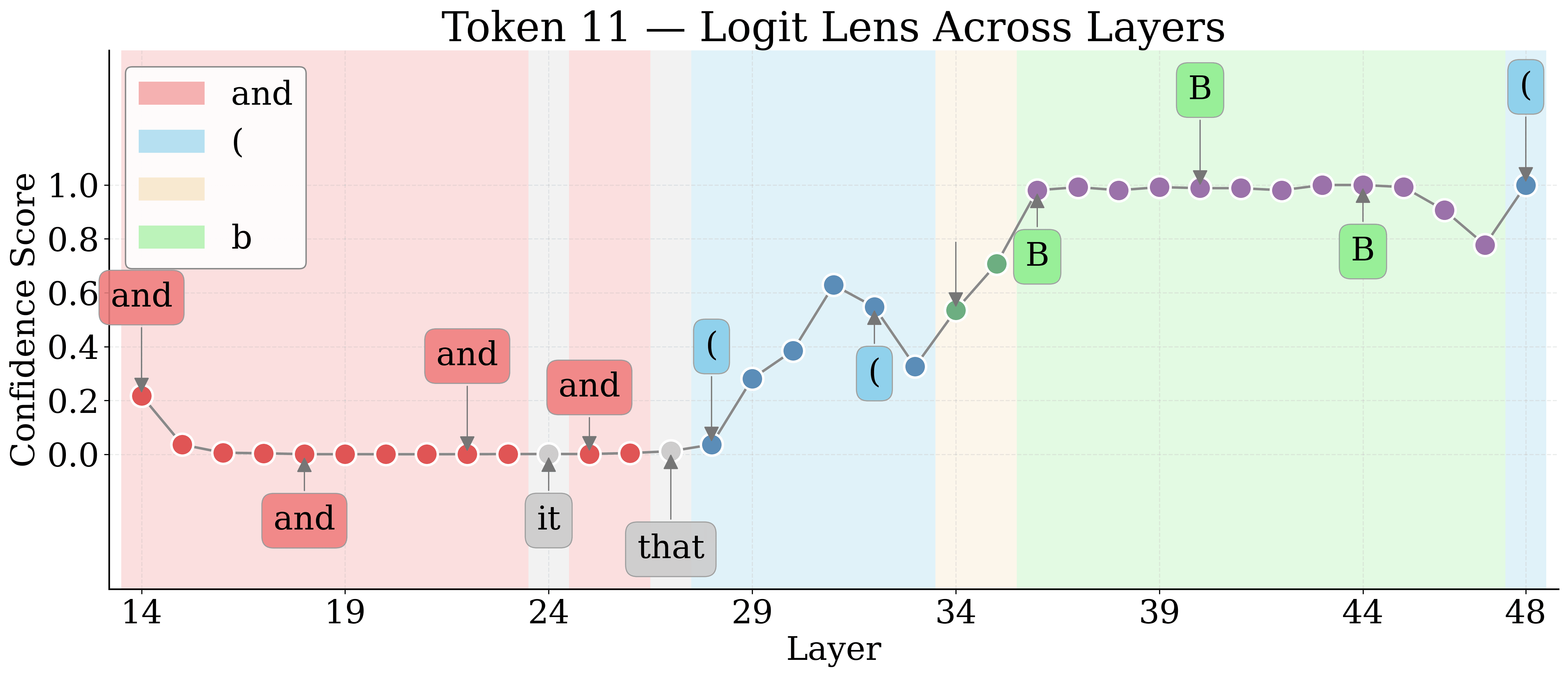}
    \end{subfigure}

        \begin{subfigure}[b]{0.33\textwidth}
        \centering
        \includegraphics[width=\textwidth]{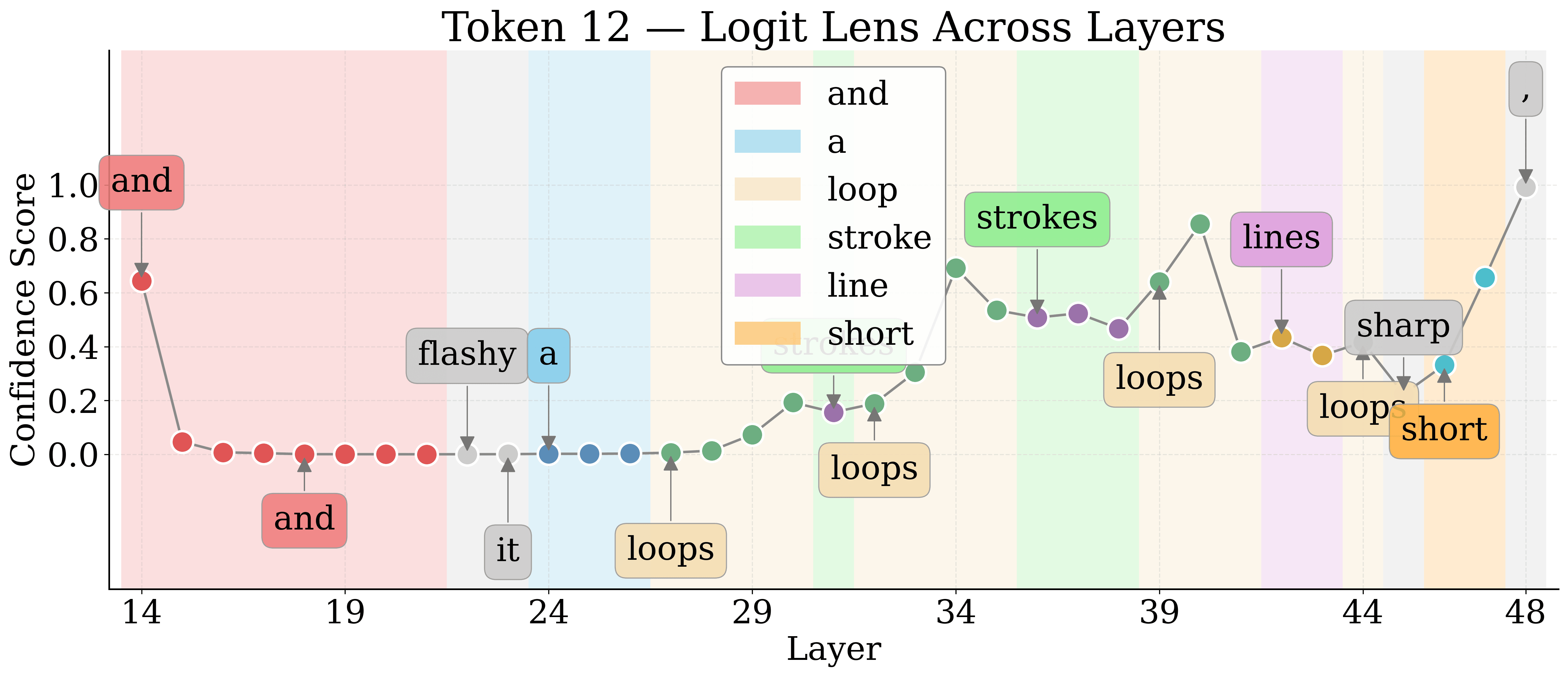}
    \end{subfigure}
    \hfill
    \begin{subfigure}[b]{0.32\textwidth}
        \centering
        \includegraphics[width=\textwidth]{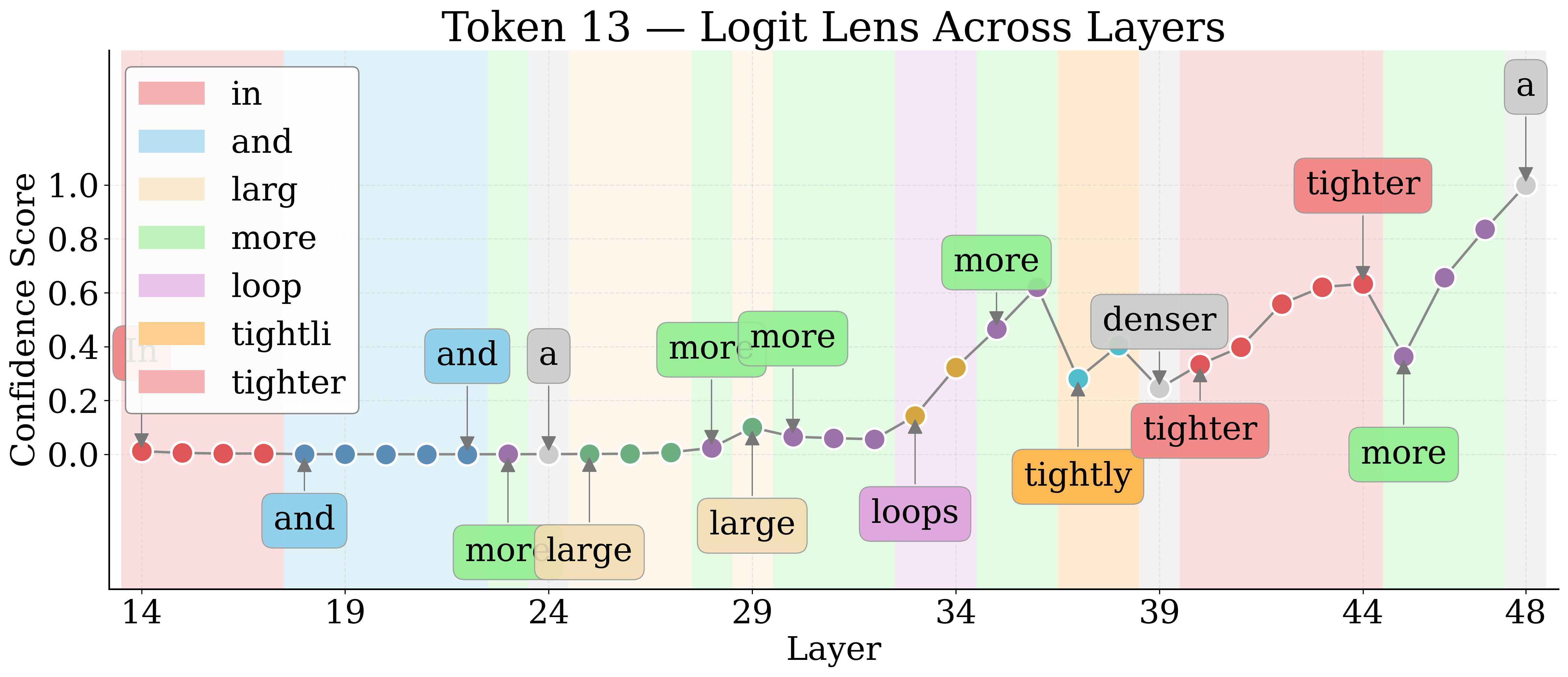}
    \end{subfigure}
    \hfill
    \begin{subfigure}[b]{0.33\textwidth}
        \centering
        \includegraphics[width=\textwidth]{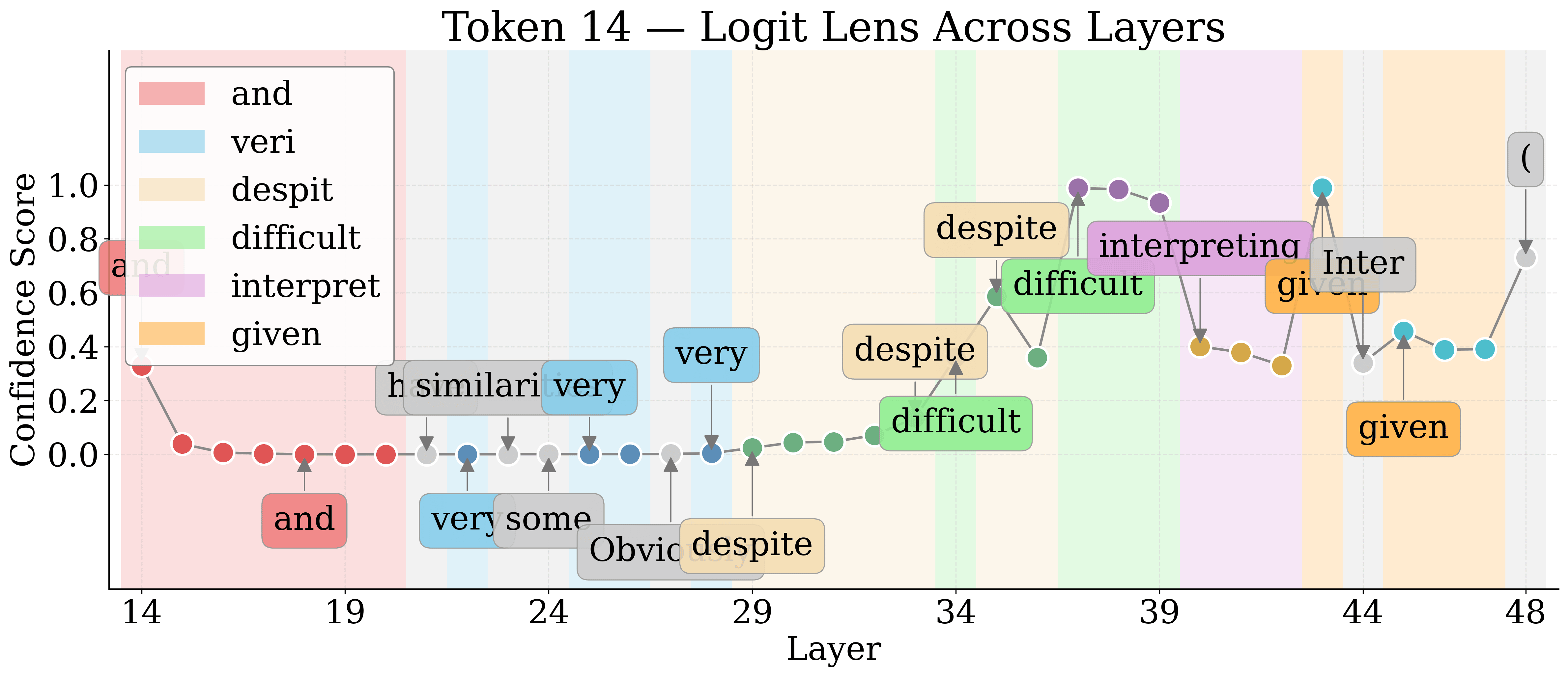}
    \end{subfigure}
    \caption{All Gemma3-12B Logit Lens tokens for \protect\squiggleZ.  (Unknown Shape 2)}
\end{figure}

\clearpage

\section{Teaching Arbitrary Names To VLMs}
\label{app:teaching_names}

\begin{table}[h]
\centering
\begin{tabular}{lp{0.8\textwidth}}
\toprule
\textbf{Name Set} & \textbf{Names} \\
\midrule
Random & 0QK2Z2, 5F1FT3, OZ0W0M, ALCTDF, DNXXB0, ION17F, K0XQNF, UTNWY7, JT1GWQ, 1VZS0M \\
Human & John, Mary, Charles, Elizabeth, William, Margaret, James, Catherine, Robert, Dorothy \\
Ordinary & cup, brick, anchor, fork, bell, shield, blade, horn, nest, arrow \\
\bottomrule
\end{tabular}
\end{table}

\subsection{Finetuning Tasks for Teaching Names}
\begin{table}[h]
\centering
\begin{tabular}{p{0.15\textwidth}p{0.15\textwidth}p{0.35\textwidth}p{0.25\textwidth}}
\toprule
\textbf{Task} & \textbf{Image Type} & \textbf{Example Prompt} & \textbf{Expected Answer} \\
\midrule
Naming & Single Image & ``What is this shape called?'' & Mary \\
\midrule
Yes/No & Single Image & ``Is this a cup?'' & Yes. \\
\midrule
Choice & Single Image & ``Which of the following: fork, cup, bell, arrow?'' & cup \\
\midrule
Comparison & Single Image (Two Shapes) & ``Which object is the arrow, A or B?'' & A \\
\midrule
Description & Single Image & ``Can you describe the shape called John?'' & A name derived from the Latin `Iohannes'\ldots \\
\bottomrule
\end{tabular}
\end{table}

\subsection{Representation Probing After Teaching Names}
\begin{figure}[h]
    \centering   
    \begin{subfigure}[b]{0.49\textwidth}
        \centering
        \includegraphics[width=\textwidth]{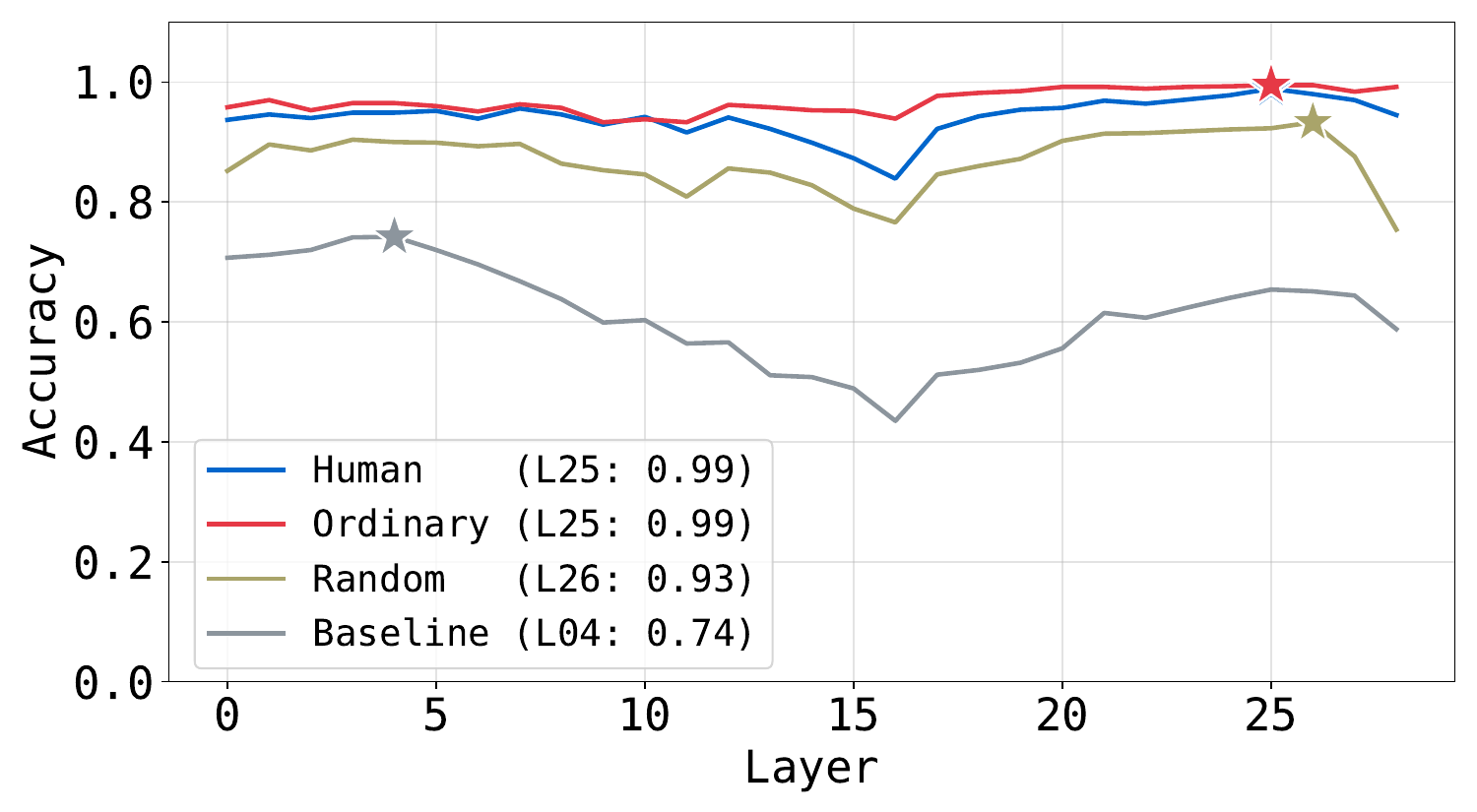}
        \caption{Qwen3-VL 2B}
    \end{subfigure}
    \hfill
    \begin{subfigure}[b]{0.49\textwidth}
        \centering
        \includegraphics[width=\textwidth]{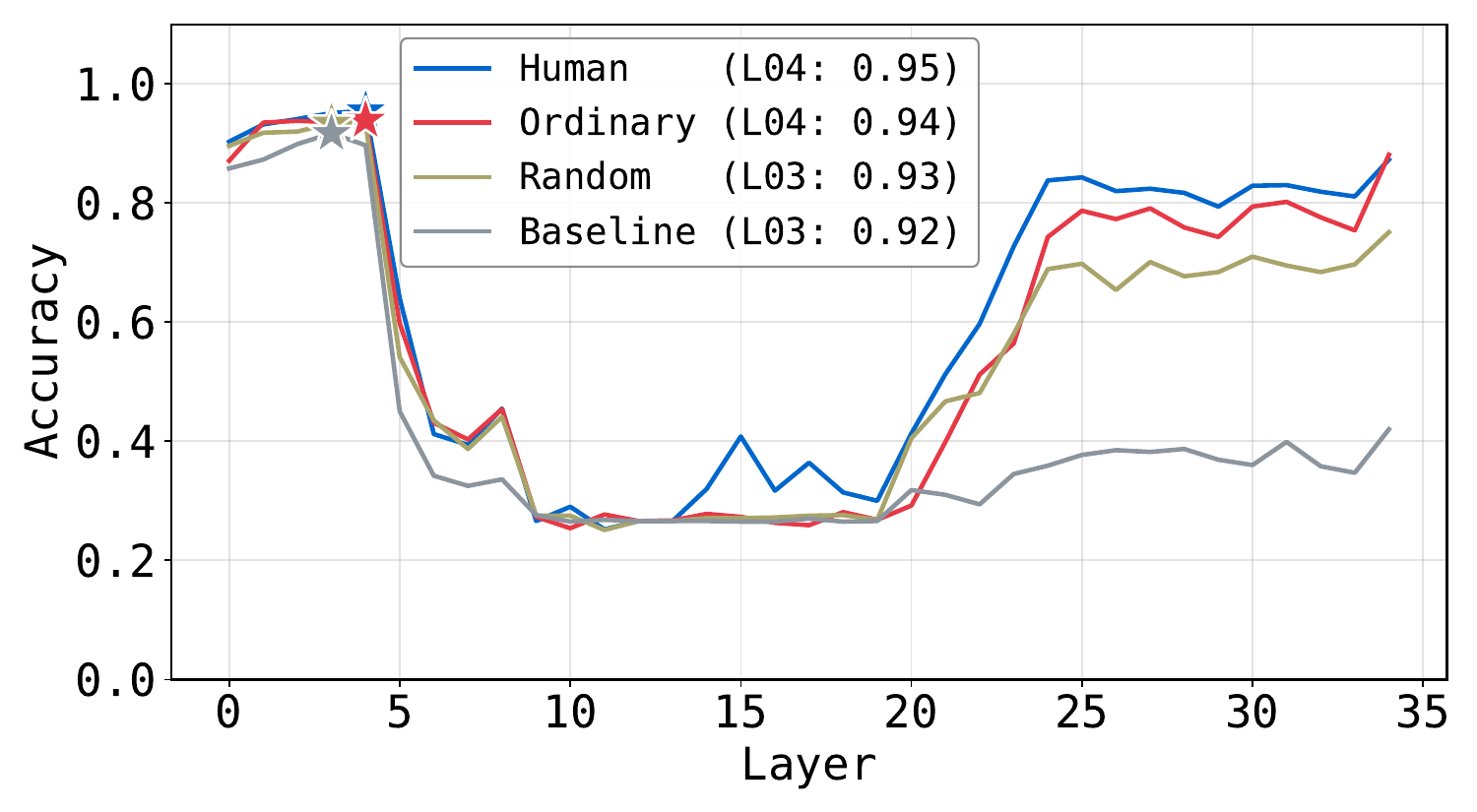}
        \caption{Gemma3-4B}
    \end{subfigure}

    \caption{Representation Probing accuracy increases after teaching VLM names through finetuning. For Qwen, the improvement is consistent across all layers, matching the gap reported in Figure \ref{fig:qwen2b_shapes_rep}. For Gemma, there is a sharp decline for all models after Layer 4, followed by a sharp recovery at Layer 19. This matches the behavior for Known shapes shown in Figure \ref{fig:gemma4b_shapes_rep}.}
    \label{fig:probe_after_names}
\end{figure}

\clearpage
\subsection{Chain-of-Thought Example of Name-Trained Qwen3VL}
\label{app:cot_examples}

\begin{figure}[!h]
    \centering
    \includegraphics[width=1\linewidth]{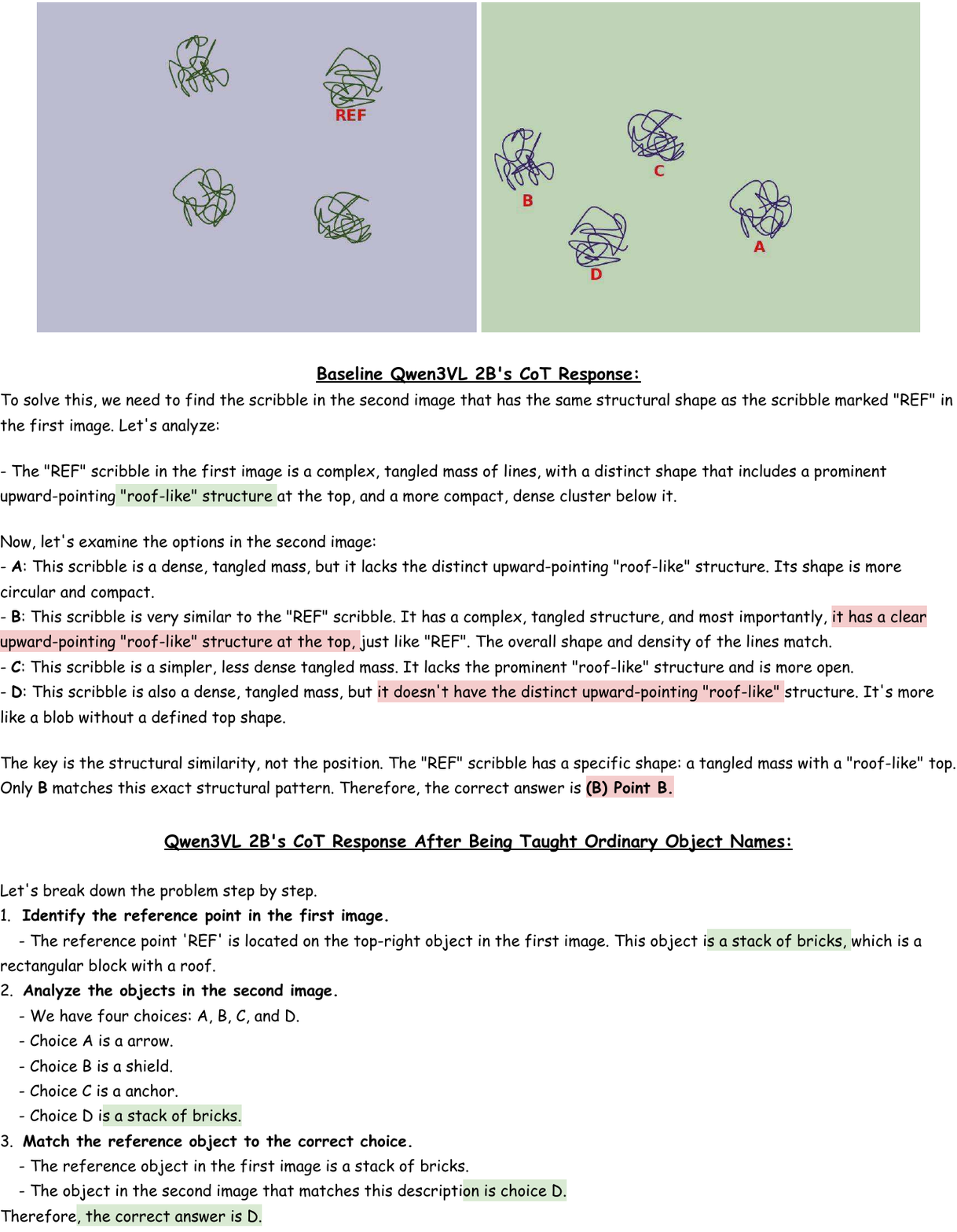}
    \caption{Chain-of-thought example on unknown shapes. Without a semantic anchor, the VLM generates hallucinated descriptions that actively mislead its final answer.}
    \label{fig:cot_example}
\end{figure}

\clearpage
\section{Extreme OOD Test After Task-Specific Finetuning With Squiggles}
\label{app:face_sem_after_ft}
\begin{table}[h]
\centering
\begin{tabular}{l|ccc|ccc}
\toprule
 &\multicolumn{3}{c}{Faces} & \multicolumn{3}{c}{Sem. Corr.}\\
\midrule
Model & Split& Base&FT & Split& Base&FT\\
      & & Acc.&Acc. & & Acc&Acc\\
\multirow{2}{*}{Qwen3VL-2B}& Known   & 77.1 & 89.8 & Named   & 36.6 & 52.3  \\
                           & Unknown & 41.1 & 60.4 & No-Name & 32.4 & 38.1  \\
\multirow{2}{*}{Gemma3-4b} & Known   & 49.8 & 60.8 & Named   & 26.1 & 35.50 \\
                           & Unknown & 32.4 & 38.8 & No-Name & 25.7 & 27.80 \\
\end{tabular}
\caption{Semantic and face correspondence performance after finetuning on squiggles ($n{=}30$). Base Acc.\ is the pre-finetuning baseline. FT Acc.\ is after finetuning.}
\label{tab:on_task_extreme}
\end{table}

\subsection{Transfer to Open-Ended Visual Change Description}
\label{app:clevr_change}

To test whether correspondence finetuning transfers beyond the multiple-choice format used during training, we evaluate Qwen3VL-2B on an open-ended visual change-description task derived from CLEVR-Change \citep{park2019robust}. CLEVR-Change contains pairs of synthetic scenes with controlled changes in object attributes and spatial configurations, making it visually distinct from the procedurally generated squiggles used for finetuning. We prompt the model: \textit{``Inspect the shapes, their colors, their positions, and their sizes in the two images. There are subtle differences between the two images. What are they?''} Because this is an open-ended generation task, we use Qwen3-8B as an automatic judge. The judge receives the model response together with the set of valid ground-truth difference descriptions and assigns a score from 0 to 10 based on how well the response matches the ground truth.

\begin{figure}[!h]
\centering
\includegraphics[width=\textwidth]{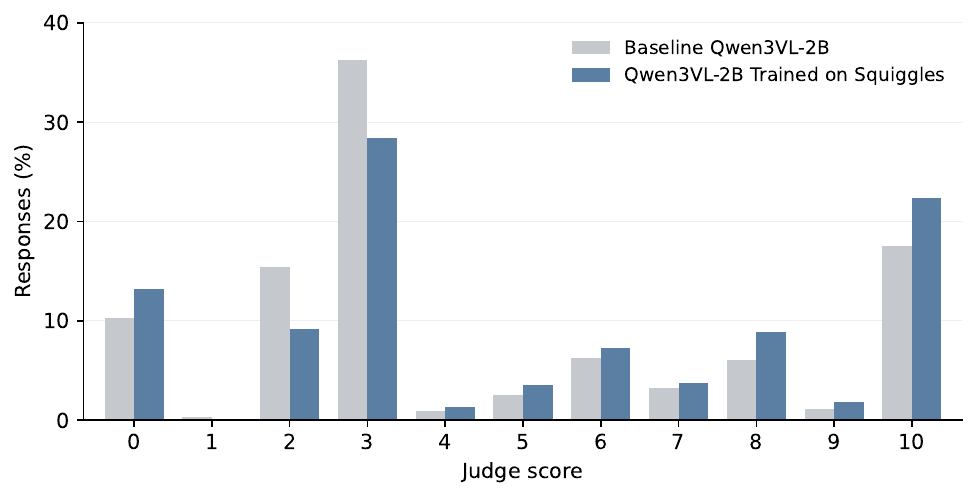}
\caption{Distribution of Qwen3-8B judge scores on open-ended CLEVR-Change responses produced by Qwen3VL-2B before and after finetuning on squiggle correspondence. The finetuned model achieves a higher mean judge score (5.087 vs.\ 4.507) and produces more high-scoring responses (scores 8--10: 33.1\% vs.\ 24.7\%).}
\label{fig:clevr_change_dist}
\end{figure}

Figure~\ref{fig:clevr_change_dist} shows that finetuning shifts the score distribution toward higher-quality responses. The mean judge score improves from 4.507 to 5.087, and the fraction of high-scoring responses (scores 8--10) increases from 24.7\% to 33.1\%.

This experiment differs from the training task in both visual domain and output format. The model is not selecting among annotated candidates or producing a fixed label, but instead must generate a free-form description of a subtle change. The improvement therefore provides evidence that correspondence finetuning does not merely teach the multiple-choice answer format, but transfers to a broader visual comparison behavior.

\section{Art Style Matching}
\label{app:art_style}

\begin{figure}[h]
    \centering   
    \begin{subfigure}[c]{0.20\textwidth}
        \centering
        \includegraphics[width=\textwidth]{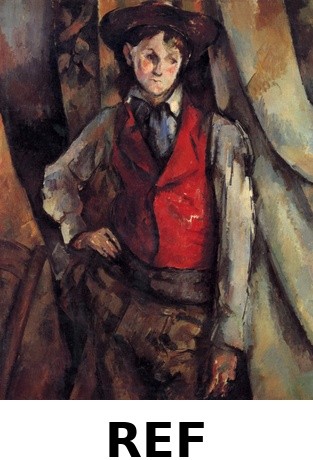}
        \caption{Art Reference.}
    \end{subfigure}
    \hspace{5em}
    \begin{subfigure}[c]{0.50\textwidth}
        \centering
        \includegraphics[width=\textwidth]{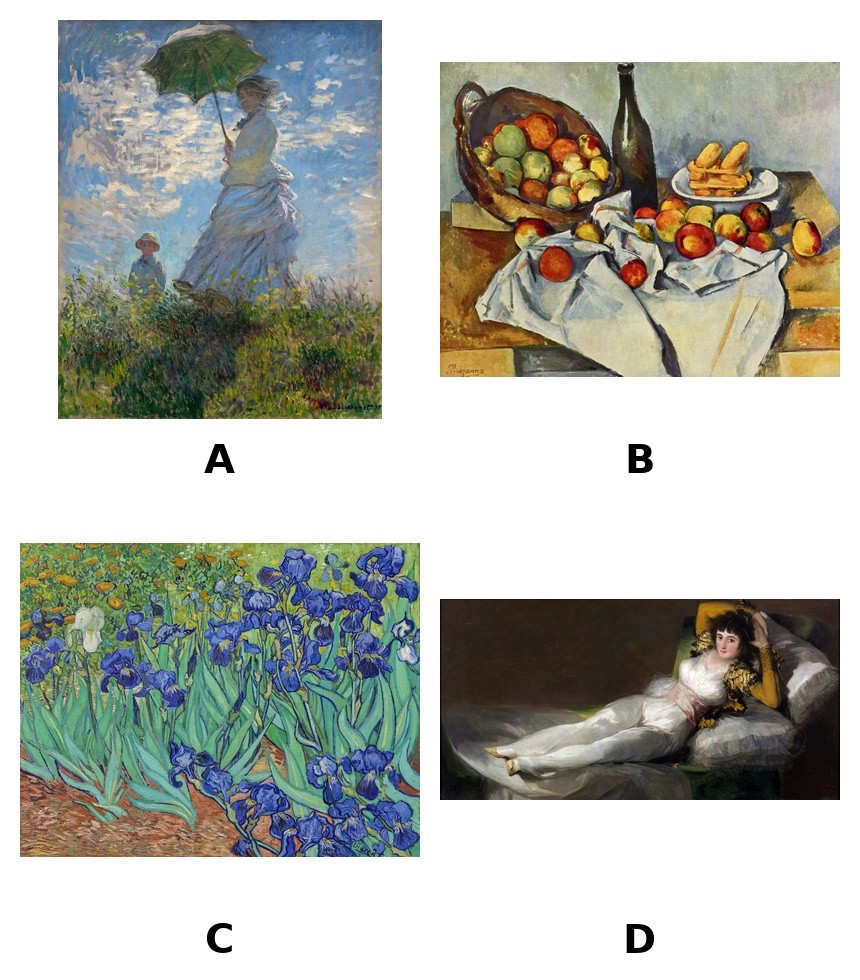}
        \caption{Art Style Option Panel.}
    \end{subfigure}
    \caption{Art Style task example. The ground truth answer is B — both the reference portrait and option B are by Paul Cézanne.}
    \label{fig:art_style_task}
\end{figure}

Our main experiments study semantic anchoring in tasks that require matching specific objects, object parts, shapes, or identities. We additionally test whether the same phenomenon appears in a more holistic visual task: \emph{art style matching}. Unlike fine-grained correspondence, this task requires comparing global properties such as texture, color usage, brushwork, and composition. However, when the painter is recognizable, a VLM may bypass this visual comparison by identifying the painter of each artwork and matching the resulting names.

\paragraph{Dataset construction.}
We collect artworks from painters with visually distinctive styles, including Goya, Rembrandt, and Monet. For each VLM, we first probe whether it recognizes the painter of each artwork using the prompt: \textit{``Which painter created this painting?''} We use the resulting predictions to construct model-specific \emph{Known} and \emph{Unknown} subsets. In the Known subset, the model can identify the relevant painters, whereas in the Unknown subset it cannot.

Each evaluation sample contains one reference artwork and a target image containing four candidate artworks. We prompt the model: \textit{``Which painting in the second image has the same art style as the painting in the first image?''} We treat two artworks as having the same style when they were created by the same painter. This is an approximation, since an artist's style may vary across works and periods. We mitigate this limitation by selecting painters with visually distinctive styles and using multiple artworks with varied subject matter.

\paragraph{Evaluation.}
We evaluate textual performance using the same Direct and Chain-of-Thought (CoT) setups. We also evaluate the information available in the models' internal representations using two training-free comparison strategies. First, we apply the MaxSim representation probe used in our correspondence experiments. Second, we compare the representations using Gram-matrix distance, which captures correlations between features and has commonly been used to represent visual style \citep{gatys2015neuralalgorithmartisticstyle}. For each representation-based method, we report the accuracy from the best-performing language-model layer.

\begin{table}[t]
\centering
\small
\resizebox{\linewidth}{!}{
\begin{tabular}{llcccc}
\toprule
& & \multicolumn{2}{c}{\textbf{Textual Evaluation}} & \multicolumn{2}{c}{\textbf{Representation Probing}} \\
\cmidrule(lr){3-4}
\cmidrule(lr){5-6}
\textbf{Model} & \textbf{Subset} & \textbf{Direct} & \textbf{CoT} & \textbf{MaxSim} & \textbf{Gram Dist.} \\
\midrule
\multirow{2}{*}{Qwen3VL-2B} & Known & \textbf{65.4} & \textbf{59.4} & \textbf{65.2} & \textbf{62.0} \\
& Unknown & 48.6 & 45.8 & 51.4 & 51.6 \\
\midrule
\multirow{2}{*}{Qwen3VL-8B} & Known & \textbf{72.0} & \textbf{73.0} & \textbf{68.2} & \textbf{62.6} \\
& Unknown & 47.8 & 37.2 & 51.2 & 51.8 \\
\midrule
\multirow{2}{*}{Gemma3-4B} & Known & 35.6 & 37.8 & \textbf{53.6} & \textbf{47.8} \\
& Unknown & \textbf{36.0} & \textbf{42.4} & 40.0 & 33.8 \\
\midrule
\multirow{2}{*}{Gemma3-12B} & Known & \textbf{58.0} & \textbf{56.0} & \textbf{55.0} & \textbf{55.2} \\
& Unknown & 39.2 & 40.6 & 31.8 & 32.2 \\
\bottomrule
\end{tabular}
}
\caption{Art style matching accuracy on model-specific Known and Unknown subsets. Known artworks are created by painters that the corresponding VLM can identify, while Unknown artworks are created by painters that it does not identify. Bold indicates the better-performing subset for each model and evaluation strategy.}
\label{tab:art_style}
\end{table}

\paragraph{Results.}
Table~\ref{tab:art_style} shows that the effect of semantic anchors extends beyond matching individual objects and local visual details. Qwen3VL-2B, Qwen3VL-8B, and Gemma3-12B perform substantially better on Known artworks than on Unknown artworks under both textual evaluation strategies. The largest textual gap occurs for Qwen3VL-8B, whose CoT accuracy decreases from $73.0\%$ on Known artworks to $37.2\%$ on Unknown artworks. This is consistent with the model identifying recognizable painters and converting the visual style-matching problem into a comparison between semantic labels.

The representation-based evaluations generally show the same pattern. For Qwen3VL-2B, Qwen3VL-8B, and Gemma3-12B, both MaxSim and Gram-matrix distance perform better on known artworks. This suggests that recognizable artworks also become more distinguishable within the models' internal representations. Gemma3-4B is the main exception: its textual performance is weak and does not show a Known-over-Unknown advantage, although both representation-based strategies still perform substantially better on the Known subset. As in our other experiments, smaller models may lack the baseline capability required to reliably exploit semantic anchors through textual generation.

Unlike the correspondence experiments, representation probing does not consistently outperform the textual strategies on art style matching. For example, Qwen3VL-8B obtains $73.0\%$ with CoT but only $68.2\%$ with MaxSim. This does not necessarily indicate that the relevant information is absent from the hidden representations. Rather, MaxSim and Gram-matrix distance are fixed, training-free comparison functions and may not capture the holistic and spatially distributed features required to distinguish artistic styles. A learned probe could potentially extract this information more effectively.

Overall, these results provide additional evidence that semantic anchoring is not limited to fine-grained correspondence. Even for a task requiring holistic visual comparison, VLM performance is generally stronger when the visual entities can first be mapped to recognizable language concepts.